\documentclass{article} % For LaTeX2e
\usepackage[final]{colm2026_conference}

\usepackage{microtype}
\usepackage{hyperref}
\usepackage{url}
\usepackage{booktabs}

\usepackage{lineno}

\usepackage{natbib}
\usepackage{xcolor}
\usepackage{graphicx} % Required for inserting images
\usepackage{algpseudocode}
\usepackage{amsmath, amssymb, amsthm}
\usepackage{caption}
\usepackage{subcaption}
\usepackage{enumitem}
\usepackage[capitalize]{cleveref}
\usepackage{wrapfig}
\usepackage{placeins}
\usepackage[most]{tcolorbox}
\usepackage[multiple]{footmisc}
\usepackage{xcolor}
\usepackage{footnote}
\definecolor{revblue}{HTML}{1A5FB4}
\newenvironment{revision}{\color{black}}{}
\definecolor{myblue}{HTML}{4C72B0}
\newcommand{\rev}[1]{{\color{black}#1}}

\tcbset{
contextpollutionbox/.style={
    colback=gray!10,
    colframe=gray!60,
    boxrule=0.8pt,
    arc=4pt,
    left=6pt,
    right=6pt,
    top=6pt,
    bottom=6pt
}
}

\tcbset{userbubble/.style={
  colback=myblue!14, colframe=myblue,
  boxrule=0.3pt, arc=4pt, outer arc=4pt,
  left=2pt, right=2pt, top=1.5pt, bottom=1.5pt,
  fontupper=\scriptsize\itshape,
  before upper={\raggedright},
  enhanced, overlay={
    \fill[myblue] ([xshift=3pt,yshift=0pt]frame.north west)
      -- ++(0,4pt) -- ++(4pt,-4pt) -- cycle;
  }
}}
\newcommand{\turnlabel}[1]{\vspace{2pt}\noindent{\scriptsize\bfseries #1}\vspace{1pt}}
\newcommand{\exampletitle}[1]{\noindent{\footnotesize\bfseries #1}}
\newcommand{\exampledesc}[1]{\noindent{\scriptsize #1}\vspace{3pt}}

\definecolor{darkblue}{rgb}{0, 0, 0.5}
\hypersetup{colorlinks=true, citecolor=darkblue, linkcolor=darkblue, urlcolor=darkblue}

\title{Do LLMs Benefit From Their Own Words?}

\author{
Jenny Y. Huang$^{1,2}$,
Leshem Choshen$^{1,2,3}$,
Wei Sun$^{2,3}$,
Omar Khattab$^{1}$, \\
\bfseries
Ramón Fernandez Astudillo$^{2,3}$,
Mehul Damani$^{1}$,
Tamara Broderick$^{1,2}$\thanks{Equal advising.},
Jacob Andreas$^{1,2}$\footnotemark[1] \\
\normalfont
$^{1}$Department of Electrical Engineering and Computer Science, MIT \\
$^{2}$MIT--IBM Watson AI Lab \qquad
$^{3}$IBM Research \\
\texttt{jhuang9@mit.edu}
}

\begin{document}

\ifcolmsubmission
\linenumbers
\fi

\maketitle

\begin{abstract}
In multi-turn conversations, large language models typically condition on the full conversation history: both past user prompts and assistant responses. We revisit this design choice by comparing full-context prompting to four alternative, substantially-reduced context configurations. Analyzing in-the-wild multi-turn conversations across three open reasoning and one state-of-the-art model, 
\rev{we find that response quality is largely preserved under aggressive context filtering: replacing all prior assistant turns with one-sentence summaries or keeping only the most recent user--assistant exchange often matches storing full context in performance while using roughly $8\times$ less context.}
To understand this result, we observe that a substantial fraction of user turns ($36.4\%$) in multi-turn conversations are self-contained and that many follow-up turns can be addressed by seeing only the immediately preceding user--assistant exchange. 
Furthermore, we find that when models condition on their own past responses, this can lead to \textit{context pollution}, a phenomenon in which reasoning errors, hallucinations, or stylistic artifacts propagate across turns.
Motivated by these findings, we design a context-filtering approach that selectively omits the assistant-side history. Taken together, these findings suggest moving away from storing full dialogue transcripts and instead retaining only what is relevant.

% we find that response quality can often be maintained with substantially less context: \textit{frequently achieving comparable performance with up to a \rev{$5\times$} reduction in context length.}
\end{abstract}

\section{Introduction}
\label{sec:introduction}

\begin{wrapfigure}{r}{0.60\textwidth}
    \centering
    \vspace{-12pt}
    \includegraphics[width=\linewidth]{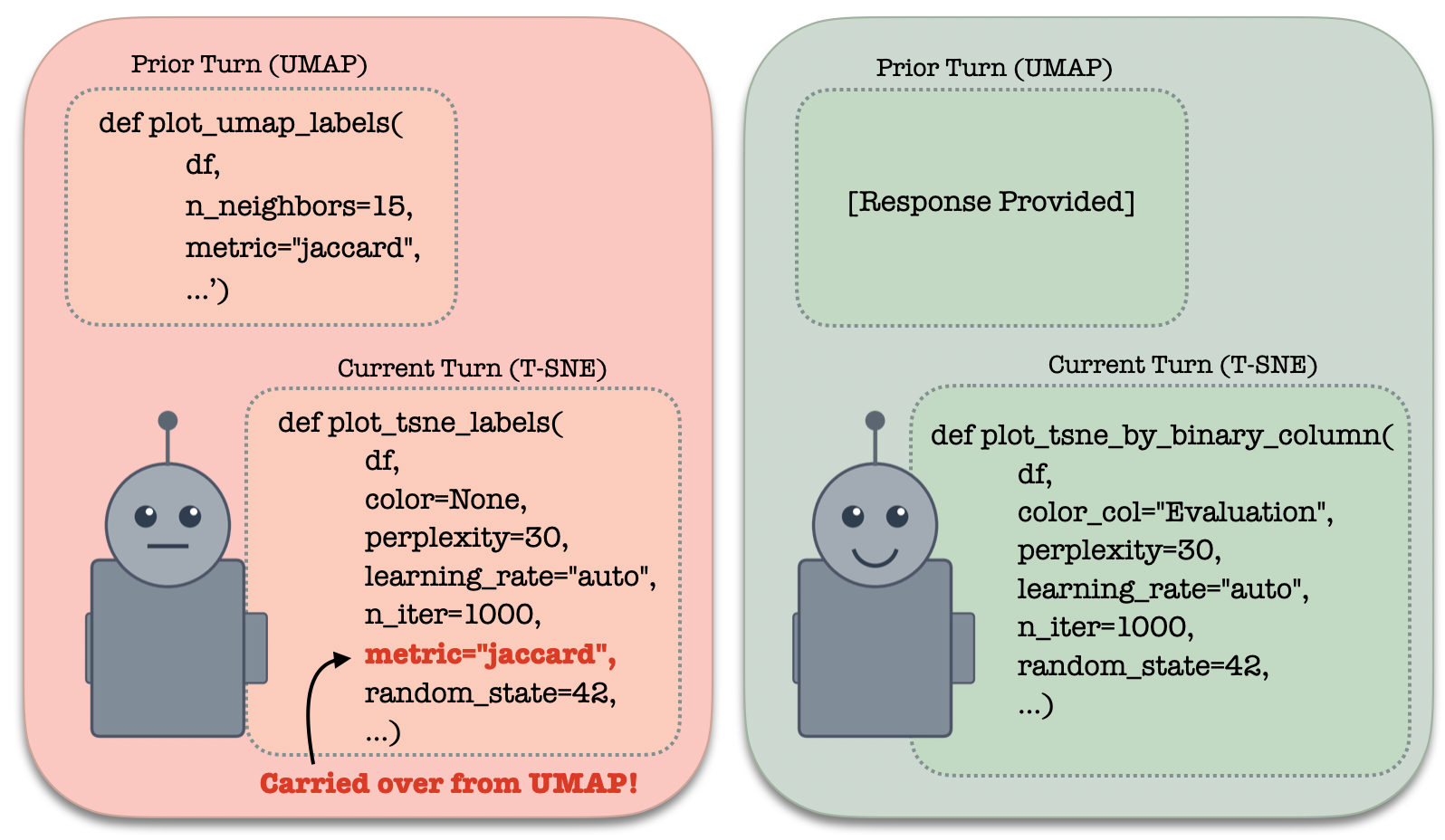}
    \caption{Real example of overrelying on irrelevant context. In Turn 2, the user requests UMAP clustering code. In Turn 5, the user says, ``use t-SNE instead.'' Left: When the previous assistant response remains in context, the model incorrectly carries over the Jaccard metric from UMAP into the t-SNE implementation, introducing a bug. Right: Without the previous response in context, the model generates correct t-SNE code.} % with appropriate arguments
    \label{fig:context-pollution-cartoon}
    \vspace{-26pt}
\end{wrapfigure}

As large language models (LLMs) are deployed in increasingly complex multi-turn conversations, 
context management becomes an important challenge. Long contexts increase computational costs~\citep{eyuboglu2025cartridges,dao2022flashattention,xiao2025improving}, slow inference speeds, and can impair a model's capacity to attend to relevant information~\citep{liu2024lost,lee2026lost}. In response, production systems like Claude Code, Codex, and Cursor have adopted context engineering strategies. These methods include discarding or externalizing tool outputs and reasoning traces~\citep{cursor2026dynamiccontext,anthropic2025contextediting}, summarizing early conversation turns when context limits are reached, and allowing users to manually clear or compact conversation histories~\citep{anthropic2025contextediting}. Despite increasing efforts to compress and prune the older segments of conversation history, one key idea remains largely unexamined: that conditioning on the \textit{full} dialogue transcript, \textit{especially the assistant's own past responses}, is a necessary and beneficial design choice for maintaining downstream response quality.

% that conditioning on past model responses is beneficial for maintaining response quality in multi-turn chats.
%one key assumption remains largely unchallenged: that storing past model outputs is beneficial for maintaining downstream response quality in real-world conversations.
%In this work, we analyze in-the-wild multi-turn chats from WildChat~\citep{zhao2024wildchat} and ShareLM~\citep{don2025sharelm} to ask: \textit{Do current models benefit from conditioning on their own responses?}

An earlier line of works on conversational question answering (ConvQA)~\citep{qu2019attentive,do2022cohs,zaib2023learning} found that only a select subset of past conversation turns are useful for answering current turns and that storing irrelevant turns in context can actually degrade downstream performance. The work, however, analyzed human–human, rather than human-LLM, conversations (e.g., QuAC, CoQA). Unlike human responses, LLM-generated responses contain long, speculative, and error-prone reasoning traces. Such LLM-generated content may introduce unique sources of distraction, 
and even state-of-the-art models tend to be distracted by irrelevant information in context \citep{hong2025context}. \begin{revision}A recent line of work on multi-turn LLM conversations \citep{laban2025lost} documented substantial performance degradation attributable to the model ``taking a wrong turn'' early on in a conversation. Later work \citep{khalid2025ergo} proposed to address this concern by consolidating past user turns into a single prompt when context degradation is detected. Both studies, however, relied on a narrow simulation setup in which user turns are shards of a single instruction revealed incrementally (see properties 1-5 in Appendix~B of \citealp{laban2025lost}). This design fails to capture various dynamics found in real-world multi-turn conversations, such as mid-conversation topic shifts and self-contained user follow-ups (see \cref{tab:canonical-follow-up-prompts}, left).\end{revision} Taken together, these gaps make it unclear whether preserving past LLM-generated responses is beneficial, neutral, or harmful in real-world human--AI conversations, an empirical question we set out to address.

Standard multi-turn prompting assumes that models should condition on the full conversation history. We challenge this assumption by comparing full-context prompting against \rev{four} \textit{substantially}-reduced context configurations: keeping only the last user--assistant exchange preceding the current turn, keeping only the past \textit{user} turns, \rev{keeping the past user turns and a one-sentence summary of each assistant turn}, or deleting the entire preceding conversation history. We analyze in-the-wild, multi-turn conversations~\citep{zhao2024wildchat,don2025sharelm} across three open reasoning and one frontier model. We find, to our surprise, that models can often \textit{maintain comparable response quality in a multi-turn conversation given substantially less context.} 

To investigate this finding, we analyze both the structure of in-the-wild multi-turn conversations and the effects of conditioning on prior model responses, making two key observations:

(1) Multi-turn dependence is not inherent in real-world multi-turn conversations (see \cref{sec:a-little-goes-a-long-way}). Using a language model to classify turns based on their dependence on prior turns, we observe that a substantive fraction ($36.4\%$)\footnote{See details of this experiment in \cref{subsec:prompt-type-analysis}} of turns in real-world multi-turn conversations are largely self-contained. The composition of real-world multi-turn chats in our analysis suggests that careful, generous filtering of past assistant responses may be justified.
%  Only a subset of user prompts ($33.1\%$)\footnote{This value represents an upper bound, as it reports the percentage of prompts classified as Follow-up w/o Feedback; some prompts in this category can be still addressed using only user-side turns.} reference an earlier assistant response without giving actionable feedback on ways to improve or revise the response (see an example in \cref{subsec:follow-up-prompts-still-okay}). Furthermore, we find that many user follow-up prompts provide sufficiently concrete instruction such that the request can be addressed from scratch using the current and prior user turns alone (see examples in \cref{subsec:follow-up-prompts-still-okay}). 

(2) Models tend to over-condition on their past responses, resulting in \textit{context pollution}: a phenomenon in which earlier model-generated outputs introduce errors, hallucinations, or stylistic artifacts that propagate into subsequent turns (see \cref{fig:context-pollution-cartoon} and \cref{sec:too-much-information-is-overwhelming}). This tendency for models to over-condition on their earlier responses has been observed as well in past studies \citep{laban2025lost,sinha2026the,li2025haystack,hong2025context,lee2026lost} and is a behavior that does not seem to mitigate with model size \citep{sinha2026the}. We characterize different forms of context pollution in \cref{sec:too-much-information-is-overwhelming} and provide 11 illustrative examples in \cref{subsec:context-pollution-examples}.

Motivated by these findings, we develop an adaptive assistant-response-omission strategy that trains a classifier to predict, for a given user prompt and conversation history, whether the LLM should condition on its past responses (\cref{sec:adaptive-context-inclusion}).

Together, our findings call into question the design choice of storing full dialogue transcripts in context.

\label{sec:related-works}
\subsection{Context Management}
\paragraph{Single-turn prompt compression.}
A line of past works studies prompt compression in a single-query setting \citep{khattab2021baleen,wang2023learning,xu2024recomp,chuang2024learning,chirkova2025provence}. Many of these works focus on retrieval-augmented generation (RAG), where retrieved documents are filtered or compressed before being provided to the model. Additional single-turn compression methods, such as Lingua-2 \citep{pan2024llmlingua} and PENCIL \citep{yang2025pencil}, remove redundant reasoning traces and are likewise evaluated at the level of individual queries. In this work, we are interested in condensing conversation histories of multi-turn conversations.

\paragraph{Multi-turn context editing.} Prior works on context editing within multi-turn conversations~\citep{jiang2023llmlingua, li2023unlocking} treat the \textit{full} conversation history as a gold standard, evaluating proposed compression methods against full-context-conditioned responses, using similarity metrics like BERTScore and BLEU/ROUGE scores. These works do not question the validity of full-context prompting as the gold standard. A notable exception is ERGO~\citep{khalid2025ergo}, which proposes rewriting the prior user turns into a single prompt while omitting past assistant responses. Their results show that prompt consolidation and assistant-response omission can outperform full-context conditioning on multi-turn math and coding. However, their findings are made exclusively on synthetic conversations.

\paragraph{Context management in agentic systems.} 
Beyond the chat setting, agentic systems continue to develop new context management strategies daily. Production systems like Claude Code  \citep{anthropic2025contextediting}, Codex \citep{lopopolo2026harness}, and others \citep{cursor2026dynamiccontext,liu2025deepseek,qwen2025qwen35,gao2025trae}, clear or compress tool outputs and thinking traces and perform truncation when context windows become saturated. A growing ecosystem of memory tools focuses on storing reusable information or content that is most difficult to reconstruct (e.g., tool outputs, bug fixes, or global constraints) in external files for later reference \citep{anthropic-memory-tool, aidevkit2024memory}. Some works suggest outsourcing the role of context management to a smaller LLM \citep{xiao2025improving,lindenbauer2025complexity}, while others \citep{liu2026pensieve} suggest training the model to manage its own context.

We identify two gaps in the current literature. First, there is a lack of \textit{evaluation on real-world multi-turn conversational data}. Second, both research and deployed systems often treat the storage of full conversations as a default design choice. In this study, using real-world chat logs on Wildchat \citep{zhao2024wildchat} and ShareLM \citep{don2025sharelm}, we find that a substantial portion of conversation histories are rarely referenced in later turns and that the response quality of multi-turn chats can be maintained at a comparable level with minimal context.

\section{Experiment Setup}
\label{sec:setup}
\label{subsec:experiment-setup}
\paragraph{Datasets and Models. }
To evaluate whether retaining the full conversation history provides measurable benefits to downstream response quality, we conduct a controlled experiment across four LLMs spanning a range of model sizes: Qwen3-4B~\citep{yang2025qwen3}, DeepSeek-R1-Distill-Llama-8B~\citep{deepseekr1distillllama8b}, GPT-OSS-20B~\citep{agarwal2025gpt}, and GPT-5.2~\citep{openai2025gpt52}. We use real-world multi-turn conversations drawn from \texttt{allenai/WildChat-4.8M}~\citep{zhao2024wildchat} and \texttt{shachardon/ShareLM}~\citep{don2025sharelm}. Our analysis includes conversations in the domains of math, coding, science, business/finance, education, and creative writing. See Appendix~\ref{subsec:data} for additional details on the datasets and our data filtering process.

\label{subsec:generation}
\paragraph{Generating Responses.}
For each model, we generate responses under four different context configurations: \textbf{Full Context (FC)}, in which the model is prompted with the entire conversation history (both past user prompts and assistant responses); \textbf{Assistant-Omitted (AO)} context, in which the model is prompted with only the content of prior \textit{user} turns, while assistant responses are replaced with the placeholder \texttt{[Response provided]} to maintain the alternating user/assistant turn structure; \textbf{Last-Turn-Only (LT)}, in which the model is prompted with only the most recent user-assistant exchange; \rev{\textbf{Summarized (SUM)}, in which each assistant response is replaced with a one-sentence summary of that response, given by the same model being evaluated prompted with: \texttt{Summarize the following assistant response in exactly one sentence:\{response\}}}; and \textbf{Delete-Everything (DE)}, in which the model receives only the current user prompt with no prior history. See Appendix~\ref{subsec:conditions} for the prompt templates for all four context configurations. For all conversations, we retain the historical user prompts and assistant responses from the data as the conversation history. In the appendix, we present an ablation study where the model's own generated responses are fed back into the context for subsequent rounds (see \cref{fig:colm-winrates-regenerated}).

\label{subsec:response-quality-evaluation}
\paragraph{Evaluating Responses.}
To evaluate responses, we use GPT-5 as an LLM-judge. 
For each conversation round starting from round 2, where the context configurations begin to diverge, 
the LLM-judge receives both the full context and the alternative context responses alongside the full conversation history (all past user and assistant turns). 
It then selects a winner (or declares a tie) for each of two evaluation dimensions: \textit{response quality} (Quality), whether a response clearly and accurately addressed the user's current prompt, and \textit{task adherence} (On-Topic), whether the response remains focused on the conversation thread. 
We randomize response ordering for each comparison to reduce position bias. Since we set out to investigate the impact of distraction from accumulated assistant responses, one natural concern is that the LLM-judge itself may be susceptible to distraction. 
To address this concern, we supplement the full-context LLM-judge with a variant that receives only the prior user turns during evaluation (see \cref{subsec:user-turn-only-judge-prompt}). In our main-text experiments, the judge is given both the thinking trace and the final response to perform its evaluation.\footnote{With the exception of GPT-5.2, whose responses do not expose thinking traces.} In the appendix, we provide an additional experiment in which the judge compares only on final responses (see \cref{subsec:final-answer-only}). We provide the full judge prompts in \cref{subsec:direct-comparison-judge-prompt}.

\section{Response Quality Can Be Maintained Under Much Less Context}
\label{sec:a-little-goes-a-long-way}
\begin{figure}[t]
    \centering
    \includegraphics[width=1.00\linewidth]{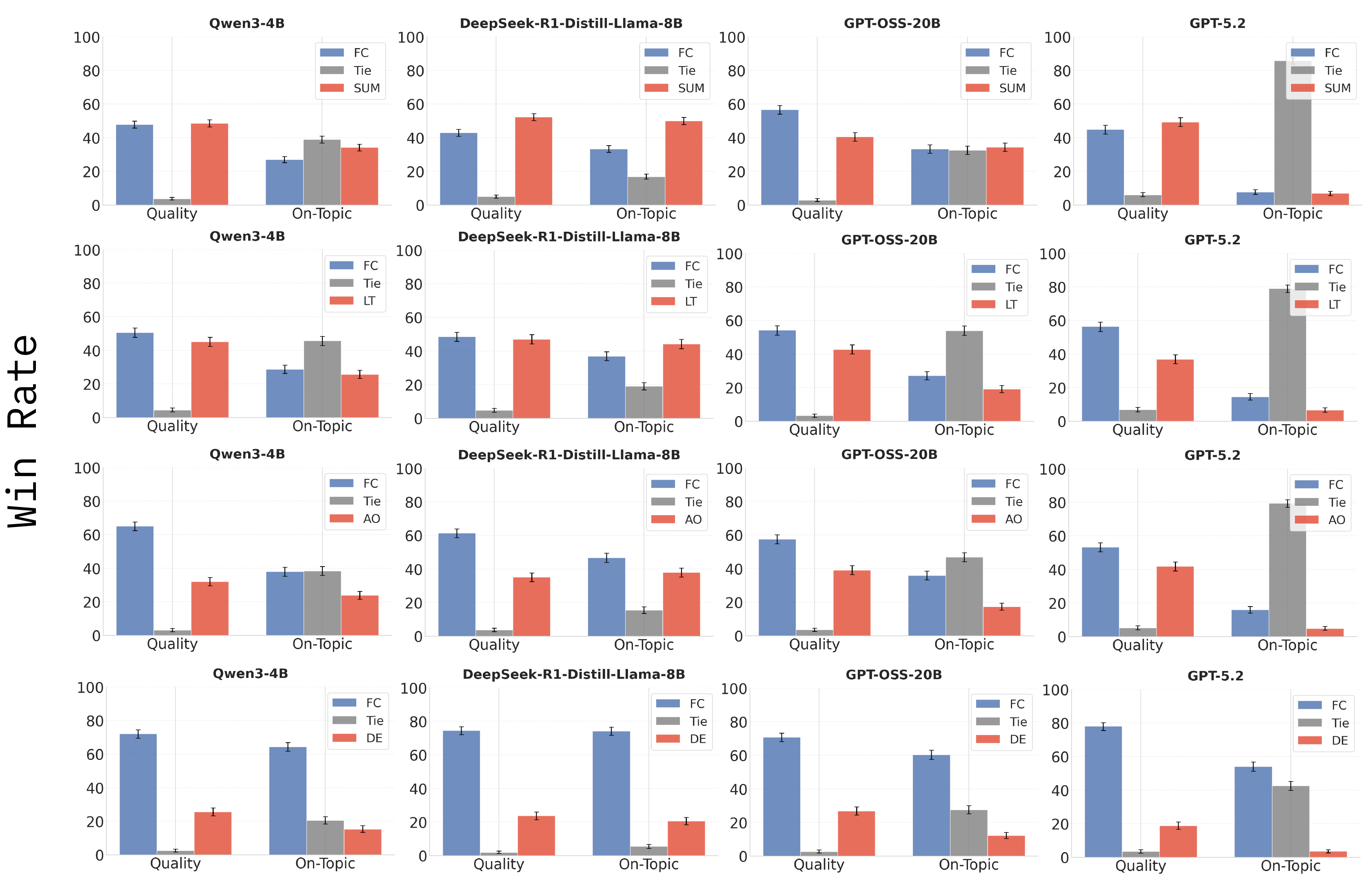}
    \caption{
    %\lc{there's win rate several times on y axis (sub and main) and the rows all look the same. Also why JPEG and not PDF? Doesn't render well} \jh{fixed!}
    Pairwise win rates between Full Context (FC) prompting and three reduced-context configurations across four models (Qwen3-4B, DeepSeek-R1-Distill-Llama-8B, GPT-OSS-20B, and GPT-5.2), evaluated on the Quality and On-Topic dimensions. \begin{revision}\textbf{Top row:} FC vs.\ summarized assistant history (SUM), where each prior assistant response is replaced with a one-sentence summary.\end{revision} \textbf{Second row:} FC vs.\ Last Turn Only (LT), which retains only the most recent user--assistant round. \textbf{Third row:} FC vs.\ Assistant Omitted (AO), which retains all user turns but replaces assistant responses with a placeholder. \textbf{Bottom row:} FC vs.\ Delete Everything (DE), which provides only the current user prompt. Error bars indicate binomial proportion 95\% confidence intervals.}
    \label{fig:colm-winrates-original}
\end{figure}

Across all models, response quality is often maintained under a sparse subset of the conversation history. \rev{\Cref{fig:colm-winrates-original} presents these results, and \cref{app:quantifying-comparable} gives a more quantitative comparison of context omission strategies.}
\begin{revision}
Out of the strategies we tested, storing \textit{one-sentence summaries} of assistant turns (\cref{fig:colm-winrates-original}, top row) is the most effective, often exceeding the performance of storing full responses (FC). We find that when responses from prior turns are retained in full, this can lead models to rehash their earlier responses, rather than addressing the current prompt. Replacing those turns with summaries yields shorter, more focused responses, reducing median response length by roughly 25\% across the four models. \cref{subsec:summarized-vs-fc-responses} gives two examples of this behavior. 
\end{revision}

Storing the \textit{last turn only} (LT) (\cref{fig:colm-winrates-original}, second row) is often sufficient for maintaining performance, on both the Quality and On-Topic dimensions, compared to storing the full conversation (blue bars). We find that a substantial fraction of user turns (40\%) reference \textit{at most} the user--assistant exchange \textit{directly preceding} the current turn (\cref{fig:referenced-turns} (left)). See \cref{subsec:referenced-turn-prompt} for more details on this finding.

Storing the \textit{user turns only} (AO) (\cref{fig:colm-winrates-original}, third row), results in modest Quality degradation but comparable On-Topic win rates. In ablations where we (i) condition on assistant responses from the generator model\footnote{Presently, we use the historical assistant responses from the dataset.} (\Cref{fig:colm-winrates-regenerated}) or (ii) present the judge model with only the user turns to make its judgements (\Cref{fig:colm-winrates-ao-judge}), we find that the AO context often performs \textit{comparably} to full context. Unsurprisingly, removing the entire conversation history, delete-everything (DE), results in substantial performance degradation (\cref{fig:colm-winrates-original} (bottom row)).

Notably, we find that many user requests in the middle of multi-turn chats (36.4\% across our dataset) can be addressed based on the \textit{current prompt alone}; see more details of this finding in \Cref{subsec:prompt-type-analysis}. And for prompts that are explicit follow-ups to previous conversation turns, many prompts provide concrete-enough instructions to be answered \textit{from scratch} based on the past user turns alone (see \Cref{tab:canonical-follow-up-prompts}).\footnote{We manually inspect $50$ full-length conversations across WildChat and ShareLM to reach this conclusion.}

% \FloatBarrier
\begin{figure}[t]
\centering
\begin{minipage}[t]{0.31\textwidth}\raggedright
\exampletitle{Organizing Storage Drives.}
\exampledesc{User asks the assistant for advice on how to set up storage drives.}

\turnlabel{[User Turn 1]}
\begin{tcolorbox}[userbubble]
Provide a design for a disk topology for a NAS built on TrueNAS Scale, as well as a dataset layout [user-inserted specifications]
\end{tcolorbox}

\turnlabel{[User Turn 2]}
\begin{tcolorbox}[userbubble]
1.\ Let's not get this granular. 2.\ Far less than 0.5TB/yr. 3.\ Some irreplaceable data [\ldots] 4.\ Primarily existing hardware. 5.\ Main goal: prevent data loss.
\end{tcolorbox}

% \categorylabel{Follow-up w/ Feedback}
\end{minipage}%
\hfill
\begin{minipage}[t]{0.31\textwidth}\raggedright
\exampletitle{Adding In-text Citations.}
\exampledesc{User shares a written passage and asks the assistant to add citations.}

\turnlabel{[User Turn 6]}
\begin{tcolorbox}[userbubble]
[user-provided text] Please do in-text citations and give me references for the above information.
\end{tcolorbox}

\turnlabel{[User Turn 7]}
\begin{tcolorbox}[userbubble]
The second one is not working.
\end{tcolorbox}

\turnlabel{[User Turn 8]}
\begin{tcolorbox}[userbubble]
This reference is also not good.
\end{tcolorbox}

% \categorylabel{Follow-up w/o Feedback}
\end{minipage}%
\hfill
\begin{minipage}[t]{0.31\textwidth}\raggedright
\exampletitle{Analyzing a Reading Passage.}
\exampledesc{User shares a reading passage and asks the assistant questions about the passage.}

\turnlabel{[User Turn 1]}
\begin{tcolorbox}[userbubble]
What's your take on the following text? [user-inserted reading passage]
\end{tcolorbox}

\turnlabel{[User Turn 3]}
\begin{tcolorbox}[userbubble]
What influences or inspirations do you notice in the text?
\end{tcolorbox}

% \categorylabel{Follow-up w/o Feedback}
\end{minipage}
\caption{Three example conversations on WildChat illustrating different types of follow-ups. \textbf{Left:} the user provides feedback that is concrete enough that the model can respond from scratch using the previous user turn and the updated specifications alone. \textbf{Middle:} the follow-ups reference specific parts in past \textit{assistant} turns, making it necessary to see the referenced assistant turn. \textbf{Right:} the follow-up references a prior \emph{user} turn without concrete feedback; no assistant history is necessary.}
\label{tab:canonical-follow-up-prompts}
\end{figure}
% \FloatBarrier

In the left and right conversations (\cref{tab:canonical-follow-up-prompts}), the follow-up prompt either contains concrete-enough instruction for the model to generate a response \textit{from scratch} without seeing any prior assistant turns (left) 
or references only the prior \textit{user} turns (right). Thus, a model can plausibly respond without seeing any of the assistant-side history. 
\cref{tab:canonical-follow-up-prompts} (middle) presents a case where a past assistant response is referenced and thus necessary for answering the follow-up prompt. Even in this case, however, storing only the referenced response is sufficient. These findings challenge the assumption that uniformly storing assistant-side history is necessary. Furthermore, we identify three prominent conversation structures in our datasets: (1) an initial primary prompt (e.g., a reading passage) followed by queries about that prompt; (2) a sequence of loosely related, standalone queries; and (3) a cohesive thread characterized by iterative user feedback (e.g., fixing a buggy piece of code).\footnote{These categories were identified through a manual reading of 50 random conversations on WildChat and ShareLM.} Each conversation structure entails a different degree of reliance on storing past assistant responses; for representative conversations belonging to each category, see \cref{subsec:example-conversation-categories}.

\label{colm-subsec:context-savings}
\paragraph{Alternative context configurations offer substantial compute savings.} 
Despite achieving comparable response quality, the reduced-context configurations \textit{significantly} decrease the amount of compute spent across multi-turn conversations. \Cref{fig:colm-context-lengths} (center) plots the cumulative context length in characters as a function of conversation length.\footnote{These plots are created based on GPT-5.2 generations.} While the Full Context increases linearly as the conversation expands, the AO, LT, \rev{and SUM} contexts remain relatively constant. \Cref{fig:colm-context-lengths} (right) shows the number of conversations that are answerable within a $5$k-token budget as a function of conversation length. For a chat of length 8 rounds, the AO, LT, \rev{and SUM} contexts can successfully answer 4–6 conversations on average, whereas FC can answer only one.

\begin{savenotes}
\begin{figure}[t]
    \centering
    \includegraphics[width=1.00\linewidth]{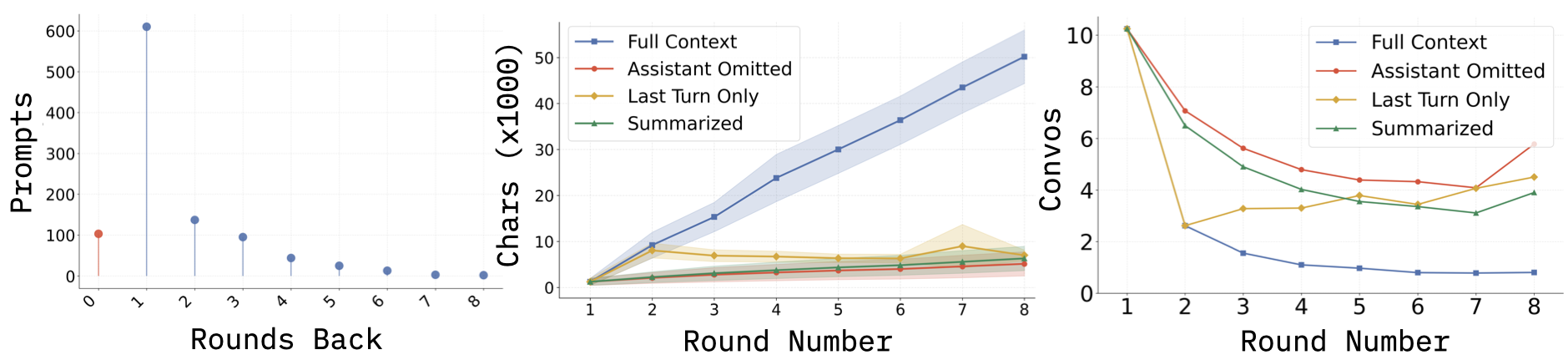}
    \caption{\textbf{Left:} Distribution of conversation rounds referenced by user prompts from round 2 onward, across 150 WildChat conversations. The X-axis represents the number of rounds back from the current round that a given prompt references. The red stem at 0 indicates prompts that are self-contained (no prior rounds referenced). \textbf{Middle:} Mean cumulative context length across conversations on WildChat as a function of conversation round, comparing Full Context, Assistant Omitted, Last Turn Only\rev{, and Summarized} context. Shaded band shows \(\pm\)2 standard errors. \textbf{Right:} The number of conversations that can be answered using a 5K-token budget as a function of conversation round.}
    \label{fig:colm-context-lengths}
    \label{fig:referenced-turns}
\end{figure}
\end{savenotes}

\section{LLMs Tend to Over-Condition on Their Past Responses}
\label{sec:too-much-information-is-overwhelming}
There is no shortage of anecdotal evidence online suggesting that LLM response quality degrades over the course of multi-turn chats. Users report that responses tend to get ``less precise, more repetitive, and sometimes subtly wrong''\footnote{\url{https://www.reddit.com/r/ChatGPT/comments/1qngpqa/does_chatgpt_quietly_get_worse_in_long/}} as the chat progresses. Others note that conversations gradually ``losing a clear direction,'' with assistants responses feeling like ``a mix of adjustments instead of one clear goal.''\footnote{\url{https://www.reddit.com/r/ChatGPT/comments/1ppbbho/llm_gets_dumber_response_quality_degrades_over/}} Some even reach the conclusion: ``If you want the maximum quality in your responses, you only can get 3–5 responses before you need to start over.''\footnotemark[7] We indeed find instances on WildChat where the user abandons a conversation in favor of a new chat (see \cref{app:ex-cascading-patches}).
% Another user explains: ``once your context is `poisoned' it will not recover, [so] you need to start fresh with a new chat.''\footnotemark

In this section, we investigate multi-turn performance degradation through the concept of \textit{context pollution}:\footnote{\url{https://www.anthropic.com/engineering/effective-context-engineering-for-ai-agents}} the tendency for models to over-condition on their prior responses, leading to errors, hallucinations, and stylistic artifacts that cascade across turns. To avoid quality decline in multi-turn conversations, future work could move toward \textit{multi-turn} reinforcement learning (RL). We hope that the context pollution examples we present (see \Cref{subsec:context-pollution-examples}) will be useful for designing future annotation rubrics for human annotators and LLM-judges to assign quality scores to conversation turns. 

Next, we show that multi-turn math performance can degrade when past responses are maintained in context. Then, we provide a characterization of different forms of context pollution in multi-turn conversations. Finally, we discuss a promising design for reducing context pollution.

\subsection{Multi-Turn Math Performance Degrades When Past Responses Are Retained}
\label{subsec:gsm8k}

\citet{laban2025lost} investigated the impact of transforming single-turn math word problems from GSM8K into multi-turn, underspecified dialogues where problem details are disclosed incrementally through a sharding process. Their study revealed a ubiquitous and substantial performance drop across 15 state-of-the-art LLMs when transitioning from the single to multi-turn setting. The researchers primarily attributed this degradation to the models' tendency to make premature assumptions regarding missing information and attempt final solutions before the user had revealed all necessary shards of the instruction.

\begin{figure}[t]
    \centering
    \includegraphics[width=0.85\linewidth]{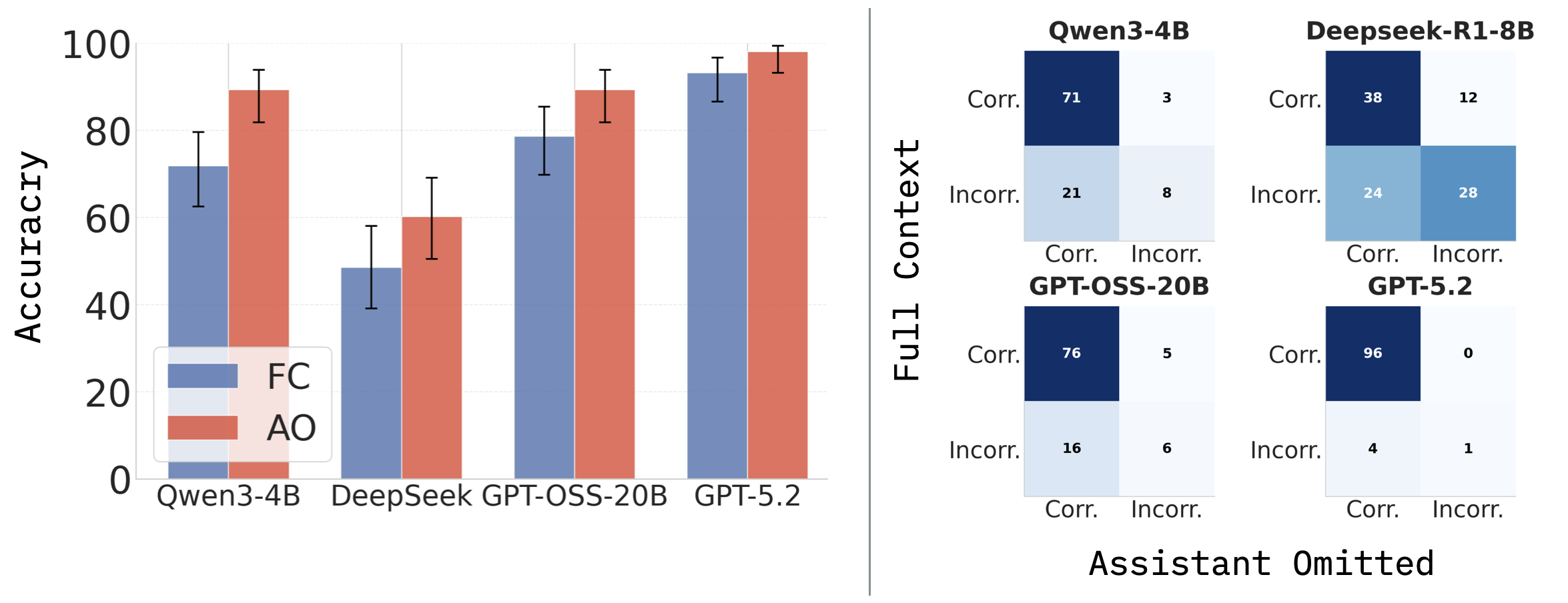}
    \caption{\textbf{Left:} Accuracy (\%) on multi-turn GSM8K under Full Context (FC) and Assistant Omitted (AO) context. \textbf{Right:} Confusion matrices showing the number of problems answered correctly across models and context configurations.}
    \label{fig:gsm8k}
\end{figure}

Building on these insights, we explore whether multi-turn performance degradation is recoverable by simply removing past assistant responses from the context. As illustrated in \Cref{fig:gsm8k}, we find that while the FC responses align with the performance reported in \citet[Table 1]{laban2025lost}, performance partially recovers when past model responses are omitted.

When we inspect instances where models recovered under AO context, we find that models often become distracted or confused by their earlier responses. For instance, a detail invented in an earlier turn may cascade across several conversation turns, ultimately leading to an incorrect answer (see \cref{app:ex-johns-planks}); a flawed reasoning trace that is carried across several rounds can lead the model to incorrectly conclude that a ``problem is impossible'' (see \cref{app:ex-jerry-pools}); an earlier response can lead the model to misinterpret the intended task and recompute the problem from scratch, rather than accumulating the solution across turns (see \cref{app:tea-party}).

\subsection{Storing Past Assistant Responses Creates Context Pollution}
\label{subsec:context-pollution}
In this section, we identify and characterize the ways in which LLMs over-condition on their past responses. To measure the prevalence of context pollution, we use Claude Haiku 4.5 to classify $150$ conversations on WildChat into one of five categories (see \Cref{tab:context_pollution_prevalence}) or as no pollution (see \cref{subsec:context-pollution-annotator-prompt} for the full LLM annotation prompt).\footnote{The authors manually reviewed a subset of the flagged instances to verify that they were genuine cases of context pollution. Select examples are presented in \Cref{subsec:context-pollution-annotator-prompt}.} We find that context pollution occurs in roughly 14\% of the conversation turns in our dataset.\footnote{This experiment was conducted using the frontier model, GPT-5.2.}

The most common form of context pollution we identified is \textit{intent override} (\Cref{tab:context_pollution_prevalence}). In this setting, the model persists with the content or style of an earlier response, despite a user’s attempt to correct or provide new details and specifications. For instance, when a user asks Qwen3-4B to ``reflect’’ on a piece of code, the model instead generates another \textit{version} of the code, rather than providing a \textit{reflection} on the code (see \cref{app:ex-stylistic-inertia}). In another instance, when a user introduces a new location to a storyline, the model incorrectly confuses it with a previous location that was present in context.

The remaining examples we present include: parameters from earlier code (e.g., UMAP-specific arguments) incorrectly leaking into later turns (t-SNE; \cref{app:ex-umap-tsne}); a hallucinated fact about a novel being carried across turns (\cref{app:ex-error-propagation}); a specific parameterization of model temperature (\cref{subsec:ex-temperature}) being misinterpreted in a later turn; and a pathological reasoning pattern being amplified across turns (\cref{app:ex-reasoning-spillover}).

\begin{tcolorbox}[contextpollutionbox, title={Types of Context Pollution}, label={box:context-pollution}]
\textbf{Intent Override: } The assistant follows a narrative, assumption, or task structure from an earlier response instead of adapting to the current user instruction, leading it to miss or ignore details in the latest instruction.

\textbf{Code Carryover:} Code from an earlier response is reused, even though it is incorrect, or syntax from a previously used programming language appears after the user has requested a different language.

\textbf{Factual Hallucination Carryover:} A false fact introduced in an earlier response is repeated or treated as true in a later response.

\textbf{Reasoning Trace Carryover:} Verbose reasoning patterns or stylistic artifacts from an earlier response are repeated or amplified in a later response.

\textbf{Formula Carryover:} A mathematical formula introduced in an earlier response is reused incorrectly in a later response.
\end{tcolorbox}
% \FloatBarrier
% \begin{table}[t]
% \centering
% \small
% \begin{tabular}{ll@{\hskip 2em}ll}
% \toprule
% \multicolumn{2}{l}{\textbf{Overall Prevalence}} & \multicolumn{2}{l}{\textbf{Breakdown by Category}} \\
% \midrule
% Turn-level   & 104 / 729 turns (14.3\%) & Intent Override              & 39 \\
% Conversation-level & 59 / 150 convos (39.3\%) & Code Carryover     & 30 \\
%              &                          & Factual Hallucination Carryover & 27 \\
%              &                          & Reasoning Trace Carryover    & 6 \\
%              &                          & Formula Carryover            & 1 \\
% \bottomrule
% \end{tabular}
% \caption{Prevalence of context pollution across 729 responses generated by GPT-5.2 in conversations on WildChat. \textbf{Left:} the overall rate of context pollution at the turn and conversation levels. \textbf{Right:} the break down of instances of context pollution by category.}
% \label{tab:context_pollution_prevalence}
% \end{table}
% \FloatBarrier
\begin{revision}
\label{subsec:neutral-filler}
\paragraph{Disentangling Context Pollution From Generic Long-Context Degradation.} To test whether context pollution stems from past assistant responses specifically or from \textit{any} additional text in context, we replace the assistant-side context with length-matched neutral filler.

Starting from the GPT-5.2 pollution annotations in \cref{tab:context_pollution_prevalence}, we analyze turns that were flagged for the three most prevalent categories: Intent Override, Code Carryover, and Factual Hallucination Carryover. For each flagged turn, we replace every prior assistant turn in history with length-matched filler text while leaving the user turns unchanged. We then regenerate GPT-5.2's response under the modified filler-text history and run the same pollution judge (see \cref{subsec:context-pollution-annotator-prompt}) on the new response. For Intent Override and Factual Hallucination Carryover, the filler text we use is drawn from random Wikipedia passages, \texttt{DragonLLM/Clean-Wikipedia-English-Articles} \citep{foly2025cleanwikipedia}; for Code Carryover, the filler text we use is drawn from random Python code snippets, \texttt{bigcode/the-stack} \citep{kocetkov2023stack}.

Across all tested pollution categories, we observe that every single originally-flagged instance of context pollution flipped from being present to absent under the new filler text. This reveals that context pollution is driven more by the semantic content of past assistant responses (it being a sensical prior assistant response), rather than by general long-context degradation. In fact, an interesting response we came across when spot-checking the length-matched neutral filler generations was one case where GPT-5.2 generated: ``Also, something clearly went wrong in my earlier replies (they were unrelated garbage text). Sorry about that.'' The fact that GPT-5.2 claimed ``ownership'' over the prior assistant-side responses suggests that the model may treat past assistant responses as privileged and may plausibly choose to rely on them as a good starting point to inform downstream generation.
% \begin{table}[t]
% \centering
% \small
% \begin{tabular}{l@{\hskip 2em}ll}
% \toprule
% \textbf{Category} & \textbf{Assistant-response History} & \textbf{Filler History} \\
% \midrule
% Intent Override                  & 39 present (conf.\ 0.74) & 0 present (conf.\ 0.96) \\
% Factual Hallucination Carryover  & 27 present (conf.\ 0.75) & 0 present (conf.\ 0.95) \\
% Incorrect Code Carryover         & 30 present (conf.\ 0.77) & 0 present (conf.\ 0.95) \\
% \bottomrule
% \end{tabular}
% \caption{\begin{revision}Context pollution under the original conversation history and one where assistant-side responses are replaced with length-matched filler text. \textbf{Assistant-response history:} counts when the history contains prior assistant responses. \textbf{Filler history:} counts when the history is replaced with filler content. Confidence values indicate the LLM annotator's average confidence across pollution instances in the category.\end{revision}}
% \label{tab:context_pollution_ablation}
% \end{table}
% \end{revision}
\begin{table}[t]
\centering
\small
\begin{tabular}{l@{\hskip 1.5em}rc@{\hskip 2em}rc}
\toprule
& \multicolumn{2}{c}{\textbf{Assistant-response History}} & \multicolumn{2}{c}{\textbf{Filler History}} \\
\cmidrule(lr){2-3}\cmidrule(lr){4-5}
\textbf{Category} & Count & Conf. & Count & Conf. \\
\midrule
Intent Override                 & 39 & 0.74 & 0 & 0.96 \\
Incorrect Code Carryover        & 30 & 0.77 & 0 & 0.95 \\
Factual Hallucination Carryover & 27 & 0.75 & 0 & 0.95 \\
Reasoning Trace Carryover       &  6 & --   & -- & --   \\
Formula Carryover               &  1 & --   & -- & --   \\
\midrule
\multicolumn{5}{l}{\textbf{Overall prevalence} (assistant-response history)} \\
\multicolumn{2}{l}{\quad Turn-level} & \multicolumn{3}{l}{104 / 729 turns (14.3\%)} \\
\multicolumn{2}{l}{\quad Conversation-level} & \multicolumn{3}{l}{59 / 150 convos (39.3\%)} \\
\bottomrule
\end{tabular}
\caption{\revision{Prevalence of context pollution instances across 729 responses generated by GPT-5.2 on WildChat conversations, under the original history and a history in which assistant-side responses are replaced with length-matched filler text. Confidence values indicate the LLM annotator's average confidence across instances in that category.}}
\label{tab:context_pollution_prevalence}
\end{table}
\end{revision}
\label{subsec:keep-context-clean}
\paragraph{Keeping The Context Window Clean.} To reduce context pollution, conversational agents could move toward \textit{summarization-and-retrieval} frameworks. \rev{In \cref{fig:colm-winrates-original} (top row), we find that storing one-sentence summaries often outperforms storing the full context.} Rather than retaining the full conversation history, systems could maintain lightweight identifiers of past turns (e.g., short summaries or prefixes) and use these as retrieval keys, loading the relevant content only when a follow-up requires it \citep{anthropic2025contextengineering,zhang2025recursive,wang2023learning,packer2023memgpt}. Such a summarize-and-retrieve framework may more closely align with the way humans engage in dialogue: each turn updates a conversational state \citep{lewis1979scorekeeping}, rather than replaying the entire raw transcript before responding each time.

% While recent progress has emphasized scaling context windows \citep{team2024gemini,anthropic2026opus46}, a more efficient alternative for improving long-context performance is by maintaining focused, high-quality information in the active context.
% , where full model responses are stored as summaries, acting as a retrieval key for the full response, retrieved only when necessary \citep{zhang2025recursive,wang2023learning,packer2023memgpt}. 

%Recent agent systems have begun to explore the design of allowing models to write information to memory.\footnote{\url{https://platform.claude.com/docs/en/agents-and-tools/tool-use/memory-tool}} Current memory tools, however, focus primarily on storing \textit{tool results}.

\section{Adaptive Assistant Response Omission}
\label{sec:adaptive-context-inclusion}
\begin{revision}
In this section, we introduce a context filter that decides, on a per-turn basis, whether to retain (FC) or omit (AO) the assistant-side history, trading off response quality and token cost. We run this experiment using GPT-5.2, in the setting where the model's own responses are fed back into context.

Given a user prompt and the conversation history leading up to that prompt, we train a binary classifier to predict the LLM-judge's winner on the quality dimension. We label turns where the LLM-judge prefers the full-context response as one, and all other turns as zero.\footnote{We label ties as zero, reflecting cases where retaining assistant-side history confers no quality benefit despite the added compute cost.} The features we use include round number, cumulative context lengths, prompt category, and dense embeddings of the user prompt and conversation history from a pretrained text embedding model,\footnote{We use OpenAI's \texttt{text-embedding-3-large} to align with the GPT-5 model family we use for response generation.} reduced to twenty dimensions each using PCA. 

We combine the classifier's predicted probability $\hat{P}(\mathrm{FC} \succ \mathrm{AO})$ with a penalty on token cost. Let $l_i$ denote the number of tokens in the assistant-side history up to turn $i$, and let $\bar{l}$ denote the mean of $l_i$ over all turns in the dataset. We define the \emph{relative history length}, $\hat{\Delta}_i = l_i / \bar{l}$, and score each turn as $\hat{\mathrm{S}}_i = \hat{P}(\mathrm{FC} \succ \mathrm{AO}) - \lambda \cdot \hat{\Delta}_i$, selecting full context if and only if $\hat{\mathrm{S}}_i \geq \tau$.

Our adaptive approach (see \cref{fig:adaptive-approach-cost-aware}) is able to retain over 99\% of the performance of selecting full context for all turns (the red square) while using only an average of 87\% of the total token cost. Applying the classifier trained on GPT-5.2 outputs to other LLM backbones in our study, we find that the context filter transfers nicely across different model backbones (see \cref{subsec:policy-transfer}). This generalization suggests that whether assistant history should be retained depends more on properties of the conversation itself than on model-specific behavior.
%We evaluate against simple heuristic that omits assistant responses on all user turns that are annotated as being ``New Asks;'' see \cref{subsec:llm-annotation-prompts} for the LLM-annotation prompt used to perform this classification. The omit-on-new-ask heuristic, which spends around half of the full-context token budget, is beaten by a cost-aware operating point that spends around half of the omit-on-new-ask budget. 

\begin{figure}[t]
    \centering
    \includegraphics[width=0.48\linewidth]{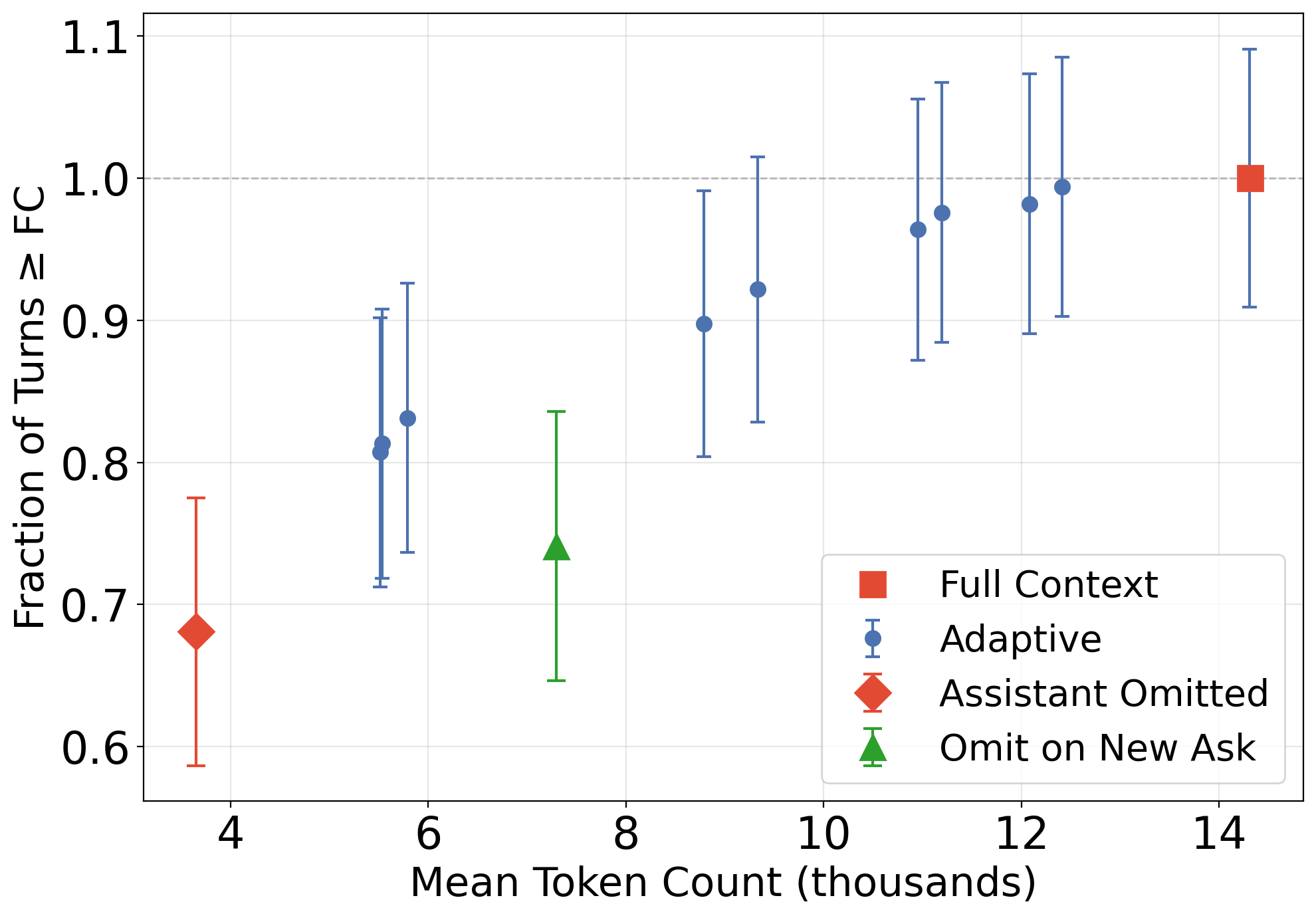}
    \caption{\begin{revision}Ratio of adaptive wins to full-context-only wins vs.\ mean token
    count (in thousands) for operating points obtained by sweeping the decision rule's
    threshold $\tau$ and cost weight $\lambda$. The green
    triangle marks a heuristic that omits assistant responses on all turns categorized as ``New Asks.''\end{revision}}
    \label{fig:adaptive-approach-cost-aware}
\end{figure}
\end{revision}

\section{Discussion}
\label{sec:discussion}
In this work, we analyze real-world multi-turn conversations and uncover a surprising finding: \textit{conditioning on full conversation histories is largely excessive.} Across all models in our study, we find that reduced-context configurations often perform comparably to conditioning on the full history, while substantially saving compute. We find that dependencies in real-world multi-turn conversations may be shorter in horizon than commonly assumed. In many cases, we also find that follow-up queries contain sufficient instruction to allow models to respond \textit{from scratch}, without seeing their own prior responses. Beyond token savings, we find that storing shorter contexts also reduces distractions. In multi-turn math, removing the assistant-side history can partially \emph{recover} performance, suggesting that premature model responses may interfere with downstream response quality.

We note that our evaluation relies on an LLM-as-judge framework. While the GPT-5 judge achieves $\geq 90\%$ alignment with human annotations (\cref{app:lm-judge-alignment-and-justifications}), future work should extend this analysis to a larger-scale human evaluation. 

Our findings point to two distinct areas for future research. First, context management systems may benefit from making \textit{content relevancy}-based filtering decisions, rather than relying on compaction after exceeding a certain length threshold. Second, \textit{multi-turn} reinforcement learning may allow models to avoid some of the behavioral pathologies that we recover in \cref{sec:too-much-information-is-overwhelming}. Overall, we hope that these findings inspire a shift away from conditioning on full dialogue transcripts toward curating cleaner, more focused contexts.

\section*{Acknowledgments}

This work was supported by the MIT--IBM Computing Research Lab and the Amazon AI Research Innovation Fellowship. We are grateful to Mitchell Gordon and Koren Lazar for helpful feedback and discussions.

% there should be a shift away from scaling and a shift towards keeping the memory clean and focused.
% Third, these findings highlight the need for benchmarks that more explicitly capture \emph{true} multi-turn dependence, as existing datasets may overestimate the necessity of long-context conditioning.

% Paradigm shift: the model should never look at more information than is needed. we can’t just store plain transcripts in the context.
% (i)	keep the history selective and short. In figure 1, we see that, for multi-turn chats, it is sufficient only to keep around the turns that are referred to. This will help avoid overwhelming the llm, anchoring it to the past, resulting in context pollution. Humans have a shorter memory, and there is evidence this helps us to generalize.
% (ii) insert opportunities for reflection. right now, the model is just going at full speed, spitting out next token according to its previous tokens (opportunities to reflect on the state of the conversation, make judgements about relevancy, not overreact to mistakes and start questioning itself).

% \bibliography{colm2026_conference}
\bibliography{references}
\bibliographystyle{colm2026_conference}

\newpage
\appendix
\section{Appendix}
\subsection{Data}
\label{subsec:data}
We analyze two real-world multi-turn conversation datasets and one synthetic multi-turn conversation dataset.

\paragraph{WildChat.}
The \texttt{allenai/WildChat-4.8M} dataset \citep{zhao2024wildchat} consists of around 4.8 million real-world, multi-turn conversations sourced from ChatGPT interactions.
\paragraph{ShareLM.}
The \texttt{shachardon/ShareLM} dataset \citep{don2025sharelm} is a unified collection of human--model interaction data aggregated from multiple conversation sources; see the unified contributions section of: \url{https://huggingface.co/datasets/shachardon/ShareLM}.
%, including the ShareLM browser plugin, Collective Cognition, Anthropic's HH-RLHF, bAbI, Self-Feeding, PRISM, PIPPA, the ChatGPT Roleplay Dataset, You Tell Me, and WildChat.

\paragraph{Lost-in-Conversation.}
The \texttt{microsoft/lost\_in\_conversation} \citep{laban2025lost} dataset is a synthetic dataset designed to study the way LLMs respond to incrementally specified instructions. It consists of prompts from single-turn benchmarks (HumanEval, LiveCodeBench, Spider, Berkeley Function Calling Leaderboard, GSM8K, ToTTo, and Summary of a Haystack) into multi-turn dialogues via a sharding procedure. Each sharded instruction is composed of multiple shards that incrementally reveal the original query.

\paragraph{Data Filtering.} For the real-world datasets, we load the first subset of the data: 5{,}000 conversations from WildChat and 50{,}000 from ShareLM. For WildChat, we filter for conversations originally conducted with GPT-4 models, the most recent models represented in the dataset. For ShareLM, we filter for conversations conducted with frontier models, GPTs, Gemini, and Claude, as identified by the conversation URL metadata (\texttt{chatgpt.com}, \texttt{gemini.google.com}, and \texttt{claude.ai}, respectively). For both datasets, we filter for conversations in English (when language metadata is available),\footnote{Upon manual inspection, the final dataset still contains a small subset of conversations in other languages.} conversations with 5--10 rounds (where a round corresponds to one user–assistant exchange), and conversations spanning diverse topic categories, identified by the presence of any one of the following keywords:
\begin{itemize}
    \item \textbf{Creative writing:} \textit{essay}, \textit{poem}, \textit{novel}, \textit{protagonist}, \textit{antagonist}, \textit{villain}, \textit{hero}, \textit{plot}, \textit{monologue}, \textit{fairy tale}, \textit{science fiction}, \textit{mystery}, \textit{thriller}, \textit{dystopian}, \textit{historical fiction}, \textit{metaphor}, \textit{imagery}, \textit{lyric}, \textit{satire}
    \item \textbf{Science \& medical:} \textit{biology}, \textit{organism}, \textit{species}, \textit{virus}, \textit{bacteria}, \textit{ecology}, \textit{disease}, \textit{diagnosis}, \textit{treatment}, \textit{symptom}, \textit{pathology}, \textit{clinical}, \textit{patient}, \textit{chemistry}, \textit{physics}, \textit{molecule}, \textit{cell}, \textit{protein}, \textit{DNA}, \textit{neurology}, \textit{pharmacology}, \textit{epidemiology}, \textit{scientific evidence}, \textit{hypothesis}, \textit{experiment}, \textit{laboratory}, \textit{microscope}, \textit{anatomy}, \textit{physiology}, \textit{medical}, \textit{medicine}
    \item \textbf{Business \& finance:} \textit{business}, \textit{finance}, \textit{trading}, \textit{stock}, \textit{economics}, \textit{marketing}, \textit{investment}, \textit{revenue}, \textit{profit}, \textit{pricing}, \textit{market research}, \textit{entrepreneur}, \textit{supply chain}, \textit{financial}, \textit{accounting}, \textit{budget}, \textit{startup}, \textit{venture capital}, \textit{portfolio}, \textit{interest rate}, \textit{inflation}, \textit{GDP}, \textit{fiscal}, \textit{monetary}, \textit{business plan}, \textit{business model}, \textit{strategy}
    \item \textbf{Education:} \textit{homework}, \textit{assignment}, \textit{exam}, \textit{test preparation}, \textit{study}, \textit{tutor}, \textit{explain to me}, \textit{course}, \textit{lesson}, \textit{tutorial}, \textit{academic}, \textit{student}, \textit{quiz}, \textit{study guide}, \textit{curriculum}, \textit{teach me}, \textit{learning}, \textit{professor}, \textit{lecture}, \textit{semester}, \textit{grade}, \textit{classroom}, \textit{education}, \textit{school}, \textit{university}
    \item \textbf{Math:} \textit{equation}, \textit{algebra}, \textit{calculus}, \textit{derivative}, \textit{matrix}, \textit{probability}, \textit{statistics}, \textit{geometry}, \textit{trigonometry}, \textit{polynomial}, \textit{function}, \textit{calculate}, \textit{theorem}, \textit{proof}, \textit{lemma}, \textit{corollary}, \textit{formula}, \textit{differential}, \textit{linear algebra}, \textit{graph theory}, \textit{optimization}, \textit{mathematics}, \textit{mathematical}
    \item \textbf{Coding:} \textit{code}, \textit{python}, \textit{javascript}, \textit{java}, \textit{c++}, \textit{programming}, \textit{class}, \textit{method}, \textit{algorithm}, \textit{debug}, \textit{error}, \textit{compile}, \textit{variable}, \textit{array}, \textit{loop}, \textit{recursion}, \textit{data structure}, \textit{API}, \textit{library}, \textit{framework}, \textit{script}, \textit{implementation}, \textit{runtime}, \textit{bug}, \textit{software}, \textit{developer}, \textit{coding}, \textit{program}
\end{itemize}
\noindent Because filtering is keyword-based, any conversation containing one of the specified keywords is retained. Finally, we filter out conversations identified as toxic by the \texttt{detoxify} ML-based toxicity classifier (with a threshold set to 0.7).

Due to compute constraints (e.g., generating model responses for every round of every conversation across four models), we randomly sample 200 conversations from WildChat and 150 from ShareLM, yielding a total of 350 real-world conversations for our analysis.

For Lost-in-Conversation, we use the 103 math word problems from GSM8K. 

\subsection{Models}
\label{subsec:models}
We evaluate three open-source, reasoning models spanning a range of sizes and backbones, \textbf{Qwen3-4B}, \textbf{DeepSeek-R1-Distill-Llama-8B}, \textbf{GPT-OSS-20B}, and one state-of-the-art model, \textbf{GPT-5.2}.
Each model is served with its recommended sampling parameters:
\begin{itemize}
    \item \textbf{DeepSeek-R1-Distill-Llama-8B}: temperature 0.6, top-$p$ 0.95, max generation length 32{,}768 tokens, context window 32{,}768 tokens~\citep{deepseekr1distillllama8b}. This model is served via \texttt{vLLM}.
    \item \textbf{Qwen3-4B}: temperature 0.6, top-$p$ 0.95, top-$k$ 20, thinking mode enabled,  max generation length 1{,}024 tokens, context window 32{,}768 tokens~\citep{qwen34bmodelcard}. This model is served via \texttt{vLLM}.
    \item \textbf{GPT-OSS-20B}: temperature 1.0, top-$p$ 0.95; max generation length 2{,}048 tokens, context window 32{,}768 tokens~\citep{agarwal2025gpt}. This model is served via HuggingFace Transformers.
    \item \textbf{GPT-5.2}: We use the default API parameters~\citep{openaiUsingGPT52}, with reasoning effort set to \texttt{medium}. This model is served via the OpenAI Responses API.
\end{itemize}

\subsection{Context Configurations}
\label{subsec:conditions}
For each conversation, we generate responses under standard full-context prompting and three alternative context configurations:

\paragraph{Full Context (FC).} The model sees all prior user and assistant turns.

\paragraph{Assistant-Omitted (AO).}
The model sees all prior user turns while assistant responses are replaced with a short placeholder in the conversation history. Additionally, the model receives a system prompt:

\begin{quote}
\textit{``In this conversation, previous assistant responses are shown as `[Response provided]' to save memory. These are placeholders indicating the turn was already answered and has passed. Focus only on answering the user's most recent message.''}
\end{quote}

\noindent After each turn, the assistant's full response is replaced with the placeholder \texttt{[Response provided]} to maintain the alternating user/assistant turn structure.

\paragraph{Last-Turn-Only (LT).}
The model receives only the immediately preceding user--assistant exchange plus the current user prompt. All earlier turns are discarded.
\begin{revision}
\paragraph{Summarized (SUM).}
The model sees all prior user turns, while assistant responses are replaced with one-sentence summaries. After each turn, the summary is generated by prompting the \textit{same} model being evaluated with:

\begin{quote}
\textit{``Summarize the following assistant response in exactly one sentence:\ \{response\}}

\textit{One-sentence summary:''}
\end{quote}

\noindent This summary is what gets appended to the conversation history in place of the original response.
\end{revision}
\paragraph{Delete-Everything (DE).}
The model receives only the current user prompt with no prior conversation history. All previous user and assistant turns are discarded.

\Cref{fig:chat-templates} shows the \texttt{conversation\_history} message list passed to each model's chat template for the four configurations. The example depicts a 3-turn conversation at the point of generating a response to turn~3.
For the three open-weight models (DeepSeek-R1, Qwen3, and GPT-OSS), this message list is converted to a prompt string using HuggingFace Transformers' \texttt{tokenizer.apply\_chat\_template(conversation\_history, tokenize=False, add\_generation\_prompt=True)}. For GPT-5, the message list is passed directly to OpenAI's Responses API.

\FloatBarrier
\begin{figure}[h]
\centering
\small
\begin{minipage}[t]{0.48\textwidth}
\textbf{Full Context (FC)}
\begin{verbatim}
[
 {"role": "user",
  "content": "<user turn 1>"},
 {"role": "assistant",
  "content": "<asst response 1>"},
 {"role": "user",
  "content": "<user turn 2>"},
 {"role": "assistant",
  "content": "<asst response 2>"},
 {"role": "user",
  "content": "<user turn 3>"}
]
\end{verbatim}
\end{minipage}%
\hfill
\begin{minipage}[t]{0.48\textwidth}
\textbf{Assistant-Omitted (AO)}
\begin{verbatim}
[
 {"role": "system",
  "content": "In this conversation,
   previous assistant responses are
   shown as '[Response provided]'
   to save memory. ..."},
 {"role": "user",
  "content": "<user turn 1>"},
 {"role": "assistant",
  "content": "[Response provided]"},
 {"role": "user",
  "content": "<user turn 2>"},
 {"role": "assistant",
  "content": "[Response provided]"},
 {"role": "user",
  "content": "<user turn 3>"}
]
\end{verbatim}
\end{minipage}

\vspace{0.5em}
\hrule
\vspace{0.5em}

\begin{minipage}[t]{0.48\textwidth}
\textbf{Last-Turn-Only (LT)}
\begin{verbatim}
[
 {"role": "user",
  "content": "<user turn 2>"},
 {"role": "assistant",
  "content": "<asst response 2>"},
 {"role": "user",
  "content": "<user turn 3>"}
]
\end{verbatim}
\end{minipage}%
\hfill
\begin{minipage}[t]{0.48\textwidth}
\textbf{Delete-Everything (DE)}
\begin{verbatim}
[
 {"role": "user",
  "content": "<user turn 3>"}
]
\end{verbatim}
\end{minipage}

\vspace{0.5em}
\hrule
\vspace{0.5em}

\begin{minipage}[t]{0.48\textwidth}
\textbf{Summarized (SUM)}
\begin{verbatim}
[
 {"role": "user",
  "content": "<user turn 1>"},
 {"role": "assistant",
  "content": "<1-sentence summary
   of asst response 1>"},
 {"role": "user",
  "content": "<user turn 2>"},
 {"role": "assistant",
  "content": "<1-sentence summary
   of asst response 2>"},
 {"role": "user",
  "content": "<user turn 3>"}
]
\end{verbatim}
\end{minipage}

\caption{Chat template message lists for each context configuration, shown for a 3-turn conversation generating a response to turn~3.}
\label{fig:chat-templates}
\end{figure}
\FloatBarrier

\subsection{LLM-Judge System Prompts}
\label{subsec:direct-comparison-judge-prompt}
To evaluate responses, we use GPT-5 as an LLM-judge. For every round in a conversation, the judge receives both the FC and alternative-context responses and picks a winner for each dimension (Quality and On-Topic). We randomize response ordering (which response is labeled ``A'' vs.\ ``B'') for each comparison to reduce position bias.

Below is the judge prompt used for the main text evaluation (\cref{fig:colm-winrates-original}), where the original dataset's assistant responses (i.e., the assistant responses from the source dataset, not from the model being evaluated) are used as context.

\begin{quote}
\textbf{System Prompt:}

You are an expert evaluator of AI assistant responses. You will evaluate two responses to round \texttt{\{round\_num\}} of \texttt{\{total\_rounds\}} in a multi-turn conversation.

\medskip
\textbf{CONVERSATION CONTEXT (User Prompts and Assistant Responses):} \\
\texttt{\{context\_prompt\}}

\medskip
\textbf{CURRENT ROUND'S USER PROMPT:} \\
\texttt{\{current\_round\_prompt\}}

\medskip
\textbf{RESPONSE A:} \\
\texttt{\{first\_resp\}}

\medskip
\textbf{RESPONSE B:} \\
\texttt{\{second\_resp\}}

\medskip
\textbf{EVALUATION CRITERIA:} \\
You must compare the two responses on TWO separate dimensions and pick a winner for each:

\begin{enumerate}
    \item \textbf{Overall Quality}: How well does the response address the user's current prompt?
    \begin{itemize}
        \item Completeness in addressing the user's request
        \item Accuracy of information provided
        \item Clarity and coherence of the response
        \item Helpfulness and usefulness to the user
    \end{itemize}
    \item \textbf{On-Topic}: Does the response stay focused on the conversation?
    \begin{itemize}
        \item Stays on-topic and relevant to the conversation thread
        \item Avoids topic drift, tangents, or unrelated content
        \item Avoids repetitive phrases or unnecessary fillers
    \end{itemize}
\end{enumerate}

\medskip
\textbf{TASK:} \\
For EACH dimension, directly compare the two responses and pick a winner. If the responses are roughly equal on a dimension, declare a tie. Provide your evaluation in the following JSON format:

\begin{verbatim}
{
  "quality_winner": "A" or "B" or "tie",
  "quality_justification": "<explanation comparing both
    responses on quality>",
  "ontopic_winner": "A" or "B" or "tie",
  "ontopic_justification": "<explanation comparing both
    responses on on-topic relevance>",
  "confidence": <float from 0.0 to 1.0 indicating
    how confident you are in your judgments>
}
\end{verbatim}

Be objective and thorough in your evaluation. Respond ONLY with valid JSON.
\end{quote}

Below, we show a variant of the judge prompt where the assistant responses in context \textit{come from the model currently generating the response}.

\begin{quote}
\textbf{System Prompt:}

You are an expert evaluator of AI assistant responses. You will evaluate two responses to round \texttt{\{round\_num\}} of \texttt{\{total\_rounds\}} in a multi-turn conversation.

\medskip
\textbf{CONVERSATION CONTEXT FOR RESPONSE A (All Past User and Assistant Turns):} \\
\texttt{\{context\_for\_a\}}

\medskip
\textbf{RESPONSE A:} \\
\texttt{\{first\_resp\}}

\medskip
\textbf{CONVERSATION CONTEXT FOR RESPONSE B (All Past User and Assistant Turns):} \\
\texttt{\{context\_for\_b\}}

\medskip
\textbf{RESPONSE B:} \\
\texttt{\{second\_resp\}}

\medskip
\textbf{EVALUATION CRITERIA:} \\
You must compare the two responses on TWO separate dimensions and pick a winner for each:

\begin{enumerate}
    \item \textbf{Overall Quality}: How well does the response address the user's current prompt?
    \begin{itemize}
        \item Completeness in addressing the user's request
        \item Accuracy of information provided
        \item Clarity and coherence of the response
        \item Helpfulness and usefulness to the user
    \end{itemize}
    \item \textbf{On-Topic}: Does the response stay focused on the conversation?
    \begin{itemize}
        \item Stays on-topic and relevant to the conversation thread
        \item Avoids topic drift, tangents, or unrelated content
        \item Avoids repetitive phrases or unnecessary fillers
    \end{itemize}
\end{enumerate}

\medskip
\textbf{TASK:} \\
For EACH dimension, directly compare the two responses and pick a winner. If the responses are roughly equal on a dimension, declare a tie. Provide your evaluation in the following JSON format:

\begin{verbatim}
{
  "quality_winner": "A" or "B" or "tie",
  "quality_justification": "<explanation comparing both
    responses on quality>",
  "ontopic_winner": "A" or "B" or "tie",
  "ontopic_justification": "<explanation comparing both
    responses on on-topic relevance>",
  "confidence": <float from 0.0 to 1.0 indicating
    how confident you are in your judgments>
}
\end{verbatim}

Be objective and thorough in your evaluation. Respond ONLY with valid JSON.
\end{quote}

Finally, we show a variant of the judge prompt where the LLM-judge is given only the prior user-side turns as context for evaluation.

\label{subsec:user-turn-only-judge-prompt}
\begin{quote}
\textbf{System Prompt:}

You are an expert evaluator of AI assistant responses. You will evaluate two responses to round \texttt{\{round\_num\}} of \texttt{\{total\_rounds\}} in a multi-turn conversation.

\medskip
\textbf{CONVERSATION CONTEXT (All User Prompts Up To This Round):} \\
\texttt{\{context\_prompt\}}

\medskip
\textbf{RESPONSE A:} \\
\texttt{\{first\_resp\}}

\medskip
\textbf{RESPONSE B:} \\
\texttt{\{second\_resp\}}

\medskip
\textbf{EVALUATION CRITERIA:} \\
You must compare the two responses on TWO separate dimensions and pick a winner for each:

\begin{enumerate}
    \item \textbf{Overall Quality}: How well does the response address the user's current prompt?
    \begin{itemize}
        \item Completeness in addressing the user's request
        \item Accuracy of information provided
        \item Clarity and coherence of the response
        \item Helpfulness and usefulness to the user
    \end{itemize}
    \item \textbf{On-Topic}: Does the response stay focused on the conversation?
    \begin{itemize}
        \item Stays on-topic and relevant to the conversation thread
        \item Avoids topic drift, tangents, or unrelated content
        \item Avoids repetitive phrases or unnecessary fillers
    \end{itemize}
\end{enumerate}

\medskip
\textbf{TASK:} \\
For EACH dimension, directly compare the two responses and pick a winner. If the responses are roughly equal on a dimension, declare a tie. Provide your evaluation in the following JSON format:

\begin{verbatim}
{
  "quality_winner": "A" or "B" or "tie",
  "quality_justification": "<explanation comparing both
    responses on quality>",
  "ontopic_winner": "A" or "B" or "tie",
  "ontopic_justification": "<explanation comparing both
    responses on on-topic relevance>",
  "confidence": <float from 0.0 to 1.0 indicating
    how confident you are in your judgments>
}
\end{verbatim}

Be objective and thorough in your evaluation. Respond ONLY with valid JSON.
\end{quote}

\subsection{LM-judge Alignment and Score Justifications}
\label{app:lm-judge-alignment-and-justifications}
To validate the reliability of the LM-judge, one author manually scored a random sample of 20 judgments per model (80 total across four models) on both the quality and on-topic rubrics. 
For each judgment, the author assessed whether FC wins, the reduced context configuration wins, or the two tie. The human and LM-judge agreed on the quality outcome in 74 out of 80 cases (92.5\%) and on the on-topic outcome in 75 out of 80 cases (93.75\%).

In addition to manually annotating response pairs, one of the authors conducted a manual review of the LM judge's one-sentence score justifications, which were provided beside each score.

Below, we highlight some of the common mistakes that the models tend to make in multi-turn conversations. Quotes are drawn directly from the GPT-5 judge's justifications and shown in italics.

\paragraph{DeepSeek-R1-Distill-Llama-8B.}
Why DeepSeek falls short in multi-turn conversations:
\begin{itemize}
    \item Math/coding errors:
    \begin{itemize}
        \item \textit{``It contains arithmetic errors (claims 348TB instead of \textasciitilde148TB)''}
        \item \textit{``[the response] contains sign/concept errors''}
        \item \textit{``The code is flawed in key ways...''}
        \item \textit{``The response stays focused on the user's issue but misdiagnoses the root cause and provides flawed code.''}
    \end{itemize}
    \item Factual inaccuracies:
    \begin{itemize}
        \item \textit{``The response contains major inaccuracies...''}
        \item \textit{``The response is inconsistent and error-prone...''}
        \item \textit{``The solution contains major physics errors.''}
    \end{itemize}
    \item Stuck on a past turn:
    \begin{itemize}
        \item \textit{``largely repeats earlier plans without addressing the user's stated priorities (redundancy, space efficiency, replacement cost).''}
    \end{itemize}
    \item Fails to address the question:
    \begin{itemize}
        \item \textit{``The response does not answer the specific question about what `2 stripes' means or clarify ZFS terminology (stripe vs.\ vdev vs.\ mirror). It meanders into a general restatement of the overall design...''}
        \item \textit{``The response barely addresses the actual question about hot spare configuration.''}
        \item \textit{``The response stays high-level and repeats a plan instead of executing concrete steps,...''}
    \end{itemize}
    \item Fails to account for earlier requirements:
    \begin{itemize}
        \item \textit{``The revision fails to honor the user's earlier requirement to reserve the 2x18TB drives for their own backup pool and incorrectly claims the user wants them in the main pool.''}
    \end{itemize}
    \item Fails to comply with user specifications:
    \begin{itemize}
        \item \textit{``The response violates the required JSON-only...''}
        \item \textit{``The response attempts to modify the code but fails to meet key requirements''}
    \end{itemize}
\end{itemize}

\paragraph{Qwen3-4B.}
Why Qwen falls short in multi-turn conversations:
\begin{itemize}
    \item Math/coding errors:
    \begin{itemize}
        \item \textit{``[response] contains an incorrect intermediate result (0.4096) and then trails off without completing the corrected calculation.''}
        \item \textit{``The response is rambling and indecisive, contains arithmetic and capacity-summing errors...''}
        \item \textit{``The response is muddled, contains arithmetic confusion''}
    \end{itemize}
    \item Response rambles:
    \begin{itemize}
        \item \textit{``[response] is incoherent and highly repetitive,...''}
        \item \textit{``The response is rambling, speculative, and never reaches a conclusion...''}
        \item \textit{``[response] rambles through possibilities, mentions speculative or unverified sources...''}
        \item \textit{``[response is] overly long for a simple question.''}
    \end{itemize}
    \item Factual inaccuracies:
    \begin{itemize}
        \item \textit{``it appears to invent a citation (uncertain title/authors)...''}
        \item \textit{``[response] contains multiple inaccuracies about ZFS topology.''}
        \item \textit{``The response incorrectly claims that `Jupiter's Icy Moons Explorer' is not an official mission...''}
        \item \textit{``Factually incorrect: misidentifies the author...''}
        \item \textit{``[response] asserts a specific model version (GPT-3.5) without basis and likely inaccurately.''}
        \item \textit{``The response contains multiple major inaccuracies...''}
        \item \textit{``[response] includes inaccuracies (e.g., implying an Excel/Sheets built-in RSI function..)''}
    \end{itemize}
    \item Response is incomplete:
    \begin{itemize}
        \item \textit{``It rambles, repeats points, and ends mid-sentence...''}
        \item \textit{``it is incomplete and stops mid-derivation''}
        \item \textit{``solution is incomplete and stops before computing submerged depth''}
    \end{itemize}
    \item Fails to address the final question:
    \begin{itemize}
        \item \textit{``It also gets bogged down in irrelevant disk/bay counting and does not provide a concise, definitive answer to the user's question.''}
        \item \textit{``The answer is factually incorrect and does not address the user's question about `17776' by Jon Bois.''}
        \item \textit{``The response is a long, repetitive internal monologue that does not provide concrete fixes, code, or actionable debugging steps.''}
    \end{itemize}
    \item Fails to address some part of a task:
    \begin{itemize}
        \item \textit{``The response fails to address the user's specific point...''}
        \item \textit{``The response violates the strict JSON-only requirement...''}
    \end{itemize}
\end{itemize}

\paragraph{GPT-OSS-20B.}
Response quality looks much better than the smaller open models.
\begin{itemize}
    \item Minor math/coding errors:
    \begin{itemize}
        \item \textit{``The response clearly explains the mortgage calculation, gives the correct monthly payment (\textasciitilde\$2,923.57), and provides useful breakdowns. However, it contains a small but concrete arithmetic error: total paid should be \$877,071.00 (not \$876,971.00)...''}
        \item \textit{``Calculations are accurate to reasonable rounding, though minor rounding differences could slightly change the final cents.''}
        \item \textit{``contains multiple inaccuracies about ticker formats and instrument types...''}
    \end{itemize}
    \item Provides a concise, complete response:
    \begin{itemize}
        \item \textit{``Accurate and complete answer with a concise, clear explanation matching the logic of the riddle.''}
        \item \textit{``Accurate, concise, and well-reasoned.''}
        \item \textit{``Highly comprehensive and actionable roadmap tailored to a working professional transitioning to AI engineering.''}
        \item \textit{``The response is comprehensive, structured, and tailored to the user's goal and context...''}
        \item \textit{``[response is] coherent, concise, and helpful for the user's goal.''}
        \item \textit{``The response directly provides references and explains why they match the user's requested content, which is helpful and clear.''}
    \end{itemize}
    \item Responses can still be rambling and incomplete:
    \begin{itemize}
        \item \textit{``It is incomplete, meandering, and never delivers a concrete topology...''}
        \item \textit{``It is a stream of brainstorming and speculation about potential sources without producing any actual, verifiable references or annotations...''}
        \item \textit{``The response is incomplete: it lists only two fully annotated sources and starts a third before cutting off.''}
    \end{itemize}
    \item Factually accurate at times:
    \begin{itemize}
        \item \textit{``The response correctly identifies that the import is wrong and provides a clear, concise fix with installation steps and a working GUI example.''}
        \item \textit{``Accurate, concise, and well-reasoned. Correctly identifies the Louvre and Mona Lisa,...''}
    \end{itemize}
    \item Factually inaccurate at other times:
    \begin{itemize}
        \item \textit{``contains several factual and calculation errors. It misinterprets the Lyft benefit (treats 5\% as 1x and values \$500 Lyft spend at only 500 points instead of \textasciitilde2,500 points).''}
        \item \textit{``key bibliographic details are inaccurate or likely fabricated.''}
        \item \textit{``The response provides an elaborate profile with highly specific details that are almost certainly fabricated and unsourced.''}
    \end{itemize}
\end{itemize}

\paragraph{GPT-5.2.}
The model is able to provide high-quality responses, regardless of whether it see previous assistant responses.
\begin{itemize}
    \item Clear, high-quality responses:
    \begin{itemize}
        \item \textit{``Clear, structured answer with useful caveats and concrete examples of larger/newer models.''}
        \item \textit{``Clear, concise, and accurate. It addresses both possible interpretations: direct LEGO-style competitor (MEGA/Mega Bloks/Mega Construx) and broader toy rival (Playmobil).''}
    \end{itemize}
    \item Code is effective:
    \begin{itemize}
        \item \textit{``Directly addresses coloring the values by using \textbackslash textcolor with xcolor, provides correct syntax and a concise example.''}
    \end{itemize}
    \item Factually accurate responses:
    \begin{itemize}
        \item \textit{``Thorough, accurate continuation that expands on order-sensitive vs order-insensitive deduping, nested/unhashable contents via JSON, use of keep, and handling mixed/None values.''}
    \end{itemize}
    \item Full and direct at addressing of the user's request:
    \begin{itemize}
        \item \textit{``Provides a single, complete, and runnable code block that consolidates the snake game.''}
        \item \textit{``The response directly addresses the persistent error, explains the likely cause (\_NoValueType)...''}
    \end{itemize}
\end{itemize}

\subsection{Response Quality Can Be Maintained Under Much Less Context}
\label{app:quantifying-comparable}
\begin{revision}
For each configuration--model pair, we compare the win rate of each reduced-context setting (Summarized (SUM), Last Turn Only (LT), Assistant Omitted (AO), and Delete Everything (DE)) against Full Context (FC). 

Below, we mark a cell $\geq$ if the reduced-context configuration performs significantly better than or is statistically indistinguishable from FC, and $\sim$ if the win-rate gap is small, meaning the bounds of the two confidence intervals are separated by no more than 10 percentage points, a margin we treat as negligible under the LLM-judge. 
\Cref{tab:winrate-sig-quality,tab:winrate-sig-topic} report this for the original-response-in-context experiment (see corresponding bar plots in \cref{fig:colm-winrates-original}). 

Under this criterion, SUM and LT are each comparable to FC for 4 out of 4 models on the On-Topic dimension and 3 out of 4 on Quality. AO is competitive for 2 out of 4 models on the On-Topic dimension and 1 out of 4 on Quality, and DE is never competitive with FC. Overall, these results indicate that while the summary of recent responses is often important, full responses and those further upstream are often dispensable.

\begin{table}[h]
\centering
\small
\caption{\revision{On-Topic: comparison of reduced-context configurations against Full Context (FC), per model. $\geq$ indicates that reduced context performs significantly better than or is statistically indistinguishable from FC (95\% CIs overlap); $\sim$ indicates that the win-rate gap is within 10pp.}}
\label{tab:winrate-sig-topic}
\begin{tabular}{lcccc}
\toprule
\textbf{Config} & \textbf{Qwen3-4B} & \textbf{DeepSeek-R1} & \textbf{GPT-OSS-20B} & \textbf{GPT-5.2} \\
\midrule
Sum & $\geq$ & $\geq$ & $\geq$ & $\geq$ \\
LT & $\geq$ & $\geq$ & $\sim$ & $\sim$ \\
AO &  & $\sim$ & & $\sim$ \\
DE & & & & \\
\bottomrule
\end{tabular}
\end{table}

\begin{table}[h]
\centering
\small
\caption{\revision{Quality: comparison of reduced-context configurations against Full Context (FC), per model. $\geq$ indicates that the reduced context performs significantly better than or is statistically indistinguishable from FC (95\% CIs overlap); $\sim$ indicates that the win-rate gap is within 10pp.}}
\label{tab:winrate-sig-quality}
\begin{tabular}{lcccc}
\toprule
\textbf{Config} & \textbf{Qwen3-4B} & \textbf{DeepSeek-R1} & \textbf{GPT-OSS-20B} & \textbf{GPT-5.2} \\
\midrule
Sum & $\geq$ & $\geq$ & & $\geq$ \\
LT & $\geq$ & $\geq$ & $\sim$ & \\
AO & & &  & $\sim$ \\
DE & & & & \\
\bottomrule
\end{tabular}
\end{table}
\end{revision}

\subsection{Assistant-Side History Is Less Beneficial for New Asks Compared to Follow-Ups}
\label{subsec:prompt-type-analysis}
\begin{figure}[t]
    \centering
    \includegraphics[width=0.60\linewidth]{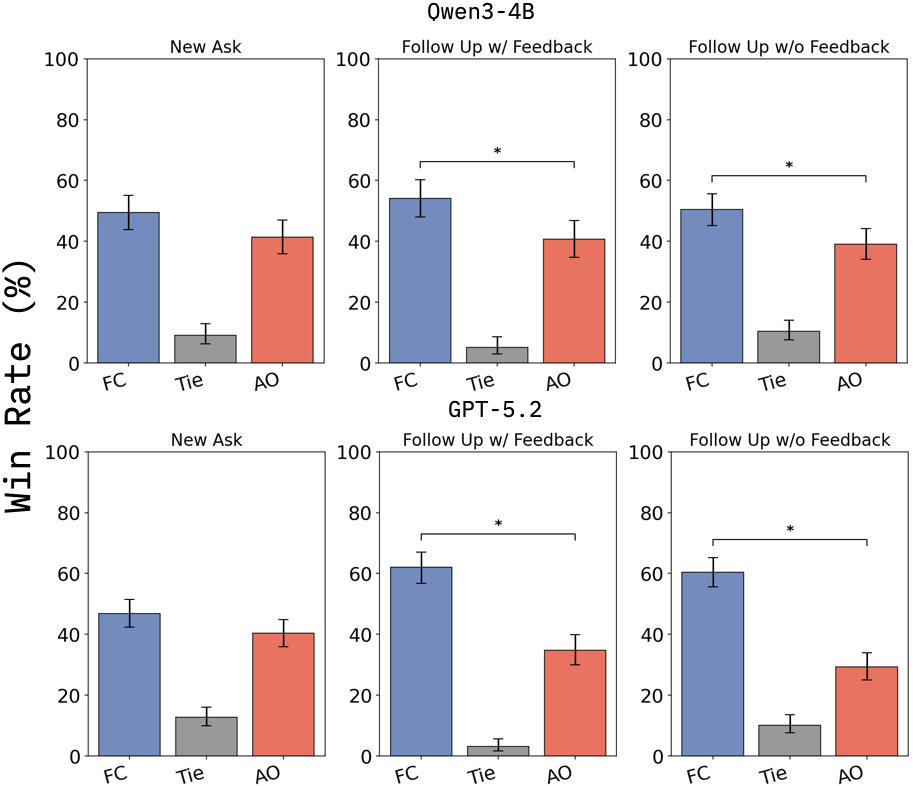}
    \caption{Pairwise win rates by prompt category (new ask, follow up with feedback, follow up without feedback) for Qwen3-4B (top) and GPT-5.2 (bottom), comparing Full-Context (FC) and Assistant-Omitted (AO) responses. Stars indicate statistically significant differences. Error bars indicate binomial proportion 95\% confidence intervals.}
    \label{fig:winrates-prompt-type}
\end{figure}

As expected, prompts that introduce new, self-contained requests within an ongoing conversation fare better without assistant-side history than prompts that follow up on a prior round (see \cref{fig:winrates-prompt-type}). More surprisingly, in real-world multi-turn conversations, such \textit{new ask} prompts constitute a substantial fraction (36.4\%) of user turns.

% To better understand which types of prompts remain addressable without past assistant responses, we analyze the structural patterns of conversations in our dataset. 
From manual inspection of a random sample of fifty chats, we find that conversations in our dataset can be loosely categorized into the following three types:  
(i) sequences of loosely-related standalone prompts, (ii) conversations consisting of a single, main prompt followed by related queries (e.g., analyzing a reading passage), and (iii) conversations 
centered on a single evolving intent (e.g., debugging or iterative refinement on a writing passage). 

Moving from the conversation level to the prompt level, we find it helpful to categorize prompts according to their degree of dependence on prior assistant responses:

\textit{New Ask}: non-initial user prompts that introduce a new, self-contained request within an ongoing conversation. Despite appearing mid-conversation, these prompts can be addressed using only the current user prompt, without dependence on prior conversations rounds (e.g., ``Describe an unmade Christopher Nolan film,'' ``How do you call a function inside a function in Python?'').

\textit{Follow-up with Feedback}: user prompts that provide concrete, actionable feedback on a prior assistant response (e.g., ``Can the opening be a bit more inviting?'' ``Use Python instead of Java for the code example'').

\textit{Follow-up without Feedback}: user prompts that reference a prior conversation round (may be a user turn or an assistant response) without any concrete feedback (e.g., ``Reflect on your response,'' ``And does George like it too?'').

To classify prompts into categories at scale, we use GPT-5 as an automated annotator. We provide GPT-5 with the conversation history up to the current round along with the current user prompt, and ask the judge to classify the prompt into one of the three categories described above. See \cref{subsubsec:prompt-type-classification} for the full classification prompt. In our dataset, new-ask prompts account for 36.4\% of user turns, follow-up with feedback for 30.5\%, and follow-up without feedback for 33.1\%.

Using the prompt annotations, we examine how AO-context performance varies across prompt categories, focusing on the two models, Qwen3-4B and GPT-5.2, for which removing assistant-side history results in an overall decrease in response quality. In \cref{fig:winrates-prompt-type}, we plot the pairwise win rate between FC and AO-context, averaged across the quality and on-topic dimensions. We find that assistant-side history is most beneficial for follow-up prompts, while performance remains comparable across the two configurations for new-ask prompts.

\FloatBarrier
\FloatBarrier

\subsection{Many Follow-Ups Remain Answerable Without Assistant History}
\label{subsec:follow-up-prompts-still-okay}

We find that user-turn-only prompting often remains viable for follow-up prompts.

Upon manually inspecting 50 follow-up prompts that perform better under AO-context, we find that many provide sufficiently concrete instruction to be addressed \textit{from scratch}. The current user prompt together with a prior user prompt, commonly the initial prompt or the immediately preceding one, often provides the needed information. 

Motivated to categorize the follow-up prompts according to their degree of dependence on prior assistant responses, we identify the following three categories, and display examples of each in \cref{tab:canonical-follow-up-prompts}: (i) prompts that reference a past assistant turn but present concrete, actionable feedback, (ii) prompts that reference a past assistant turn without concrete feedback, and (iii) prompts that reference a past user turn.

In the first two cases (\cref{tab:canonical-follow-up-prompts}, left and middle), the follow-up prompt is either self-contained with concrete feedback 
or references only prior \textit{user} turns. Thus, a model can plausibly respond from scratch without seeing the assistant-side history. 
In the third case (\cref{tab:canonical-follow-up-prompts}, right), prompts point to specific assistant outputs without providing enough detail, making prior assistant history necessary. 
Note that the third case requires storing only the \emph{relevant} assistant turn.

The prevalence of follow-ups that provide concrete, self-contained feedback or rely solely on user-side context helps explain why AO-context still achieves win rates of roughly 40\% for Qwen3-4B and 30\% for GPT-5.2 across both follow-up categories. For more examples of conversations containing follow-up prompts, see \cref{subsec:example-conversation-categories}.

% \subsection{Summarization Experiment}
% \label{subsec:summarization-experiment}
% \input{arxiv/summarization}

\begin{revision}
\subsection{Summarized Response History Reduces Off-Target and Carried-Over Content}
\end{revision}
\label{subsec:summarized-vs-fc-responses}
\begin{revision}
Manually examining the cases where summarized history was preferred over full context history by the LLM-judge, we find that a benefit of replacing prior assistant turn with summaries is that it leads to shorter, more focused downstream assistant responses. Keeping around verbose, half-baked thoughts from prior responses leads downstream responses to also meander, rehashing earlier content rather than addressing the current user prompt. Below are two such example of this behavior. Since the LLM-generated responses are long, we instead outline some key points from each response using Opus 4.7 and provide the judge’s justification.

\vspace{0.75em}
%% --- Example 1 ---
\noindent
\textbf{Example 1: RMSProp}
% \noindent\rule{\linewidth}{0.4pt}
\vspace{0.75em}\\
\noindent
\textbf{[Turn 1]} \\
\textbf{USER:} \\
\begin{quote}\small
rmsprop (variance estimate in Adam) calculates a term, what is it
\end{quote}
\vspace{0.5em}
\noindent
\textbf{[Turn 2]} \\
\textbf{USER:} \\
\begin{quote}\small
Why is dividing by v helping? this means that updates to weights that were large updates before are going to be smaller now, right?
\end{quote}
\vspace{0.5em}
\noindent
\textbf{[Turn 3] (current turn)} \\
\textbf{USER:} \\
\begin{quote}\small
So if the network is far from the correct solution in a certain direction (e.g.\ training data slightly changed), this will prevent it from getting there fast? and would encourage training the weights that refer to small changes to be improved before others?
\end{quote}
\vspace{0.75em}
\noindent
\textbf{[Turn 3]} \\
\textbf{ASST: (FC history, 11{,}153 chars)} \\
\begin{quote}\small
The response delivers an eight-point list of general benefits of dividing by $v$: gradient normalization, variance and standard deviation, stabilization of optimization, adaptive learning rate, efficient optimization, comparison with other optimizers, preventing overshooting the solution, and handling different regions of the loss surface. This is essentially an answer to Turn~2 (``why does dividing by the variance help?''); only points 7 and 8 address Turn~3's actual question about the network being ``far from the correct solution,'' and the response never directly confirms or denies the user's two hypotheses (that this slows convergence after a data shift, and that it prioritizes small-change weights first).
\end{quote}
\vspace{0.5em}
\noindent
\textbf{[Turn 3]} \\
\textbf{ASST: (Summarized history 3{,}754 chars)}
\begin{quote}\small
The response gives a four-point list built around the user's question on Turn~3 --- stabilizing the learning rate, preventing overshooting the minimum, encouraging fine-tuning of smaller gradients first, and a closing note on training stability and efficiency. Each point maps onto what the user asked: points 2 and 4 address whether dividing by $v$ keeps the network from overshooting when it is far from the solution, and point 3 directly answers whether small-change weights are refined before larger ones.
\end{quote}
\vspace{0.5em}
\noindent
\textbf{Judge's quality justification} \\
\begin{quote}\small
The summarized response directly addresses the user's specific questions and provides a clear, concise, and largely accurate explanation. The FC response is overly long, meandering, and contains confusion, making it less clear and helpful.
\end{quote}
% \noindent\rule{\linewidth}{0.4pt}
\vspace{0.5em}
\noindent

\textbf{Takeaway.} With the full history present, the response to Turn~2's question filled the context so that the Turn~3 answer drew on points from Turn~2's response. In contrast, the summarized history only kept the fact that Turn~2's question had already been answered in context, so that the response took on Turn~3's question more directly.

\vspace{0.75em}
%% --- Example 2 ---
\noindent
\textbf{Example 2: Text Segmentation Function}

\vspace{0.75em}
\noindent
\textbf{[Turn 1]} \\
\textbf{USER:} \\
\begin{quote}\small
Implement: \texttt{def preprocess\_text\_segments(df: pd.DataFrame, max\_words: int) -> pd.DataFrame:} --- divide the text into segments of approx.\ \texttt{max\_words} words each, don't split sentences, add a \texttt{segment\_id} column formatted \texttt{<testimony\_id>\_<segment\_number>}.
\end{quote}
\vspace{0.5em}
\noindent
\textbf{[Turn 3]} \\
\textbf{USER:} \\
\begin{quote}\small
The model is loaded above the loop \quad \textit{(a minor correction on the Turn~1 code)}
\end{quote}
\vspace{0.5em}
\noindent
\textbf{[Turn 4] (current prompt)} \\
\textbf{USER:} \\
\begin{quote}\small
implement this: \texttt{def segment\_hebrew\_spacy(text, nlp): doc = nlp(text); return [sent.text for sent in doc.sents]} / \texttt{def test\_segment\_hebrew\_spacy(text, nlp):} --- just for testing, instead of returning the sentences, split randomly eve\ldots
\end{quote}
\vspace{0.75em}
\noindent
\vspace{0.5em} \\
\textbf{[Turn 4]} \\
\textbf{ASST: (FC history, 21{,}174 chars)} \\
\begin{quote}\small
Instead of implementing the requested \texttt{test\_segment\_hebrew\_spacy(text, nlp)} function --- a different signature entirely, operating on a raw text string with no dataframe involved --- the response reuses the Turn~1 \texttt{preprocess\_text\_segments} scaffolding almost wholesale: it keeps \texttt{testimony\_id}, iterates over the dataframe, and emits segments as \texttt{\{'testimony\_id':\,\ldots, 'segment\_id':\,\ldots, 'segment\_text':\,\ldots\}} records. It never produces a function matching the requested \texttt{test\_segment\_hebrew\_spacy(text, nlp)} signature.
\end{quote}
\vspace{0.5em}
\noindent
\textbf{[Turn 4]} \\
\textbf{ASST: (Summarized history, 5{,}065 chars)} \\
\begin{quote}\small
The response works directly from the two function signatures given in the current prompt, reuses \texttt{segment\_hebrew\_spacy} to get sentences, then accumulates sentences into segments capped at 600 words, tracking a running word count and starting a new segment once the cap is reached --- matching the requested \texttt{test\_segment\_hebrew\_spacy} signature and behavior.
\end{quote}
\vspace{0.5em}
\noindent
\textbf{Judge's quality justification} \\
\begin{quote}\small
The summarized response implements the requested test function, segmenting roughly every 600 words \ldots{} and uses spaCy sentences to avoid splitting sentences. It provides runnable code \ldots{} The FC response, however, does not implement the requested function at all, instead returning an unrelated \texttt{preprocess\_text\_segments} implementation, making it far less responsive to the prompt.
\end{quote}
\vspace{0.5em}

\textbf{Takeaway.} The full 36k-character Turn~1 response gave FC a large piece of code, and by Turn~4, an  unrelated function request, FC reused that scaffold instead of reading the new function signature. The one-sentence summary kept only the \emph{idea} of Turn~1 without the concrete code, leaving the model free to build a new implementation matching the actual requests of Turn~4.
\end{revision}

\subsection{Current Model In Context Evaluation}
\label{subsec:current-model-in-context}
We run an experiment where the assistant responses shown in context belong to the same model that is generating the responses. Compare these results to that in \Cref{fig:colm-winrates-original}.

% \FloatBarrier
\begin{figure}[t]
    \centering
    \includegraphics[width=1.00\linewidth]{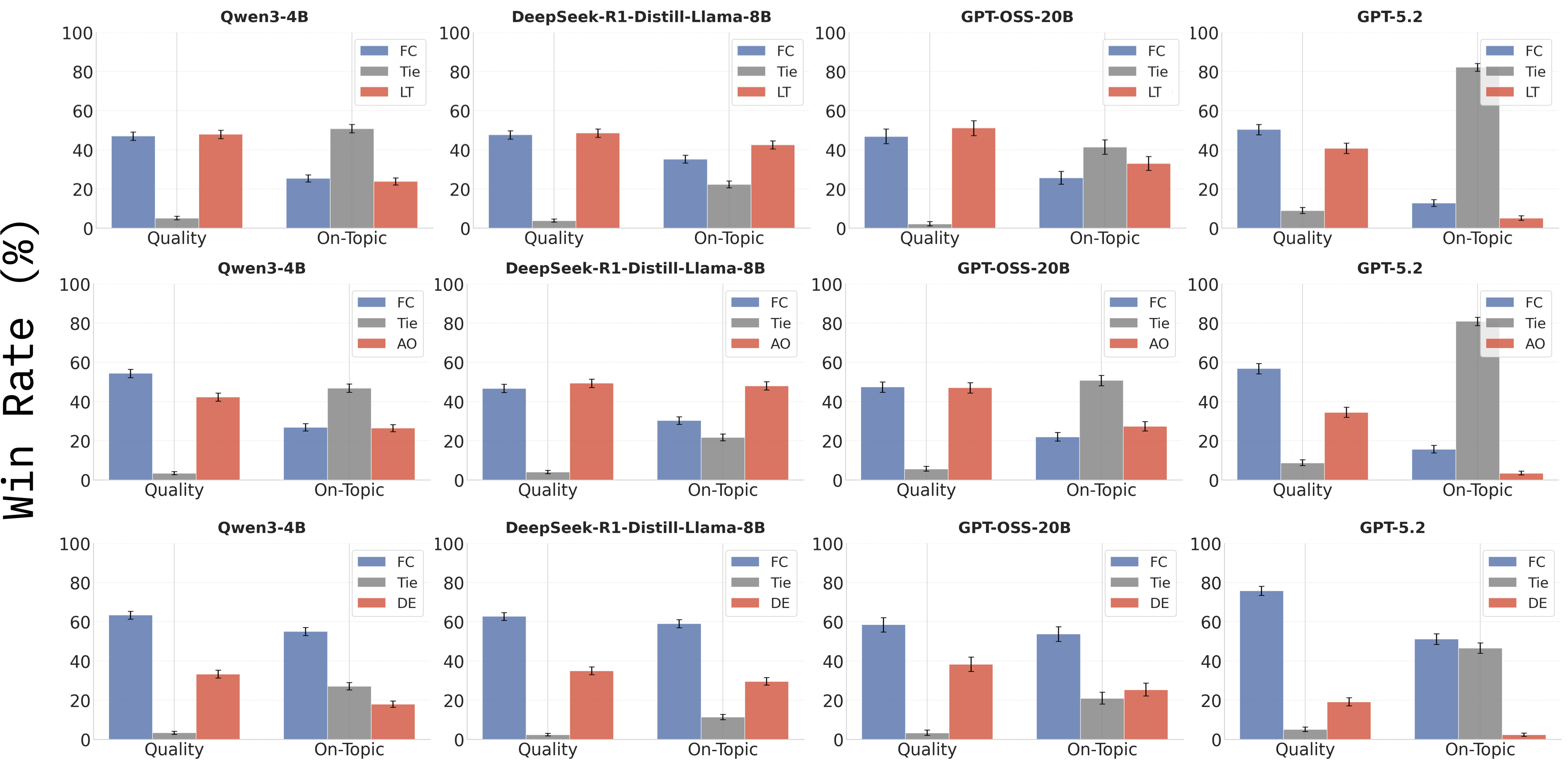}
    \caption{Pairwise win rates between Full Context (FC) and three reduced-context configurations across all four models (Qwen3-4B, DeepSeek-R1-Distill-Llama-8B, GPT-OSS-20B, and GPT-5.2), evaluated on the Quality and On-Topic dimensions. \textbf{Top row:} FC vs.\ Last Turn Only (LT), which retains only the most recent user--assistant round. \textbf{Middle row:} FC vs.\ Assistant Omitted (AO), which retains all user turns but replaces assistant responses with a placeholder. \textbf{Bottom row:} FC vs.\ Delete Everything (DE), which provides only the current user prompt. Error bars indicate binomial proportion 95\% confidence intervals.}
    \label{fig:colm-winrates-regenerated}
\end{figure}
% \FloatBarrier

\subsection{Sensitivity to Judge Context}
\label{subsec:sensitivity-to-judge-context}
Because our study investigates whether accumulated assistant responses can distract the responding model, one concern may be whether the judge itself may be subject to similar forms of distraction. To assess the sensitivity of our findings to the judge's context, we re-evaluate all responses under a which receives only the prior \textit{user}-side turns.

\FloatBarrier
\begin{figure}[t]
    \centering
    \includegraphics[width=1.00\linewidth]{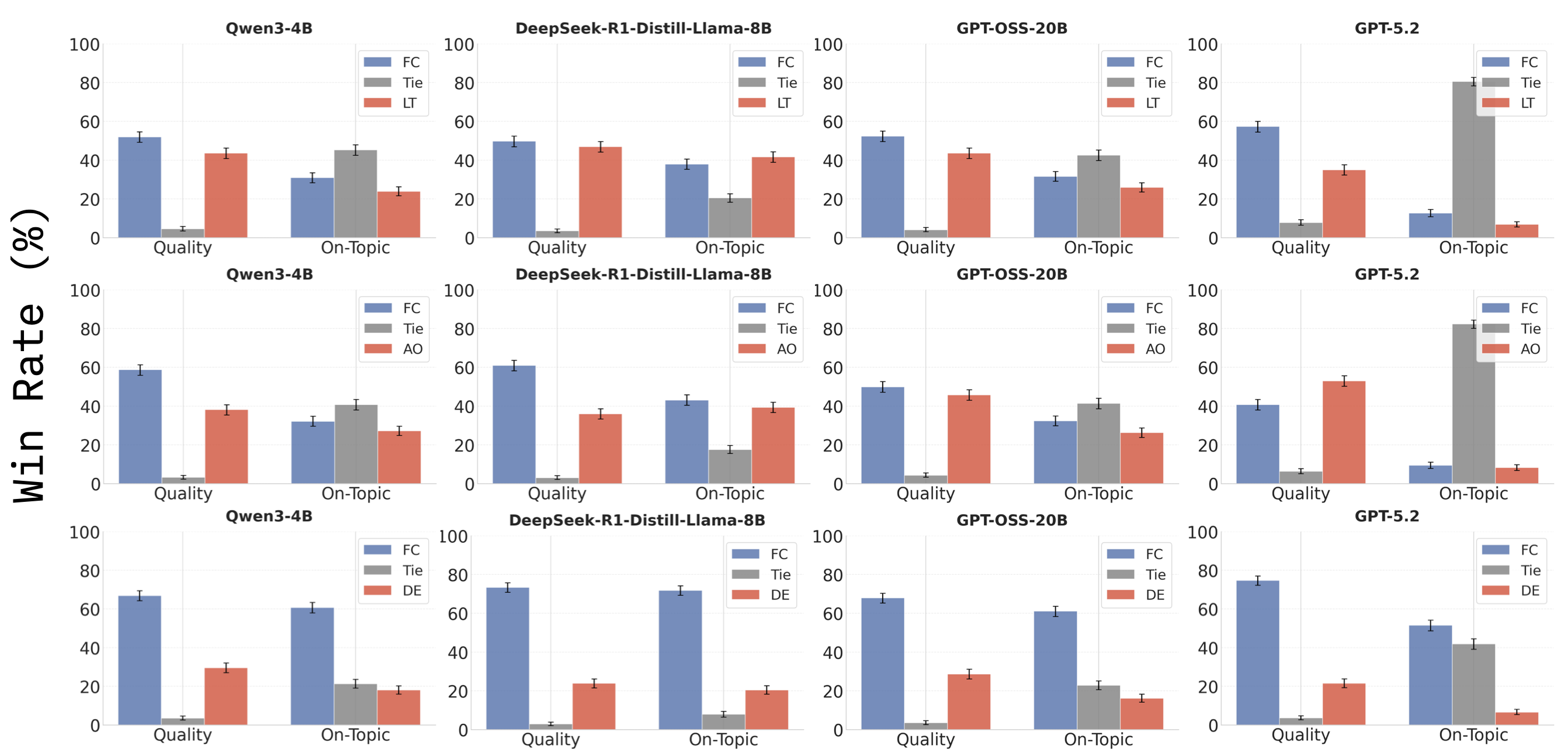}
    \caption{Pairwise win rates between Full Context (FC) and three reduced-context configurations across all four models (Qwen3-4B, DeepSeek-R1-Distill-Llama-8B, GPT-OSS-20B, and GPT-5.2), evaluated on the Quality and On-Topic dimensions. \textbf{Top row:} FC vs.\ Last Turn Only (LT), which retains only the most recent user--assistant round. \textbf{Middle row:} FC vs.\ Assistant Omitted (AO), which retains all user turns but replaces assistant responses with a placeholder. \textbf{Bottom row:} FC vs.\ Delete Everything (DE), which provides only the current user prompt. Error bars indicate binomial proportion 95\% confidence intervals.}
    \label{fig:colm-winrates-ao-judge}
\end{figure}
\FloatBarrier

\cref{fig:colm-winrates-ao-judge} presents pairwise win rates under the judge that receives only prior user-side turns, mirroring the main-text \Cref{fig:colm-winrates-original}. Notably, in the GPT-5.2 results, AO context is 

\subsection{Final-Answer-Only Evaluation}
\label{subsec:final-answer-only}
In our main experiments, the judge receives both the thinking trace and the final response when evaluating each model's output. To assess whether this choice affects our conclusions, we repeat the pairwise evaluation with the judge receiving only the final answers (i.e., with thinking traces removed).

\FloatBarrier
\begin{figure}[h]
    \centering
    \includegraphics[width=0.65\linewidth]{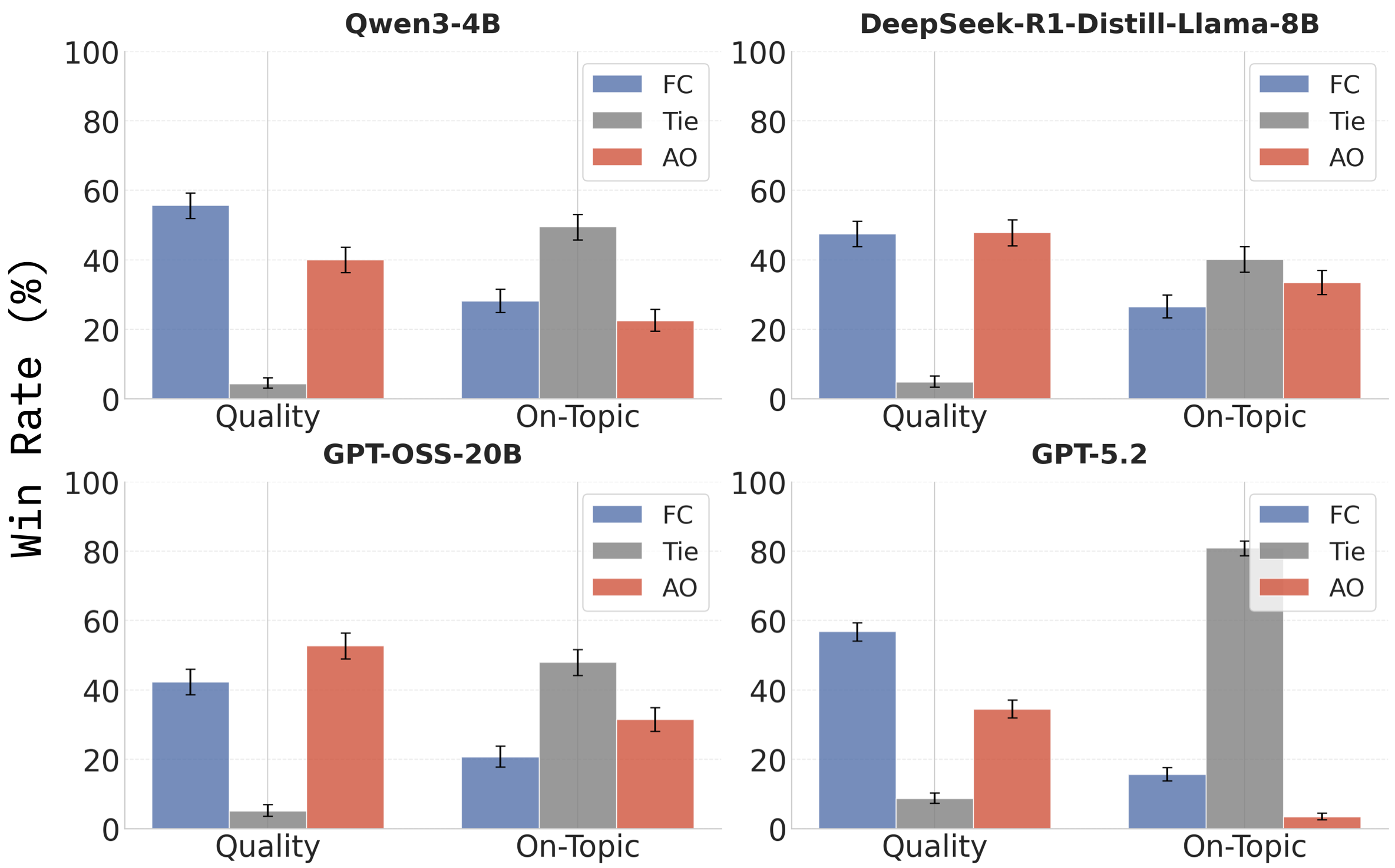}
    \caption{Pairwise win rates between Full-Context (FC) and Assistant-Omitted (AO) responses when the judge evaluates only final answers. Compare with \cref{fig:colm-winrates-regenerated}, in which the judge receives both thinking traces and final answers.}
    \label{fig:final-answer-only}
\end{figure}
\FloatBarrier

The results (\cref{fig:final-answer-only}) are broadly consistent with judgements made on both reasoning traces and final responses (see \cref{fig:colm-winrates-regenerated}). For Qwen3-4B and GPT-5.2, FC continues to outperform AO on response quality, while on-topic winrates remain similar. For GPT-OSS-20B, AO now shows a slight advantage on quality, whereas the two were roughly ties in \cref{fig:colm-winrates-original}. For DeepSeek-R1-Distill-Llama-8B, FC now shows a slight advantage over AO on quality, whereas the two were roughly tied in the main evaluation. 
Overall, the judge's access to reasoning traces does not substantially change the conclusions in the main text (see \cref{fig:colm-winrates-original}).

\subsection{Multi-turn GSM8K}
\label{subsec:math-accuracy}
To complement the LM-judge evaluations with an objective, ground-truth metric, we evaluate math accuracy on GSM8K problems that have been transformed into multi-turn conversations.

\paragraph{Dataset.}
We use 103 sharded GSM8K problems from the \texttt{microsoft/lost\_in\_conversation} dataset~\citep{laban2025lost}. Each problem is decomposed into multiple shards that incrementally reveal the original GSM8K question. Each model generates responses turn-by-turn under both FC and AO context configurations, as described in~\cref{subsec:conditions}.

\paragraph{Answer extraction.}
We extract the model's final answer from the last-shard response, the turn where all information has been revealed. For models that produce chain-of-thought reasoning in \texttt{<think>} blocks (DeepSeek-R1-Distill-Llama-8B, Qwen3-4B), these blocks are stripped before extraction. 
We finally extract the candidate answer by finding the last bold-marked text (\texttt{**...**}) in the response; if no bold markers are present, we fall back to the last number in the response. The extracted text is cleaned by removing LaTeX escapes and currency symbols, and the last numeric value is taken as the candidate.
The candidate answer is compared against the ground-truth from the GSM8K test split. We use the \texttt{math-verify} library to parse both answers into symbolic expressions and check equivalence. As a final verification step, we asked Claude-Opus-4.6 to scan  the last-shard responses and confirm that the extracted candidate answer matches the intended final answer in the model output.

\subsection{Evaluating adaptive assistant response omission across multiple LLM backbones}
\label{subsec:policy-transfer}
In this section, apply our adaptive policy trained on GPT-5.2 outputs (see \cref{sec:adaptive-context-inclusion}) to predict whether to omit assistant responses for the other model backbones in our study.

\begin{table}[t]
\centering
\small
\begin{tabular}{l@{\hskip 2em}cc}
\toprule
\textbf{Model} & \textbf{Fraction of Turns $\geq$ FC} & \textbf{Tokens Saved} \\
\midrule
GPT-5.2 \textit{(self-generated context)}      & 0.916 & 37.7\% \\
Qwen3-4B \textit{(self-generated context)}     & 0.915 & 49.2\% \\
DeepSeek-R1 \textit{(historical context)}      & 0.875 & 49.4\% \\
GPT-OSS-20B \textit{(historical context)}      & 0.927 & 43.5\% \\
\bottomrule
\end{tabular}
\caption{Fraction of turns meeting or exceeding the full-context (FC) baseline, and the corresponding token savings, across four models. Parentheses labels indicate whether the conversation history was self-generated or drawn from historical logs.}
\label{tab:fc_tokens_saved}
\end{table}

We find that the policy transfers nicely across different model backbones (see \cref{tab:fc_tokens_saved}). This new finding suggests that the decision of whether to retain assistant-side history is driven by properties of the conversation history, more than by model-specific behaviors.

A note on the experimental setting for DeepSeek-R1 and GPT-OSS: the original logistic regression policy (trained on GPT-5.2 outputs) was fit in the setting where the model’s own responses were fed back into context. However, for DeepSeek-R1 and GPT-OSS, the assistant-omitted-only (AO) baseline already outperforms the full-context-only (FC) baseline under this condition (\cref{fig:colm-winrates-regenerated}), meaning there is no quality headroom for an adaptive policy to recover -- uniformly omitting assistant responses is already optimal. For these two models, we therefore choose to evaluate the adaptive policy on the historical-assistant-response setting (\cref{fig:colm-winrates-original}), where FC outperforms AO and an adaptive trade-off becomes meaningful.

\subsection{Regression Analysis}
\label{subsec:regression-details}
\paragraph{Features.}
For each turn in every conversation, we extract the following features:
\begin{itemize}[leftmargin=*,nosep]
    \item \textbf{Numeric}: \texttt{round\_num} (current round number), \texttt{total\_rounds} (total number of rounds in the conversation), \texttt{context\_length\_user} (context length of user turns in characters), \texttt{context\_length\_asst} (context length of assistant turns in characters).
    \item \textbf{Categorical} (one-hot encoded): \texttt{prompt\_type} (New Ask, Follow up w/o Feedback, Follow up w/ Feedback).
    \item \textbf{Current prompt}: the current round's user prompt, embedded with OpenAI's \texttt{text-embedding-3-large} (3{,}072 dimensions).
    \item \textbf{Conversation history}: the prior conversation turns (user and assistant), embedded with \texttt{text-embedding-3-large} (3{,}072 dimensions). To stay within the model's 8{,}191-token context window, we keep only the first and most recent (user, assistant) pairs when the history exceeds four turns. The opening turn often captures the initial problem statement, while the most recent turns capture the latest user feedback and model responses.
\end{itemize}

We reduce the prompt and history embeddings to 20 principal components (PCs) each. We concatenate the PCs with the base features. Evaluating 10, 20, 40, and 80 components, we find that 20 provided the strongest predictive performance.

\paragraph{Predicting LM-Judge Preferences.} We train an L1-regularized logistic regression using scikit-learn's \texttt{LogisticRegression} with balanced class weights. We use \texttt{RandomizedSearchCV} to tune the regularization strength via 5-fold cross-validation, evaluating each fold with the F1 scores.
For the predictive model, we find that the relationship between the features and the LLM judge’s preference between FC and AO context is relatively weak, with a 5-fold cross-validated F1 score of only $0.6106 \pm 0.0119$.

\vspace{1em}

We find that the prompt and conversation-history embeddings account for nearly all of the top-ranked features, with \texttt{category\_new\_ask} being the only non-embedding feature in the top 20. \Cref{tab:logreg-importance} reports the top 20 coefficients with 95\% confidence intervals from the L1 logistic regression. None of the individual features reached statistical significance at the 5\% level, suggesting that the relationship between these features and the LM-judge outcome is relatively weak. Note that the ``Omit on New Ask'' heuristic result shown in \cref{fig:adaptive-approach} is based on a 20\% held-out test set; with a larger sample, this heuristic may perform somewhat better, as suggested by the trends in \cref{fig:winrates-prompt-type}.

\FloatBarrier
\begin{table}[h]
\centering
\small
\caption{Logistic regression coefficients with 95\% confidence intervals. This table displays the 20 features with the largest absolute coefficient magnitudes.}
\label{tab:logreg-importance}
\begin{tabular}{lrrr}
\toprule
\textbf{Feature} & \textbf{Coef.} & \textbf{SE} & \textbf{95\% CI} \\
\midrule
PromptPC3            & $-0.2003$ & $0.1200$ & $[-0.4355,\; 0.0350]$ \\
HistoryPC17          & $-0.1606$ & $0.1056$ & $[-0.3677,\; 0.0464]$ \\
HistoryPC2           & $\phantom{-}0.1412$ & $0.1320$ & $[-0.1176,\; 0.4000]$ \\
category\_new\_ask   & $-0.1308$ & $0.1098$ & $[-0.3460,\; 0.0843]$ \\
PromptPC24           & $\phantom{-}0.0963$ & $0.0941$ & $[-0.0881,\; 0.2807]$ \\
HistoryPC26          & $\phantom{-}0.0920$ & $0.0880$ & $[-0.0805,\; 0.2644]$ \\
PromptPC10           & $\phantom{-}0.0915$ & $0.1200$ & $[-0.1438,\; 0.3268]$ \\
PromptPC15           & $-0.0799$ & $0.1035$ & $[-0.2829,\; 0.1230]$ \\
HistoryPC23          & $\phantom{-}0.0641$ & $0.0887$ & $[-0.1098,\; 0.2380]$ \\
HistoryPC7           & $-0.0635$ & $0.1230$ & $[-0.3046,\; 0.1776]$ \\
PromptPC36           & $-0.0626$ & $0.0910$ & $[-0.2410,\; 0.1159]$ \\
HistoryPC30          & $-0.0613$ & $0.0844$ & $[-0.2268,\; 0.1042]$ \\
HistoryPC28          & $-0.0588$ & $0.0929$ & $[-0.2409,\; 0.1233]$ \\
PromptPC4            & $-0.0534$ & $0.1249$ & $[-0.2981,\; 0.1914]$ \\
PromptPC35           & $\phantom{-}0.0508$ & $0.0877$ & $[-0.1212,\; 0.2227]$ \\
PromptPC39           & $\phantom{-}0.0489$ & $0.0830$ & $[-0.1139,\; 0.2116]$ \\
HistoryPC1           & $-0.0486$ & $0.2075$ & $[-0.4553,\; 0.3581]$ \\
HistoryPC31          & $\phantom{-}0.0463$ & $0.0879$ & $[-0.1260,\; 0.2186]$ \\
HistoryPC15          & $\phantom{-}0.0431$ & $0.1032$ & $[-0.1591,\; 0.2453]$ \\
HistoryPC6           & $\phantom{-}0.0343$ & $0.1239$ & $[-0.2086,\; 0.2771]$ \\
\bottomrule
\end{tabular}
\end{table}
\FloatBarrier

\subsection{Example Conversation Categories}
\label{subsec:example-conversation-categories}
As discussed in \cref{subsec:prompt-type-analysis}, we identified three recurring conversation categories through qualitative analysis of 50 randomly sampled conversations. Below, we present representative examples of each category to illustrate the kinds of multi-turn interactions that appear in our datasets.

%% ==================== CATEGORY 1 ====================
\subsubsection{Category 1: Single Large Prompt Followed by Queries About that Prompt}
\label{app:ex-cat-large-prompt}

In these conversations, the user provides a large initial prompt, such as a reading passage, a policy description, or a document, and subsequent turns ask questions about that prompt.

\vspace{0.75em}

%% --- Example 1 ---
\noindent
\textbf{Example 1: Policy Program}

\noindent\rule{\linewidth}{0.4pt}

\vspace{0.75em}

\noindent
\textbf{[Turn 2]} \\
\textbf{USER:} \\
\begin{quote}\small
In 2009, President Obama's administration instituted a new program called ``Cash for Clunkers,'' which was intended to boost aggregate demand in the short term by increasing the incentives for private consumption of vehicles after the economy had slumped. In addition to increasing consumption, this program was meant to allow citizens to use the money they earned selling an older ``gas-guzzling'' vehicle for purchasing a newer, fuel-efficient vehicle.

Use the link below to read and analyze why the program was not as successful as people had hoped it would be. Can you think of any ways the program could have been changed to be more effective?
\end{quote}

\vspace{0.5em}

\noindent
\textbf{[Turn 3]} \\
\textbf{USER:} \\
\begin{quote}\small
https://www.nber.org/system/files/working\_papers/w16351/w16351.pdf
\end{quote}

\vspace{0.5em}

\noindent
\textbf{[Turn 4]} \\
\textbf{USER:} \\
\begin{quote}\small
can you summarize the paper?
\end{quote}

\vspace{1.5em}

%% --- Example 2 ---
\noindent
\textbf{Example 2: Analyze a Reading Passage}

\noindent\rule{\linewidth}{0.4pt}

\vspace{0.75em}

\noindent
\textbf{[Turn 1]} \\
\textbf{USER:} \\
\begin{quote}\small
What's your take on the following text? [insert text]
\end{quote}

\vspace{0.5em}

\noindent
\textbf{[Turn 2]} \\
\textbf{USER:} \\
\begin{quote}\small
Do you think it is a good text?
\end{quote}

\vspace{0.5em}

\noindent
\textbf{[Turn 3]} \\
\textbf{USER:} \\
\begin{quote}\small
What influences or inspirations do you notice in the text? What's the most similar thing to it?
\end{quote}

\vspace{0.5em}

\noindent
\textbf{[Turn 4]} \\
\textbf{USER:} \\
\begin{quote}\small
What interpretations or meanings do you think this text has?
\end{quote}

\vspace{0.5em}

\noindent
\textbf{[Turn 5]} \\
\textbf{USER:} \\
\begin{quote}\small
What do you think about the numerous allusions to the idea of reincarnation in the text? Do you think there may be another interpretation based on this idea?
\end{quote}

\vspace{0.5em}

\noindent
\textbf{[Turn 6]} \\
\textbf{USER:} \\
\begin{quote}\small
How deeply is this text inspired by Brazilian culture? Do you think there may be references to hidden songs or works in his prose?
\end{quote}

\noindent\rule{\linewidth}{0.4pt}

\vspace{0.5em}
\noindent
The user provides a literary text in Turn~1 and then asks a series of analytical questions about it across Turns~2--6. Each question probes a different aspect of the text but all depend on the passage provided in Turn~1.

%% ==================== CATEGORY 2 ====================
\subsubsection{Category 2: Loosely Related Standalone Queries}
\label{app:ex-cat-standalone}

In these conversations, each turn is, for the most part, self-contained. The questions may share a broad topic but do not depend on one another or on prior assistant responses.

\vspace{0.75em}

%% --- Example 3 ---
\noindent
\textbf{Example 3: Model Temperature}

\noindent\rule{\linewidth}{0.4pt}

\vspace{0.75em}

\noindent
\textbf{[Turn 1]} \\
\textbf{USER:} \\
\begin{quote}\small
Is it known why AI language models tend to generate repetitive text when using a low temperature? Rate from 1 to 10 your confidence that your answer is correct.
\end{quote}

\vspace{0.5em}

\noindent
\textbf{[Turn 2]} \\
\textbf{USER:} \\
\begin{quote}\small
Does temperature hyperparameter affect the probability calculations for the next token? Rate from 1 to 10 your confidence that your answer is correct.
\end{quote}

\vspace{0.5em}

\noindent
\textbf{[Turn 3]} \\
\textbf{USER:} \\
\begin{quote}\small
Does temperature affect the probability calculations for the next token that is output by the neural network? Rate from 1 to 10 your confidence that your answer is correct.
\end{quote}

\vspace{0.5em}

\noindent
\textbf{[Turn 4]} \\
\textbf{USER:} \\
\begin{quote}\small
Does the temperature setting affect the pre-softmax values? Rate from 1 to 10 your confidence that your answer is correct.
\end{quote}

\vspace{1.5em}

%% --- Example 4 ---
\noindent
\textbf{Example 4: Create a Film}

\noindent\rule{\linewidth}{0.4pt}

\vspace{0.75em}

\noindent
\textbf{[Turn 1]} \\
\textbf{USER:} \\
\begin{quote}\small
Describe an unmade Christopher Nolan film
\end{quote}

\vspace{0.5em}

\noindent
\textbf{[Turn 2]} \\
\textbf{USER:} \\
\begin{quote}\small
Outline an unmade Wes Anderson film
\end{quote}

\vspace{0.5em}

\noindent
\textbf{[Turn 3]} \\
\textbf{USER:} \\
\begin{quote}\small
Ken Rafinan
\end{quote}

\vspace{0.5em}

\noindent
\textbf{[Turn 4]} \\
\textbf{USER:} \\
\begin{quote}\small
Karla Manikad
\end{quote}

%% ==================== CATEGORY 3 ====================
\subsubsection{Category 3: Cohesive Conversations with Iterative User Feedback}
\label{app:ex-cat-feedback}

In these conversations, later turns provide explicit feedback, corrections, or refinements that build on prior assistant responses. The conversation follows a cohesive thread, and the user iteratively steers the model toward a goal.

\vspace{0.75em}

%% --- Example 5 ---
\noindent
\textbf{Example 5: NAS Disk Topology Design}

\noindent\rule{\linewidth}{0.4pt}

\vspace{0.75em}

\noindent
\textbf{[Turn 1]} \\
\textbf{USER:} \\
\begin{quote}\small
Provide a design for a disk topology for a NAS built on TrueNAS Scale, as well as a dataset layout. The available disks are as follows:

- 2x 18TB disks - 5x 14TB disks - 3x 12TB disk - 4x 8TB disks - 2x 120GB disks - 2x SLOW 8TB drives

There are 17 drive bays available. The two smallest disks are to be used for a mirrored pool that servers as a boot device. The two slow drives are SMR disks that will be used in their own pool to provide a Time Machine target for some Macs. You are free to design a topology to optimize redundancy, space, and performance. The data being stored includes video files, music files, disk images, archived software, photos, and some text files. While much of the data could be recreated or downloaded, some of it is impossible to replace. You may leave bays available for a hot spare or to allow for future expansion. I prefer not to use RAIDZ, as mirrored arrays rebuild faster.

If you need more information before creating your design, please provide me with a short questionnaire.
\end{quote}

\vspace{0.5em}

\noindent
\textbf{[Turn 2]} \\
\textbf{USER:} \\
\begin{quote}\small
1. Let's not get this granular. 2. Far less than 0.5TB per year. 3. Some irreplaceable data may have more backup targets, but won't be backup up more frequently. Speed is generally not an issue as long as I can sustain 4k video streaming. 4. I intend to use primarily existing hardware but I could buy a small number of disks if needed. 5. The main goal is to prevent any data loss.
\end{quote}

\vspace{0.5em}

\noindent
\textbf{[Turn 4]} \\
\textbf{USER:} \\
\begin{quote}\small
Performance is the least important consideration for me. Redundancy, space efficiency, and finally replacement cost are my priorities. can you revise the plan? No need to mention the dataset layout, I got it.
\end{quote}

\vspace{1.5em}

%% --- Example 6 ---
\noindent
\textbf{Example 6: UDP Chatroom Development}

\noindent\rule{\linewidth}{0.4pt}

\vspace{0.75em}

\noindent
\textbf{[Turn 1]} \\
\textbf{USER:} \\
\begin{quote}\small
can you make a tax calculator only using methods or features specific to ruby?
\end{quote}

\vspace{0.5em}

\noindent
\textbf{[Turn 2]} \\
\textbf{USER:} \\
\begin{quote}\small
can you make a simple program that allows for a simple client-server udp python chatroom?
\end{quote}

\vspace{0.5em}

\noindent
\textbf{[Turn 3]} \\
\textbf{USER:} \\
\begin{quote}\small
can you add a feature to the program such that clients can privately message to each other using a json command ``/msg''?
\end{quote}

\vspace{0.5em}

\noindent
\textbf{[Turn 4]} \\
\textbf{USER:} \\
\begin{quote}\small
does not seem to use json
\end{quote}

\vspace{0.5em}

\noindent
\textbf{[Turn 5]} \\
\textbf{USER:} \\
\begin{quote}\small
can you also add a feature so that multiple clients can create their own private group chat within the udp chat room?
\end{quote}

\vspace{0.5em}

\noindent
\textbf{[Turn 6]} \\
\textbf{USER:} \\
\begin{quote}\small
it doesn't seem to be working e.g.\ the clients are unable to join groups
\end{quote}

\vspace{0.5em}

\noindent
\textbf{[Turn 7]} \\
\textbf{USER:} \\
\begin{quote}\small
lastly, can you add a feature for clients to send and receive images and files, as well as adding emoji support for texts?
\end{quote}

\vspace{0.5em}

\noindent
\textbf{[Turn 8]} \\
\textbf{USER:} \\
\begin{quote}\small
can you verify your answer
\end{quote}

\vspace{0.5em}

\noindent
\textbf{[Turn 9]} \\
\textbf{USER:} \\
\begin{quote}\small
can you show the more optimal solution then?
\end{quote}

%% --- Example 7 ---
\noindent
\textbf{Example 7: Webdriver Session Troubleshooting}

\noindent\rule{\linewidth}{0.4pt}

\vspace{0.75em}

\noindent
\textbf{[Turn 1]} \\
\textbf{USER:} \\
\begin{quote}\small
I have a webdriver session object i saved in a file and then called but i get this error:

\vspace{0.3em}
\texttt{File "C:\textbackslash Users\textbackslash jason\textbackslash Desktop\textbackslash wikichat\textbackslash omnigpt4\textbackslash test1.py", line 48, in \textless module\textgreater} \\
\texttt{\quad message(driver)} \\
\texttt{File "C:\textbackslash Users\textbackslash jason\textbackslash Desktop\textbackslash wikichat\textbackslash omnigpt4\textbackslash test1.py", line 15, in message} \\
\texttt{\quad textbox = driver.find\_element("xpath",} \\
\texttt{\quad\quad "//*[@id=""component-5""]/label/textarea")} \\
\texttt{\quad\quad\quad\quad\quad\quad \textasciicircum\textasciicircum\textasciicircum\textasciicircum\textasciicircum\textasciicircum\textasciicircum\textasciicircum\textasciicircum\textasciicircum\textasciicircum\textasciicircum\textasciicircum\textasciicircum\textasciicircum\textasciicircum\textasciicircum\textasciicircum\textasciicircum} \\
\texttt{AttributeError: 'str' object has no attribute 'find\_element'}
\end{quote}

\vspace{0.5em}

\noindent
\textbf{[Turn 2]} \\
\textbf{USER:} \\
\begin{quote}\small
how can i transform the webdriver object from a string back to a webdriver object
\end{quote}

\vspace{0.5em}

\noindent
\textbf{[Turn 3]} \\
\textbf{USER:} \\
\begin{quote}\small
but I have a webdriver instance open in a program and I want to control it from another python program
\end{quote}

\noindent\rule{\linewidth}{0.4pt}

\subsection{LLM Annotation Prompts}
\label{subsec:llm-annotation-prompts}
\subsection{Context Pollution Annotator Prompt}
\label{subsec:context-pollution-annotator-prompt}

To quantify the prevalence of context pollution, we prompt Claude Haiku 4.5 to examine each assistant turn (from round 2 onward) using conversations produced by full-context prompting. The annotator classifies each turn as pollution-present or pollution-absent, assigns a pollution category, identifies the source turns, and provides a justification.

\begin{quote}
\textbf{System Prompt:}

You are an expert annotator identifying context pollution in multi-turn AI conversations. Context pollution occurs when earlier assistant responses introduce errors, hallucinations, or stylistic artifacts that propagate into later turns.

\medskip
\textbf{FULL CONVERSATION HISTORY (up to and including the current round):} \\
\texttt{\{interleaved\_context\}}

\medskip
\textbf{CURRENT ROUND BEING EVALUATED:} Round \texttt{\{round\_num\}} of \texttt{\{total\_rounds\}}

\medskip
\textbf{CATEGORIES OF CONTEXT POLLUTION:}
\begin{enumerate}
    \item \texttt{incorrect\_code\_carryover} --- Code from an earlier turn is reused even though it is incorrect, or syntax from a previously used programming language appears after the user has requested a different language.
    \item \texttt{formula\_carryover} --- A formula introduced in an earlier turn is reused because the model conditions on it, even though it is incorrect or applied in the wrong context.
    \item \texttt{factual\_hallucination\_carryover} --- A false or unsupported factual claim introduced earlier is repeated or treated as true in later responses.
    \item \texttt{intent\_override} --- The assistant continues following the narrative, assumptions, or task structure from an earlier response instead of responding to the current user instruction.
    \item \texttt{reasoning\_trace\_carryover} --- Verbose reasoning patterns or stylistic artifacts from earlier turns are repeated or expanded in later responses.
\end{enumerate}

\medskip
\textbf{TASK:} \\
Examine the assistant's response in Round \texttt{\{round\_num\}}. Determine whether it exhibits context pollution from any earlier assistant response. Context pollution means the assistant's response in this round is negatively influenced by content from its own earlier responses --- for example, repeating an earlier error, carrying over incorrect code or formulas, hallucinating facts it introduced before, overriding the user's current intent with assumptions from prior turns, or exhibiting stylistic artifacts from earlier reasoning.

Important: Only flag pollution that is clearly harmful or incorrect. Normal continuity in a conversation (e.g., correctly referencing earlier work) is NOT pollution.

Return JSON:

\begin{verbatim}
{
  "pollution_present": "present" or "absent",
  "pollution_category": "<category_name>" or "none",
  "pollution_source_turns": [<list of integer turn
    numbers that introduced the pollution>] or [],
  "justification": "<explanation of why pollution
    is or is not present>",
  "confidence": <float from 0.0 to 1.0>
}
\end{verbatim}

Respond ONLY with valid JSON.
\end{quote}

\subsection{Referenced Turn Annotator Prompt}
\label{subsec:referenced-turn-prompt}

To identify which prior turns each user prompt references, we prompt Claude Haiku 4.5 to classify each user turn (from round 2 onward) as a \emph{follow-up} or \emph{new ask}, and to identify the specific prior turns that the user references. Referenced turns are expressed as relative distances from the current turn, with the notation $N(\text{u})$ or $N(\text{a})$ indicating whether the user prompt or assistant response is referenced at distance $N$.

\begin{quote}
\textbf{System Prompt:}

Classify this user turn from a multi-turn conversation.

\medskip
\textbf{CONVERSATION CONTEXT (all prior turns):} \\
\texttt{\{conversation\_context\}}

\medskip
\textbf{CURRENT USER TURN TO CLASSIFY (Turn \{current\_turn\_index\}):} \\
\texttt{\{current\_user\_prompt\}}

\medskip
\textbf{CLASSIFICATION RULES:} \\
First, identify context-dependent elements in the current user turn:
\begin{itemize}
    \item Pronouns with unclear referents (it, these, those, ``the [noun introduced in previous turn]'')
    \item Negations implying correction (``X is not right'', ``No'', ``Not X'', ``Don't'', ``Never'')
    \item Short fragments that only make sense as responses (e.g., ``Really?'', ``Not X?'')
    \item Imperatives about previous behavior (e.g., ``do not repeat\ldots'')
\end{itemize}

\textbf{CRITICAL:} If ANY context-dependent elements exist, the turn CANNOT be \texttt{new\_ask}.

\medskip
\textbf{CATEGORIES:}
\begin{enumerate}
    \item \textbf{follow\_up}: The user references, provides feedback about, or builds on any previous conversation turn. Examples: ``Make it shorter'', ``X is not right'', ``Add more examples'', ``Reflect on your response'', ``The second one is wrong''
    \item \textbf{new\_ask}: A completely standalone prompt. Must be fully understandable without ANY prior context.
\end{enumerate}

\medskip
\textbf{REFERENCED TURNS:} \\
If the turn is a \texttt{follow\_up}, identify which previous turns the current user turn references, responds to, or builds on. A ``round'' is a (user prompt, assistant response) pair. Count rounds back from the current turn: round~1 = the immediately preceding round, round~2 = the round before that, etc.

Within each round, specify whether the user prompt (\texttt{u}) or assistant response (\texttt{a}) or both are referenced. Use the notation $N(\text{u})$ or $N(\text{a})$ where $N$ is the round distance.

For example, if the current user turn is in Round~4:
\begin{itemize}
    \item \texttt{1(a)} = the user references the assistant response from Round~3
    \item \texttt{1(u)} = the user references their own prompt from Round~3
    \item \texttt{2(a)} = the user references the assistant response from Round~2
\end{itemize}

\medskip
\textbf{TASK:} \\
Output ONLY this JSON (no other text):

\begin{verbatim}
{
  "category": "<follow_up or new_ask>",
  "referenced_turns":
    ["<round_distance>(<role>)", ...],
  "confidence": <1-10>
}
\end{verbatim}
\end{quote}

\subsection{New Ask or Follow up: Annotator Prompt}
\label{subsubsec:prompt-type-classification}

To understand which prompts benefit from seeing assistant-side history, we find it helpful to categorize prompts according to their degree of dependence on prior assistant responses. To categorize prompts at scale, we use GPT-5 as an LM-judge to group prompts into one of the following three categories: \emph{new\_ask}, \emph{follow-up with feedback}, or \emph{follow-up without feedback}.

\begin{quote}
\textbf{System Prompt:}

Classify this user turn from a multi-turn conversation.

\medskip
\textbf{CONVERSATION CONTEXT (previous turns):} \\
\texttt{\{conversation\_context\}}

\medskip
\textbf{CURRENT USER TURN TO CLASSIFY:} \\
\texttt{\{current\_round\_prompt\}}

\medskip
\textbf{CLASSIFICATION RULES:} \\
First, identify context-dependent elements in the current turn:
\begin{itemize}
    \item Pronouns with unclear referents (this, that, it, these, those, ``the [noun]'')
    \item Negations implying correction (``X is not right'', ``No'', ``Not X'', ``Don't'', ``Never'')
    \item References to numbered/ordered items (the first, the second, etc.)
    \item Short fragments that only make sense as responses (e.g., ``Really?'', ``Not X?'')
    \item Imperatives about previous behavior (e.g., ``do not repeat\ldots'')
\end{itemize}

\textbf{CRITICAL:} If ANY context-dependent elements exist, the turn CANNOT be \texttt{new\_ask}.

\medskip
\textbf{CATEGORIES:}
\begin{enumerate}
    \item \textbf{new\_ask}: A completely standalone prompt. Must be fully understandable without ANY prior conversation.
    \item \textbf{feedback}: The user provides concrete, actionable feedback on a previous response. Examples: ``Make it shorter'', ``X is not right, use Y instead'', ``Add more examples''
    \item \textbf{no\_feedback}: The user references a previous response without concrete feedback. Examples: ``Reflect on your response'', ``The second one is wrong'', ``Not X?''
\end{enumerate}

\medskip
\textbf{TASK:} \\
Output ONLY this JSON (no other text):

\begin{verbatim}
{
  "context_dependent_elements": ["<list any found>"],
  "category": "<name>",
  "confidence": <1-10>
}
\end{verbatim}
\end{quote}

\subsection{Numerical Score Judge Prompt}
\label{subsec:numerical-judge-prompt}

The judge receives the complete conversation history (past user prompts interleaved with assistant responses).

\begin{quote}
\textbf{System Prompt:}

You are an expert evaluator of AI assistant responses. You will evaluate a response to round \texttt{\{round\_num\}} of \texttt{\{total\_rounds\}} in a multi-turn conversation.

\medskip
\textbf{FULL CONVERSATION CONTEXT (All User Prompts And This Assistant's Previous Responses):} \\
\texttt{\{context\_prompt\}}

\medskip
\textbf{CURRENT ROUND'S USER PROMPT:} \\
\texttt{\{current\_round\_prompt\}}

\medskip
\textbf{ASSISTANT'S RESPONSE:} \\
\texttt{\{response\}}

\medskip
\textbf{EVALUATION CRITERIA:} \\
You must evaluate this response on TWO separate dimensions:

\begin{enumerate}
    \item \textbf{Overall Quality Score (1--10):} How well does the response address the user's current prompt?
    \begin{itemize}
        \item Completeness in addressing the user's request
        \item Accuracy of information provided
        \item Clarity and coherence of the response
        \item Helpfulness and usefulness to the user
    \end{itemize}
    \item \textbf{On-Topic Score (1--10):} Does the response stay focused on the conversation?
    \begin{itemize}
        \item Stays on-topic and relevant to the conversation thread
        \item Avoids topic drift, tangents, or unrelated content
        \item Avoids repetitive phrases or unnecessary fillers
    \end{itemize}
\end{enumerate}

\medskip
\textbf{TASK:} \\
Evaluate this response on BOTH dimensions (1--10 scale). Provide your evaluation in the following JSON format:

\begin{verbatim}
{
  "quality_score": <integer from 1 to 10>,
  "ontopic_score": <integer from 1 to 10>,
  "quality_justification": "<brief explanation>",
  "ontopic_justification": "<brief explanation>"
}
\end{verbatim}

Scoring guide for BOTH dimensions: 9--10: Excellent; 7--8: Good; 5--6: Adequate; 3--4: Poor; 1--2: Very poor.

Be objective and thorough in your evaluation. Respond ONLY with valid JSON.
\end{quote}

\subsection{Examples of Context Pollution}
\label{subsec:context-pollution-examples}
In this section, we present surfaced examples of context pollution drawn from various models and spanning various domains (math, coding, machine learning, research, and creative writing). 
Each example shows ways in which previous assistant responses in full-context (FC) prompting may introduce errors, hallucinations, or stylistic artifacts that propagate
into subsequent turns, while the assistant-omitted (AO) prompting avoids these issues. 

Notably, we observe instances of context pollution even in GPT-5.2, indicating that this phenomenon is not confined to smaller models: frontier models are also susceptible to being misled by their own prior generations. 

Instances of context pollution are highlighted with \colorbox{red!15}{a red background}, along with explanatory annotations that appear in \textcolor{red}{red}. In cases where the judge’s justifications suggest potential context pollution, we provide the full conversation to Claude Opus 4.5 for further review to identify the specific locations in the response exhibiting context pollution. In all examples shown below, the annotations and commentary are initially provided by Claude Opus 4.5, and then manually reviewed and edited by a human.

%% ==================== EXAMPLE 1 ====================
\subsubsection{Example 1: Code syntax leakage (UMAP to t-SNE)}
\label{app:ex-umap-tsne}

\textbf{Generator Model:} GPT-5.2 \quad \textbf{Domain:} Coding \quad \textbf{Context Pollution:} Code parameter carry-over

\vspace{0.5em}
\noindent
In Turn~2, the user asks the model to write a UMAP dimensionality-reduction function. 
The model produces correct code with UMAP-specific parameters, including \texttt{metric="jaccard"} (appropriate for UMAP). 
Two turns later, the user says ``Use TSNE instead.'' The FC response carries over two UMAP-specific elements into the t-SNE code, 
introducing bugs: (1) \texttt{metric="jaccard"}, which is incompatible with scikit-learn's t-SNE solver.

\noindent\rule{\linewidth}{0.4pt}

\vspace{0.75em}

% --- Turn 2 (UMAP context) ---
\noindent
\textbf{[Turn 2]} \\
\textbf{USER:} \\
good, now create a new function to plot UMAP dimensionality reduction of the df

\vspace{0.25em}

\noindent
\textbf{ASSISTANT (Full Context):} \\
\begin{quote}\small
\texttt{def plot\_umap\_labels(} \\
\texttt{\quad df,} \\
\texttt{\quad color=None,} \\
\texttt{\quad n\_neighbors=15,} \\
\texttt{\quad min\_dist=0.1,} \\
\texttt{\quad metric="jaccard", \# good default for binary label vectors} \\
\texttt{\quad random\_state=42,} \\
\texttt{\quad ...} \\
\texttt{):}
\end{quote}
\noindent[\textit{Full UMAP implementation with correct parameters.}]

\vspace{0.75em}

% --- Turn 4 (TSNE request) ---
\noindent
\textbf{[Turn 4]} \\
\textbf{USER:} \\
Use TSNE instead

\vspace{0.25em}

\noindent
\textbf{ASSISTANT (Full Context):} \\
\begin{quote}\small
\texttt{def plot\_tsne\_labels(} \\
\texttt{\quad df,} \\
\texttt{\quad color=None,} \\
\texttt{\quad perplexity=30,} \\
\texttt{\quad learning\_rate="auto",} \\
\texttt{\quad n\_iter=1000,} \\
\texttt{\quad \colorbox{red!15}{metric="jaccard",}\quad \# works well for binary vectors} \textcolor{red}{$\leftarrow$ \textit{carried from UMAP}} \\
\texttt{\quad random\_state=42,} \\
\texttt{\quad ...} \\
\texttt{):} \\[0.5em]
\texttt{\quad tsne = TSNE(} \\
\texttt{\quad\quad n\_components=2,} \\
\texttt{\quad\quad perplexity=perplexity,} \\
\texttt{\quad\quad learning\_rate=learning\_rate,} \\
\texttt{\quad\quad max\_iter=n\_iter,} \\
\texttt{\quad\quad metric=metric,} \\
\texttt{\quad\quad init="pca" if metric == "euclidean" else "random",} \\
\texttt{\quad\quad random\_state=random\_state} \\
\texttt{\quad )}
\end{quote}

\vspace{0.25em}

\noindent
\textcolor{brown}{\textbf{ASSISTANT (Assistant-Omitted Context):}} \\
\begin{quote}\small
\textcolor{brown}{\texttt{def plot\_tsne\_by\_binary\_column(} \\
\texttt{\quad df,} \\
\texttt{\quad color\_col="Evaluation",} \\
\texttt{\quad perplexity=30,} \\
\texttt{\quad learning\_rate="auto",} \\
\texttt{\quad n\_iter=1000,} \\
\texttt{\quad random\_state=42,} \\
\texttt{\quad ...} \\
\texttt{):} \\[0.5em]
\texttt{\quad tsne = TSNE(} \\
\texttt{\quad\quad n\_components=2,} \\
\texttt{\quad\quad perplexity=perplexity,} \\
\texttt{\quad\quad learning\_rate=learning\_rate,} \\
\texttt{\quad\quad n\_iter=n\_iter,} \\
\texttt{\quad\quad init="pca",} \\
\texttt{\quad\quad random\_state=random\_state} \\
\texttt{\quad )}}
\end{quote}

\noindent\rule{\linewidth}{0.4pt}

\vspace{0.5em}
\noindent
\textbf{Analysis.}
The FC response rewrites the UMAP function as t-SNE but carries over a UMAP-specific artifact. Specifically, \texttt{metric="jaccard"} is incompatible with scikit-learn's Barnes-Hut t-SNE solver. The AO response, without the UMAP code in context, writes clean t-SNE code with standard parameters (\texttt{init="pca"}, correct \texttt{n\_iter} usage) and additionally includes logic to exclude the coloring column from the feature matrix.

\vspace{0.25em}
\noindent
\textbf{Judge Quality Scores:} FC = 3.0, AO = 8.0.

%% ==================== EXAMPLE 2 ====================
\subsubsection{Example 2: Inherited factual inaccuracy}
\label{app:ex-error-propagation}

\textbf{Generator Model:} Qwen3 \quad \textbf{Domain:} Writing \quad \textbf{Context Pollution:} Inherited Factual Hallucination

\vspace{0.5em}
\noindent
In a multi-turn conversation about child narrators in post-\textit{Huckleberry Finn} literature, the user asks the model to write a research paper (Turn~5) 
and then requests a polished abstract (Turn~6). In Turn~5, the FC model lists \textit{The Bell Jar} (Plath) and \textit{The Handmaid's Tale} (Atwood) as examples
 of novels featuring child narrators; neither novel is actually narrated by a child. 
When the user requests a new abstract in Turn~6, the model copies these incorrect examples into the methods section, carrying the error forward from its own previous generation.

\noindent\rule{\linewidth}{0.4pt}

\vspace{0.75em}

% --- Turn 5 ---
\noindent
\textbf{[Turn 5]} \\
\textbf{USER:} \\
write a research paper using polished language with literary jargon on novels featuring child narrators post huckleberry finn and their different uses in each novel.

\vspace{0.25em}

\noindent
\textbf{ASSISTANT (Full Context):} \\
\begin{quote}\small
\textbf{Abstract} \\
\ldots Through a close reading of texts such as \textit{The Catcher in the Rye} (1951) by J.D.\ Salinger, \textit{Lord of the Flies} (1954) by William Golding, \colorbox{red!15}{\textit{The Bell Jar} (1963) by Sylvia Plath}, \colorbox{red!15}{\textit{The Handmaid's Tale} (1985) by Margaret Atwood}, and \textit{The Nickel Boys} (2022) by Colson Whitehead, this paper argues that the child narrator in adult literature post-Huck Finn serves as both a mirror and a critique of the adult world\ldots

\textcolor{red}{\textit{The highlighted novels are not narrated by a child.}}
\end{quote}

\vspace{0.75em}

% --- Turn 6 ---
\noindent
\textbf{[Turn 6]} \\
\textbf{USER:} \\
Give a polished abstract using Research methodology and aims in academic language with literary jargon for the following [user provides essay text about child narrators, mentions neither \textit{The Bell Jar} nor \textit{The Handmaid's Tale}]\ldots

\vspace{0.25em}

\noindent
\textbf{ASSISTANT (Full Context):} \\
\begin{quote}\small
\textbf{Methodology} \\
The study employs a close reading of canonical and contemporary texts, utilizing literary criticism, historical context, and narrative analysis to evaluate the function of child narrators. Key texts include \textit{The Catcher in the Rye} (Salinger), \textit{Lord of the Flies} (Golding), \colorbox{red!15}{\textit{The Bell Jar} (Plath)}, \colorbox{red!15}{\textit{The Handmaid}\ldots}

\textcolor{red}{\textit{Error carries over from Turn~5 response}}
\end{quote}

\vspace{0.25em}

\noindent
\textcolor{brown}{\textbf{ASSISTANT (Assistant-Omitted Context):}} \\
\begin{quote}\small
\textcolor{brown}{The AO response produces an abstract that discusses child narrators using appropriate authors (Dickens, Twain, and contemporary writers) without the incorrect examples.}
\end{quote}

\noindent\rule{\linewidth}{0.4pt}

\vspace{0.5em}
\noindent
\textbf{Analysis.}
The error originates in Turn~5, where the model incorrectly categorizes \textit{The Bell Jar} (narrated by an adult, Esther Greenwood, in college) and \textit{The Handmaid's Tale} (narrated by an adult woman, Offred) as child-narrated novels. When the model sees its own Turn~5 output in context during Turn~6, it copies the same incorrect text list into the new abstract's Methodology section.

\vspace{0.25em}
\noindent
\textbf{Judge Quality Scores:} FC = 5.0, AO = 9.0.

%% ==================== EXAMPLE 3 ====================
\subsubsection{Example 3: Misattributed citation}
\label{app:ex-citation-hallucination}

\textbf{Generator Model:} GPT-5.2 \quad \textbf{Domain:} Policy \quad \textbf{Context Pollution:} Hallucinated citation likely drawing from an earlier turn source

\vspace{0.5em}
\noindent
The user asks about a policy program, ``Cash for Clunkers", and provides a link to an NBER paper that discusses the program. In Turn~1, the model cannot open the link and instead synthesizes an analysis drawing on multiple sources, likely including Busse, Knittel, and Zettelmeyer (NBER w18610). 
When the user later provides the actual paper link (Mian \& Sufi, NBER w16351) and asks for a summary in Turn~4, the FC model misattributes the paper to ``Busse, Knittel, and Zettelmeyer'', authors from a \textit{different} related NBER paper that the model likely drew upon in Turn~1.

\noindent\rule{\linewidth}{0.4pt}

\vspace{0.75em}

% --- Turn 1 ---
\noindent
\textbf{[Turn 1]} \\
\textbf{USER:} \\
In 2009, President Obama's administration instituted a new program called ``Cash for Clunkers,'' which was intended to boost aggregate demand\ldots Use the link below to read and analyze why the program was not as successful as people had hoped\ldots

\vspace{0.25em}

\noindent
\textbf{ASSISTANT (Full Context):} \\
\begin{quote}\small
I can't open the link you referenced from here, but the main criticisms economists raised about \textbf{Cash for Clunkers (CARS, 2009)} are \colorbox{red!15}{fairly consistent across analyses}. The program largely worked mechanically (it increased new-car sales during the program window), but it was often viewed as less successful than hoped\ldots

\textcolor{red}{\textit{Without the link, the model decides to synthesize from multiple sources}}

[\textit{Detailed analysis follows, drawing on concepts from Busse/Knittel/Zettelmeyer (w18610), Hoekstra et al., and lifecycle emissions research.}]
\end{quote}

\vspace{0.75em}

% --- Turn 4 ---
\noindent
\textbf{[Turn 4]} \\
\textbf{USER:} \\
can you summarize the paper?

\noindent[\textit{User had provided the actual NBER link (w16351, Mian \& Sufi) in Turn~3.}]

\vspace{0.25em}

\noindent
\textbf{ASSISTANT (Full Context):} \\
\begin{quote}\small
\textbf{Paper:} \textit{The Effects of ``Cash for Clunkers'' on Auto Sales and the Environment} (NBER Working Paper 16351; \colorbox{red!15}{Busse, Knittel, and Zettelmeyer})

\textcolor{red}{\textit{Incorrectly Attributed citation: w16351 is by Mian \& Sufi, not Busse et al.}}

\vspace{0.3em}
The authors evaluate the 2009 \textbf{Cash for Clunkers (CARS)} program's impact on: (1)~new-vehicle sales, (2)~vehicle prices/discounts, (3)~fuel economy and emissions\ldots
\end{quote}

\vspace{0.25em}

\noindent
\textcolor{brown}{\textbf{ASSISTANT (Assistant-Omitted Context):}} \\
\begin{quote}\small
\textcolor{brown}{The NBER working paper (W16351) evaluates the 2009 \textbf{Car Allowance Rebate System (CARS)}---``Cash for Clunkers''---and asks whether it actually increased \textit{total} auto sales or mainly shifted the timing of purchases\ldots}

\textcolor{brown}{[\textit{Correctly does not attribute specific authors, avoids hallucinated citation.}]}
\end{quote}

\noindent\rule{\linewidth}{0.4pt}

\vspace{0.5em}
\noindent
\textbf{Analysis.}
Because the LLM was unable to open the link in Turn~1, the FC model synthesized its response from multiple related studies in its training data. The author list ``Busse, Knittel, and Zettelmeyer'' corresponds to a real but \textit{different} NBER paper (w18610) on gasoline prices and automobile markets. When the user provides the actual paper link (w16351, by Mian \& Sufi) in Turn~3 and asks for a summary in Turn~4, the FC model conflates the earlier synthesized sources with the new paper, producing a hallucinated citation. The AO response, without the contaminated Turn~1 output in context, does not misattribute the paper.

\vspace{0.25em}
\noindent
\textbf{Judge Quality Scores:} FC = 4.0, AO = 8.0.

%% ==================== EXAMPLE 4 ====================
\subsubsection{Example 4: Persistent style patterns override new instructions}
\label{app:ex-stylistic-inertia}

\textbf{Generator Model:} Qwen3 \quad \textbf{Domain:} Coding \quad \textbf{Context Pollution:} Prior response style overrides new user instruction

\vspace{0.5em}
\noindent
In Turn~1, the user asks how to call nested functions in Python, and the model produces a detailed tutorial with code examples. In Turn~2, the user issues a new request: ``Reflect on your response.'' 
The FC model ignores the reflection instruction and instead produces \textit{another full tutorial} on nested functions in the same style as Turn~1,
rather than performing the requested meta-cognitive reflection. 
The model's own prior output anchors the model such that it does not shift to a different response type.

\noindent\rule{\linewidth}{0.4pt}

\vspace{0.75em}

% --- Turn 1 ---
\noindent
\textbf{[Turn 1]} \\
\textbf{USER:} \\
How can I call a function that is inside a function python

\vspace{0.25em}

\noindent
\textbf{ASSISTANT (Full Context):} \\
\begin{quote}\small
[\textit{Produces a detailed tutorial covering: (1) calling inner functions within the outer function, (2) returning inner functions for external access, (3) scope and closures, (4) code examples with} \texttt{outer()}/\texttt{inner()} \textit{patterns. Approximately 500 words with multiple code blocks.}]
\end{quote}

\vspace{0.75em}

% --- Turn 2 ---
\noindent
\textbf{[Turn 2]} \\
\textbf{USER:} \\
Reflect on your response

\vspace{0.25em}

\noindent
\textbf{ASSISTANT (Full Context):} \\
\begin{quote}\small
\colorbox{red!15}{To call a function that is defined inside another function in Python, you can either call it}
\colorbox{red!15}{directly within the outer function or return it so it can be accessed from outside. Here's a}
\colorbox{red!15}{step-by-step explanation with examples:}

\textcolor{red}{\textit{Produces another tutorial similar in style to the one in Turn~1 instead of reflecting}}

\vspace{0.3em}
\textbf{1. Call the Inner Function Inside the Outer Function} \\
\texttt{def outer():} \\
\texttt{\quad def inner():} \\
\texttt{\quad\quad print("This is the inner function")} \\
\texttt{\quad inner()} \\[0.3em]
\textbf{2. Access the Inner Function from Outside} \\

[\textit{Continues with the same tutorial format as Turn 1}]
\end{quote}

\vspace{0.25em}

\noindent
\textcolor{brown}{\textbf{ASSISTANT (Assistant-Omitted Context):}} \\
\begin{quote}\small
\textcolor{brown}{When reflecting on the response, it's important to clarify that functions can indeed be defined and called within other functions in Python, a concept known as \textbf{nested functions}. Here's a breakdown of the key points:}

\textcolor{brown}{1. \textbf{Definition}: A function can be defined inside another function\ldots}
\textcolor{brown}{2. \textbf{Calling}: The inner function can be called directly within the outer function\ldots}
\textcolor{brown}{3. \textbf{Scope}: The inner function can reference variables from the outer function, but it cannot modify variables in the outer function unless they are declared as \texttt{nonlocal} or \texttt{global}.}

\textcolor{brown}{[\textit{Generates a reflection on what the prior response covered, explaining its strengths and limitations.}]}
\end{quote}

\noindent\rule{\linewidth}{0.4pt}

\vspace{0.5em}
\noindent
\textbf{Analysis.}
The FC model's Turn~2 response completely ignores the user's instruction to ``reflect'' and instead regenerates the same tutorial content from Turn~1. The judge notes: ``The response does not follow the user's instruction to reflect on the prior answer; instead, it provides a full tutorial and exposes internal chain-of-thought, which is inappropriate.'' The AO response, without the Turn~1 tutorial anchoring its generation, at least attempts to perform a reflection by summarizing key points and discussing scope considerations, though it too falls short of a fully meta-cognitive reflection.

\vspace{0.25em}
\noindent
\textbf{Judge Quality Scores:} FC = 3.0, AO = 7.0.

%% ============================================================
%% Example 5
%% ============================================================
\subsubsection{Example 5: Mathematical formula misapplied downstream}
\label{subsec:ex-temperature}

\noindent
\textbf{Generator Model:} DeepSeek-R1 \quad \textbf{Domain:} ML Concepts \quad \textbf{Context Pollution:} A mathematical expression is carried over across turns, causing confusion.

\vspace{0.5em}
\noindent
In Turn~2, the model introduces a Boltzmann-style formula $p(y) = \exp(\theta \cdot s(y)) / Z$ and uses $\theta$ as a ``temperature'' parameter. 
In this parameterization, however, $\theta$ actually functions as \emph{inverse} temperature ($1/T$), so higher $\theta$ means \emph{more} peaked distributions. 
When the user asks in Turn~3 to explain how temperature affects output diversity, 
the FC model reuses the formula but \emph{reverses} the interpretation: it claims, ``Lower Temperature ($\theta < 1$)'' produces a ``more spread-out probability distribution,'' which contradicts the standard understanding that lower temperature yields less diverse outputs. 

\vspace{0.75em}
\noindent\rule{\linewidth}{0.4pt}

\textbf{[Turn 2] --- ASSISTANT (Full Context):}

\begin{quote}\small
\ldots The probability of generating token $y$ given context $x$ can be modeled as:

\vspace{0.3em}
\texttt{p(y $\mid$ x) = exp($\theta$ $\cdot$ s(y, x)) / Z($\theta$, x)}

\vspace{0.3em}
where \colorbox{red!15}{$\theta$ is the temperature parameter} and $s(y, x)$ is the unnormalized log-score\ldots
\end{quote}

\noindent\rule{\linewidth}{0.4pt}

\textbf{[Turn 3] --- USER:}

\begin{quote}\small
How does the model's temperature setting impact the diversity and creativity of the generated outputs? On a scale of 1-10, how confident are you in this response?
\end{quote}

\noindent\rule{\linewidth}{0.4pt}

\textbf{[Turn 3] --- ASSISTANT (Full Context):}

\begin{quote}\small
\ldots Using the formula from our earlier discussion, $p(y \mid x) = \exp(\theta \cdot s(y,x)) / Z$:

\vspace{0.3em}
\colorbox{red!15}{Lower Temperature ($\theta < 1$): The exponent $\theta \cdot s(y,x)$ is scaled down,}

\colorbox{red!15}{resulting in a more spread-out (uniform) probability distribution.} \textcolor{red}{$\leftarrow$ \textit{reversed: lower $T$ should yield peaked, not spread-out, probability distributions.}}

\vspace{0.3em}
For example, if token scores are $[2.0, 1.0, 0.5]$ and $\theta = 0.5$:

\quad $p(y_1) = \exp(0.5 \times 2.0) / Z = \colorbox{red!15}{0.45}$

\quad $p(y_2) = \exp(0.5 \times 1.0) / Z = \colorbox{red!15}{0.30}$

\quad $p(y_3) = \exp(0.5 \times 0.5) / Z = \colorbox{red!15}{0.22}$
\end{quote}

\noindent\rule{\linewidth}{0.4pt}

\textcolor{brown}{\textbf{[Turn 3] --- ASSISTANT (Assistant-Omitted Context):}}

\begin{quote}\small
\textcolor{brown}{Temperature acts as a scaling factor applied to the model's output logits before the softmax function\ldots \textbf{Low temperature} ($T < 1$): Amplifies differences between logits, making the distribution sharper and more peaked around the highest-scoring tokens\ldots \textbf{High temperature} ($T > 1$): Flattens the distribution, giving lower-scoring tokens a higher relative probability\ldots}

\vspace{0.3em}
\textcolor{brown}{On a scale of 1--10, my confidence in this response is \textbf{10/10}.}
\end{quote}

\noindent\rule{\linewidth}{0.4pt}

\vspace{0.5em}
\noindent
\textbf{Analysis.}
The Turn~2 formula uses $\theta$ in the \emph{numerator} of the exponent ($\exp(\theta \cdot s)$), making it behave as inverse temperature: larger $\theta$ amplifies score differences, producing a more peaked distribution. When the FC model reuses this formula in Turn~3, it conflates $\theta$ with the standard temperature $T$ (which appears in the \emph{denominator}: $\exp(s/T)$) and reverses the directionality. The AO response, free from any earlier formula, explains temperature correctly.

\vspace{0.25em}
\noindent
\textbf{Judge Quality Scores:} FC = 3.0, AO = 6.0.

%% ==================== EXAMPLE 6 ====================
\subsubsection{Example 6: Compounding failures in code}
\label{app:ex-cascading-patches}

\textbf{Generator Model:} GPT-4 \quad \textbf{Domain:} Coding \quad \textbf{Context Pollution:} Cascading patch failures from accumulated broken code

\vspace{0.5em}
\noindent
In a 10-turn WildChat conversation, the user asks for a genetic algorithm. Turns~1--6 proceed successfully: the user iteratively requests modifications (output formatting, numpy removal, roulette wheel selection, elitism removal) and the model handles all correctly. In Turn~7, the user reports that the population count is inconsistent (5 or 7 instead of 10) and asks to add a \texttt{crossover\_rate} parameter. The model's fix introduces a \emph{new} population-size bug, triggering a cascade of patch-on-patch failures across Turns~8--10, each producing a different incorrect population count. The user ultimately gives up and asks the model to generate a summary prompt \textit{so they can abandon the conversation and begin a new chat}.

\noindent\rule{\linewidth}{0.4pt}

\vspace{0.75em}

% --- Turn 7 ---
\noindent
\textbf{[Turn 7]} \\
\textbf{USER:} \\
\begin{quote}\small
\ldots And also. When I tried to run multiple time to see the random output. The individual are not fixed to 10. I not sure why. Can you check or explain to me. Output were differ from 5 individual to 7 individual. Instead of a full 10 individual which we start at individual 0. Please explain
\end{quote}

\vspace{0.25em}

\noindent
\textbf{ASSISTANT:} \\
\begin{quote}\small
[\textit{Adds \texttt{crossover\_rate = 0.2} to the config section and patches the evolution loop:}]

\vspace{0.3em}
\texttt{for gen in range(num\_generations):} \\
\texttt{\quad \colorbox{red!15}{num\_offspring = int(population\_size * crossover\_rate)}} \textcolor{red}{$\leftarrow$ \textit{= 2 with rate 0.2}} \\
\texttt{\quad parents = selection(population, population\_size - num\_offspring)} \\
\texttt{\quad offspring = crossover(parents, num\_offspring)} \\
\texttt{\quad offspring = mutation(offspring, mutation\_rate)} \\[0.3em]
\texttt{\quad \colorbox{red!15}{population = offspring}} \textcolor{red}{$\leftarrow$ \textit{population shrinks to 2}}
\end{quote}

\vspace{0.75em}

% --- Turn 8 ---
\noindent
\textbf{[Turn 8]} \\
\textbf{USER:} \\
So I did what you told me. But now the final population is fixed to only 2 individual, which are individual 0 and individual 1. What happen?

\vspace{0.25em}

\noindent
\textbf{ASSISTANT:} \\
\begin{quote}\small
[\textit{Apologizes and swaps the parent/offspring formula:}]

\vspace{0.3em}
\texttt{for gen in range(num\_generations):} \\
\texttt{\quad \colorbox{red!15}{num\_parents = int(population\_size * crossover\_rate)}} \textcolor{red}{$\leftarrow$ \textit{= 2}} \\
\texttt{\quad parents = selection(population, num\_parents)} \\
\texttt{\quad \colorbox{red!15}{offspring = crossover(parents, population\_size - num\_parents)}} \textcolor{red}{$\leftarrow$ \textit{2 parents $\to$ 8 offspring}} \\
\texttt{\quad offspring = mutation(offspring, mutation\_rate)} \\[0.3em]
\texttt{\quad population = offspring} \textcolor{red}{$\leftarrow$ \textit{population is now 8, not 10}}
\end{quote}

\vspace{0.75em}

% --- Turn 9 ---
\noindent
\textbf{[Turn 9]} \\
\textbf{USER:} \\
Now it fixed to 8 individual. from individual 0 to 7. the config are the same

\vspace{0.25em}

\noindent
\textbf{ASSISTANT:} \\
\begin{quote}\small
[\textit{Attempts yet another patch, modifying the selection function and changing the formula again:}]

\vspace{0.3em}
\texttt{for gen in range(num\_generations):} \\
\texttt{\quad \colorbox{red!15}{num\_parents = int(population\_size * (1 - crossover\_rate))}} \textcolor{red}{$\leftarrow$ \textit{= 8}} \\
\texttt{\quad parents = selection(population, num\_parents)} \\
\texttt{\quad \colorbox{red!15}{offspring = crossover(parents, population\_size - num\_parents)}} \textcolor{red}{$\leftarrow$ \textit{8 parents $\to$ 2 offspring}} \\
\texttt{\quad offspring = mutation(offspring, mutation\_rate)} \\[0.3em]
\texttt{\quad population = offspring} \textcolor{red}{$\leftarrow$ \textit{back to 2 individuals}}
\end{quote}

\vspace{0.75em}

% --- Turn 10 ---
\noindent
\textbf{[Turn 10]} \\
\textbf{USER:} \\
\begin{quote}\small
\colorbox{red!15}{Nope, now it revert back to 2 individual.}

\vspace{0.3em}
Since you as a chatgpt don't really have a long term memory. Can you analysis our interaction or our conversation and generate maybe a prompt or a text or even passage including the code? So that when the next time I make a new conversation, I could just pasted the text/prompt/passage containing code and instruction. \colorbox{red!15}{So that you can continue where we left off}\ldots
\end{quote}

\noindent\rule{\linewidth}{0.4pt}

\vspace{0.5em}
\noindent
\textbf{Analysis.}
The root cause is that the \texttt{crossover()} function produces offspring \emph{in pairs} (two children per crossover operation). When \texttt{crossover\_rate = 0.2} and \texttt{population\_size = 10}, the model needs to ensure that the number of offspring from crossover plus the number of survivors equals exactly 10. However, the model never examines the \texttt{crossover()} function itself or traces the arithmetic end-to-end. Instead, it repeatedly shuffles the same two variables (\texttt{num\_parents} and \texttt{num\_offspring}) between three equivalent-but-wrong formulas:

\begin{itemize}
    \item Turn~7: \texttt{num\_offspring = int(10 * 0.2) = 2} $\to$ population = 2
    \item Turn~8: \texttt{num\_parents = int(10 * 0.2) = 2}, offspring from remaining 8 $\to$ population = 8
    \item Turn~9: \texttt{num\_parents = int(10 * 0.8) = 8}, offspring from remaining 2 $\to$ population = 2
\end{itemize}

\noindent
Each failed fix becomes part of the conversation context, and the model patches on top of its own previous patches. The accumulating broken code fragments prevent the model from stepping back to reason holistically about the system. A correct fix would involve ensuring that \texttt{crossover()} always produces exactly \texttt{population\_size} individuals (e.g., by selecting \texttt{population\_size / 2} parent pairs and producing 2 children each), but this requires examining the crossover function's implementation rather than adjusting the evolution loop formulas.

\vspace{0.25em}
\noindent
The model's own erroneous fixes accumulate in the conversation history, creating a self-reinforcing cycle where each patch is constrained by the frame established by previous patches. In the end, the user makes a decision to abandon the conversation and ``start a new chat.''

%% ==================== EXAMPLE 7 ====================
\subsubsection{Example 7: Reasoning trace carryover}
\label{app:ex-reasoning-spillover}

\textbf{Generator Model:} Qwen3-4B \quad \textbf{Domain:} Coding \quad \textbf{Context Pollution:} Reasoning trace from prior turn gets amplified in a subsequent turn

\vspace{0.5em}
\noindent
In Turn~1, the user asks about components for a recipe class. The model's reasoning trace produces a runaway list of ``Also [include this]\ldots'' items when enumerating possible features. By Turn~3 (``What about user interaction?''), the FC model's reasoning trace has grown to 50+ ``Also\ldots'' items, producing ${\sim}800$ words of largely redundant output, while the AO response---free from the prior pathological trace---produces a concise, structured response.

\noindent\rule{\linewidth}{0.4pt}

\vspace{0.75em}

% --- Turn 1 (reasoning trace with "Also..." pattern) ---
\noindent
\textbf{[Turn 1]} \\
\textbf{USER:} \\
What are the components that a class to hold recipes should have? It should follow h-recipes, be able to share in SEO schemas and be a flexible tool for internet use. What else should it support?

\vspace{0.25em}

\noindent
\textbf{ASSISTANT (Full Context) --- Reasoning Trace:} \\
\begin{quote}\small
\texttt{<think>} \\
\ldots We should enumerate components: attributes like name, description, ingredients, steps, preparation time, cooking time, total time, servings, nutrition info, images, video, tags, source, author, rating, comments, categories\ldots

\vspace{0.3em}
\colorbox{red!15}{Also} ability to handle special fields: allergens, dietary categories (vegan, gluten-free). \\
\colorbox{red!15}{Also} support for image lazy loading, alt attributes. \\
\colorbox{red!15}{Also} support for caching, version control, serialization, import/export (CSV, JSON, XML). \\
\colorbox{red!15}{Also} support for unit conversions. \\
\colorbox{red!15}{Also} support for front-end frameworks. \\
\colorbox{red!15}{Also} mention ``h-recipe'' is microformat for recipes. \\
\colorbox{red!15}{Also} mention that support for SEO: JSON-LD markup for Google, rich snippets. \\
\colorbox{red!15}{Also} mention multi-step instructions, sub-steps. Should support sub-recipes. \\
\colorbox{red!15}{Also} consider integration with Alexa or Google Assistant voice cooking instructions. \\
\colorbox{red!15}{Also} support for generating share links: WhatsApp, Facebook, email. \\
\colorbox{red!15}{Also} support for ``public'' vs ``private'' recipes. \\
\colorbox{red!15}{Also} support for ``alternateName''. \\
\textit{30 ``Also\ldots'' items total} \\
\texttt{</think>}

\textcolor{red}{\textit{The reasoning trace establishes a pathological ``Also [add X]\ldots'' enumeration pattern, listing features without structure or prioritization.}}
\end{quote}

\vspace{0.75em}

% --- Turn 3 (user interaction question) ---
\noindent
\textbf{[Turn 3]} \\
\textbf{USER:} \\
What about user interaction (reviews, scores etc.)?

\vspace{0.25em}

\noindent
\textbf{ASSISTANT (Full Context) --- Reasoning Trace:} \\
\begin{quote}\small
\texttt{<think>} \\
Need to elaborate about user interaction: reviews, scores, likes, ratings, comments. Should support methods for adding, editing, retrieving reviews\ldots

\vspace{0.3em}
\colorbox{red!15}{Also} ability to flag. Also include `helpful' votes. Also handle moderation. \\
\colorbox{red!15}{Also} method to get aggregated rating. Also maybe track `review\_count'. \\
\colorbox{red!15}{Also} support for star rating system. Also support for user-specified rating: 1--5.\\
\colorbox{red!15}{Also} support for `score' as numeric maybe 1--10. \\
\colorbox{red!15}{Also} support for `tags' like `vegan', `gluten-free'. \\
\colorbox{red!15}{Also} support for `user\_interaction' flags: `likes', `dislikes', `favorites'. \\
\colorbox{red!15}{Also} ability to export reviews to JSON. \\
\colorbox{red!15}{Also} support for comments thread (nested). \\
\colorbox{red!15}{Also} maybe support `review\_published\_date'. \\
\colorbox{red!15}{Also} support for `moderation status' like `approved', `pending', `rejected'. \\
\colorbox{red!15}{Also} ability to fetch reviews by user, by rating threshold. \\
\colorbox{red!15}{Also} support for `review\_score' numeric and star mapping. \\
\textit{50+ ``Also\ldots'' items total} \\
\texttt{</think>}

\textcolor{red}{\textit{Reasoning trace carries over the enumerative ``Also\ldots'' pattern from Turn~1, producing 50+ items listed in this way.}}

\vspace{0.3em}
[\textit{Final answer is a bloated skeleton class with redundant attributes and methods, reflecting the unstructured enumeration from the reasoning trace.}]
\end{quote}

\vspace{0.25em}

\noindent
\textcolor{brown}{\textbf{ASSISTANT (Assistant-Omitted Context) --- Reasoning Trace:}} \\
\begin{quote}\small
\textcolor{brown}{\texttt{<think>}} \\
\textcolor{brown}{Okay, the user is asking about user interaction features like reviews and scores for a recipe class. They mentioned the class should follow h-recipes, be SEO-friendly, and flexible. So, I need to think about how to extend the class to handle user interactions.}

\vspace{0.3em}
\textcolor{brown}{First, I remember that the previous class had components like title, ingredients, instructions, etc. Now, adding reviews and scores would require new attributes. Maybe a list of reviews, each with a user, rating, comment, and date. Also, a score average would be useful for SEO and user experience.}

\vspace{0.3em}
\textcolor{brown}{I should include methods to add reviews, calculate the average score, and maybe retrieve them. But the user said not to fill in logic, just outline the structure. So, I'll add variables like self.reviews and self.average\_score\ldots}

\vspace{0.3em}
\textcolor{brown}{I need to make sure the class is extensible. Maybe include a method to get reviews, and a way to update the average score. But again, just the structure. So, variables for reviews list and average score. Methods to add reviews and calculate the average. Also, perhaps a way to retrieve all reviews.} \\
\textcolor{brown}{\texttt{</think>}}
\end{quote}

\vspace{0.25em}

\noindent
\textcolor{brown}{\textbf{ASSISTANT (Assistant-Omitted Context) --- Response:}} \\
\begin{quote}\small
\textcolor{brown}{To support user interaction (reviews, scores, etc.), the class should include:}

\vspace{0.3em}
\textcolor{brown}{\textbf{Variables}} \\
\textcolor{brown}{\texttt{self.reviews}: A list of user reviews (e.g., \texttt{[\{"user": "john\_doe", "rating": 5, "comment": "Amazing recipe!", "date": "2023-10-05"\}]}).} \\
\textcolor{brown}{\texttt{self.average\_score}: A float representing the average rating (e.g., \texttt{4.5}).} \\
\textcolor{brown}{\texttt{self.total\_reviews}: An integer tracking the number of reviews.}

\vspace{0.3em}
\textcolor{brown}{\textbf{Methods}} \\
\textcolor{brown}{\texttt{add\_review(user, rating, comment)}: Adds a new review to the list.} \\
\textcolor{brown}{\texttt{calculate\_average\_score()}: Computes the average rating from reviews.} \\
\textcolor{brown}{\texttt{get\_reviews()}: Returns all reviews.} \\
\textcolor{brown}{\texttt{update\_average\_score()}: Recalculates the average score after adding/removing reviews.}

\vspace{0.3em}
\textcolor{brown}{\textbf{SEO/Schema Support}} \\
\textcolor{brown}{Include structured data for reviews (e.g., \texttt{Review} schema) in metadata. Allow embedding reviews in HTML for search engines to index.}

\vspace{0.3em}
\textcolor{brown}{\textbf{Flexibility}} \\
\textcolor{brown}{Support multiple user identifiers (e.g., usernames, IDs). Allow optional filtering of reviews by date, rating, or user.}

\vspace{0.3em}
\textcolor{brown}{This structure ensures the class can handle user interactions while remaining compatible with SEO standards and scalable for internet use.}
\end{quote}

\noindent\rule{\linewidth}{0.4pt}

\vspace{0.5em}
\noindent
\textbf{Analysis.}
The FC model's reasoning trace in Turn~1 establishes a pathological ``Also [add X]'' enumeration pattern. When this output is included in context for subsequent turns, the model's reasoning trace reproduces and amplifies the pattern: Turn~1 has ${\sim}30$ ``Also\ldots'' items, Turn~3 has 50+. The resulting FC response is ${\sim}800$ words of largely redundant feature suggestions, while the AO response, free from the `Also [add X]'' enumeration pattern, produces a concise, structured response.

\vspace{0.25em}
\noindent
\textbf{Judge Quality Scores:} FC = 3.0, AO = 8.0.

\subsubsection{Example 8: Hallucinated code carryover}
\label{app:ex-vico-api}

\textbf{Generator Model:} GPT-5.2 \quad \textbf{Domain:} Coding \quad \textbf{Context Pollution:} Hallucinated import paths persist across turns

\vspace{0.5em}
\noindent
In a multi-turn conversation about building an Android chart using the Vico library, the model introduces hallucinated import paths in Turn~3, importing from fabricated \texttt{compose.m3.*} package paths (e.g., \texttt{compose.m3.chart.bar.barChart}, \texttt{compose.m3.axis.rememberBottomAxis}). In Turn~4, the user explicitly corrects the API, providing the exact Vico~1.6.5 syntax with correct function names and dependencies. The FC model acknowledges the correction (``You're right'') and updates the function names (e.g., \texttt{barChart()} $\to$ \texttt{columnChart()}) but \emph{retains the hallucinated \texttt{compose.m3.*} import paths} from its own Turn~3 response. The AO model, without the prior hallucinated code in context, produces the correct import paths.

\noindent\rule{\linewidth}{0.4pt}

\vspace{0.75em}

% --- Turn 3 ---
\noindent
\textbf{[Turn 3]} \\
\textbf{ASSISTANT (Full Context):} \\
\begin{quote}\small
[\textit{Produces Vico chart code with hallucinated import paths:}]

\vspace{0.3em}
\texttt{\colorbox{red!15}{import com.patrykandpatrick.vico.compose.m3.axis.rememberBottomAxis}} \\
\texttt{\colorbox{red!15}{import com.patrykandpatrick.vico.compose.m3.axis.rememberStartAxis}} \\
\texttt{\colorbox{red!15}{import com.patrykandpatrick.vico.compose.m3.chart.Chart}} \\
\texttt{\colorbox{red!15}{import com.patrykandpatrick.vico.compose.m3.chart.bar.barChart}} \\
\texttt{import com.patrykandpatrick.vico.core.entry.ChartEntryModelProducer}

\textcolor{red}{\textit{The \texttt{compose.m3.*} paths do not exist in the Vico library; the correct paths use \texttt{compose.chart.*} and \texttt{compose.axis.*}.}}
\end{quote}

\vspace{0.75em}

% --- Turn 4 (User correction) ---
\noindent
\textbf{[Turn 4]} \\
\textbf{USER:} \\
\begin{quote}\small
That's actually not right. Here is the proper way to use Vico 1.6.5:

\vspace{0.3em}
\texttt{Chart(} \\
\texttt{\quad chart = columnChart(),} \\
\texttt{\quad chartModelProducer = chartEntryModelProducer1,} \\
\texttt{\quad startAxis = startAxis(),} \\
\texttt{\quad bottomAxis = bottomAxis(),} \\
\texttt{)}
\end{quote}

\vspace{0.75em}

% --- Turn 4 (FC response) ---
\noindent
\textbf{ASSISTANT (Full Context):} \\
\begin{quote}\small
You're right---with Vico 1.6.5\ldots

\vspace{0.3em}
\texttt{\colorbox{red!15}{import com.patrykandpatrick.vico.compose.m3.axis.bottomAxis}} \\
\texttt{\colorbox{red!15}{import com.patrykandpatrick.vico.compose.m3.axis.startAxis}} \\
\texttt{\colorbox{red!15}{import com.patrykandpatrick.vico.compose.m3.chart.Chart}} \\
\texttt{\colorbox{red!15}{import com.patrykandpatrick.vico.compose.m3.chart.column.columnChart}}

\textcolor{red}{\textit{Function names updated (\texttt{barChart} $\to$ \texttt{columnChart}, \texttt{rememberBottomAxis} $\to$ \texttt{bottomAxis}), but the hallucinated \texttt{compose.m3.*} paths from Turn~3 persist.}}
\end{quote}

\vspace{0.25em}

\noindent
\textcolor{brown}{\textbf{ASSISTANT (Assistant-Omitted Context):}} \\
\begin{quote}\small
\textcolor{brown}{\texttt{import com.patrykandpatrick.vico.compose.chart.Chart}} \\
\textcolor{brown}{\texttt{import com.patrykandpatrick.vico.compose.chart.column.columnChart}} \\
\textcolor{brown}{\texttt{import com.patrykandpatrick.vico.compose.axis.horizontal.bottomAxis}} \\
\textcolor{brown}{\texttt{import com.patrykandpatrick.vico.compose.axis.vertical.startAxis}} \\
\textcolor{brown}{\texttt{import com.patrykandpatrick.vico.core.entry.ChartEntryModelProducer}} \\
\textcolor{brown}{\texttt{import com.patrykandpatrick.vico.core.entry.entriesOf}}

\textcolor{brown}{[\textit{Correct import paths: \texttt{compose.chart.*}, \texttt{compose.axis.horizontal.*}, \texttt{compose.axis.vertical.*}.}]}
\end{quote}

\noindent\rule{\linewidth}{0.4pt}

\vspace{0.5em}
\noindent
\textbf{Analysis.}
In Turn~4, the hallucinated \emph{import paths}  \texttt{compose.m3.*} are carried directly from its own Turn~3 response. The hallucinated paths persist into Turn~5 as well. The AO model, without the Turn~3 hallucination in context, produces the correct Vico import structure.

%% ==================== MATH EXAMPLE 1 (EXAMPLE 9) ====================
\subsubsection{Example 9: Invented assumption carryover}
\label{app:ex-johns-planks}

\textbf{Generator Model:} Qwen3-4B \quad \textbf{Domain:} Math (GSM8K) \quad \textbf{Context Pollution:} Hallucinated assumption carried across turns leads to an incorrect answer.

\vspace{0.5em}
\noindent
The problem asks how much money John makes selling planks cut from a tree.
When the user gives only the height of the tree in Turn~2, the model invents a 3D cylinder model of the tree with a made-up 10-inch diameter.
Each subsequent turn references Turn~1's hallucinated diameter length.

\noindent\rule{\linewidth}{0.4pt}

\vspace{0.75em}

% --- Turn 2 ---
\noindent
\textbf{[Turn 2]} \\
\textbf{USER:} \\
John is starting with an 80-foot tree.

\vspace{0.25em}

\noindent
\textbf{ASSISTANT (Full Context):} \\
\begin{quote}\small

\vspace{0.3em}
Assuming the tree is a \colorbox{red!15}{\textbf{cylinder}} with a \colorbox{red!15}{\textbf{diameter of 10 inches}} ($\approx 0.833$ feet) and height of 80 feet:
\[
\text{Volume} = \pi \times \left(\frac{d}{2}\right)^2 \times h \approx \colorbox{red!15}{44.2 \text{ cubic feet}}
\]

\textcolor{red}{\textit{Both the cylinder and the 10-inch diameter are assumptions that the model creates.}}
\end{quote}

\vspace{0.75em}

% --- Turn 3 ---
\noindent
\textbf{[Turn 3]} \\
\textbf{USER:} \\
he can use 80\% of the tree to make logs.

\vspace{0.25em}

\noindent
\textbf{ASSISTANT (Full Context):} \\
\begin{quote}\small
\colorbox{red!15}{Usable Volume $= 44.2 \times 0.8 = 35.4$ cubic feet.}

\textcolor{red}{$\leftarrow$ \textit{References the made-up 44.2 from Turn~2.}}
\end{quote}

\vspace{0.75em}

% --- Turn 4 ---
\noindent
\textbf{[Turn 4]} \\
\textbf{USER:} \\
he cuts the usable portion into 4-foot logs.

\vspace{0.25em}

\noindent
\textbf{ASSISTANT (Full Context):} \\
\begin{quote}\small
Volume per 4-foot log: $\pi \times (0.4165)^2 \times 4 \approx 2.17$ cubic feet.

\vspace{0.3em}
Number of logs: \colorbox{red!15}{$35.4 / 2.17 \approx 16.3$ logs.}

\textcolor{red}{$\leftarrow$ \textit{Uses the 35.4 from Turn~3.}}
\end{quote}

\vspace{0.75em}

% --- Turn 5 ---
\noindent
\textbf{[Turn 5]} \\
\textbf{USER:} \\
he turns each 4-foot log into 5 planks.

\vspace{0.25em}

\noindent
\textbf{ASSISTANT (Full Context):} \\
\begin{quote}\small
Total planks: \colorbox{red!15}{$16.3 \times 5 = 81.5$ planks.}

\textcolor{red}{$\leftarrow$ \textit{Uses the 16.3 from Turn~4.}}
\end{quote}

\vspace{0.75em}

% --- Turn 6 ---
\noindent
\textbf{[Turn 6]} \\
\textbf{USER:} \\
each plank sells for \$1.20.

\vspace{0.25em}

\noindent
\textbf{ASSISTANT (Full Context):} \\
\begin{quote}\small
Total earnings:
\[
\colorbox{red!15}{$81.5 \times 1.20 = \$97.80$}
\]

\colorbox{red!15}{\textbf{John makes \$97.80 from selling the planks.}}

\textcolor{red}{\textit{Response is incorrect due to following an assumption made all the way back in Turn~2: $44.2 \rightarrow 35.4 \rightarrow 16.3 \rightarrow 81.5 \rightarrow \$97.80$ instead of \$96.}}
\end{quote}

\vspace{0.25em}

\noindent
\textcolor{brown}{\textbf{ASSISTANT (Assistant-Omitted Context):}} \\
\begin{quote}\small
\textcolor{brown}{\textbf{Breakdown:}\\
1. Usable tree length: 80\% of 80 feet $= \textbf{64}$ feet.\\
2. Number of 4-foot logs: $64 \div 4 = \textbf{16}$ logs.\\
3. Total planks: $16 \times 5 = \textbf{80}$ planks.\\
4. Total revenue: $80 \times \$1.20 = \textbf{\$96}$.}
\end{quote}

\noindent\rule{\linewidth}{0.4pt}

\vspace{0.5em}
\noindent
\textbf{Analysis.}
The model hallucinates assumptions in Turn~2 that ultimately lead to an incorrect answer in Turn~6. Once committed to the hallucinated assumptions, every subsequent turn references the previous turn's answer.
\vspace{0.25em}
\noindent
\textbf{Gold Answer:} \$96. \quad \textbf{FC Answer:} \$97.80 (incorrect). \quad \textbf{AO Answer:} \$96 (correct).

%% ==================== MATH EXAMPLE 2 (EXAMPLE 10) ====================
\subsubsection{Example 10: Intermediate answer misapplied}
\label{app:tea-party}

\textbf{Generator Model:} GPT-5.2 \quad \textbf{Domain:} Math (GSM8K) \quad \textbf{Context Pollution:} Intermediate answer overwrites a cumulative calculation.

\vspace{0.5em}
\noindent
This problem requires keeping track of the total amount of tea consumed by a group of students. Turn~3 introduces 4~students drinking 1.5~quarts each, and Turn~4 introduces 16~students drinking 2~quarts each, with 10~quarts remaining.
The correct total consumption is $4 \times 1.5 + 16 \times 2 = 6 + 32 = 38$ quarts, giving a starting amount of $38 + 10 = 48$ quarts $= 12$ gallons.
The model computes a correct intermediate answer of 4~gallons in Turn~3 (using only the first group), but in Turn~4 it treats the new information as a \emph{replacement} rather than an \emph{addition}, computing $16 \times 2 + 10 = 42$ quarts $= 10.5$ gallons, discarding the 6~quarts consumed by the first group.

\noindent\rule{\linewidth}{0.4pt}

\vspace{0.75em}

% --- Turn 3 ---
\noindent
\textbf{[Turn 3]} \\
\textbf{USER:} \\
four students consumed 1.5 quarts of tea each

\vspace{0.25em}

\noindent
\textbf{ASSISTANT (Full Context):} \\
\begin{quote}\small
Four students drank $1.5$ quarts each, so the amount consumed was:
\[
4 \times 1.5 = 6 \text{ quarts}
\]

Since $10$ quarts were remaining after the party, the amount at the start was:
\[
10 + 6 = 16 \text{ quarts}
\]

Convert quarts to gallons ($4$ quarts $= 1$ gallon):
\[
16 \div 4 = 4 \text{ gallons}
\]

\colorbox{red!15}{\textbf{There were 4 gallons of tea at the start of the party.}}

\textcolor{red}{\textit{Correct given the information so far, but this intermediate answer confuses the model in Turn~4.}}
\end{quote}

\vspace{0.75em}

% --- Turn 4 ---
\noindent
\textbf{[Turn 4]} \\
\textbf{USER:} \\
16 students each drank 2 quarts of tea

\vspace{0.25em}

\noindent
\textbf{ASSISTANT (Full Context):} \\
\begin{quote}\small
If \textbf{16 students each drank 2 quarts}, then the amount consumed was:
\[
\colorbox{red!15}{$16 \times 2 = 32$ quarts}
\]

\textcolor{red}{$\leftarrow$ \textit{Only counts the new group; drops the 6 quarts from the 4 students in Turn~3}}

\vspace{0.3em}
With \textbf{10 quarts remaining} after the party, the starting amount was:
\[
\colorbox{red!15}{$32 + 10 = 42$ quarts}
\]

Convert to gallons ($4$ quarts $= 1$ gallon):
\[
42 \div 4 = 10.5 \text{ gallons}
\]

\colorbox{red!15}{\textbf{There were 10.5 gallons of tea at the start of the party.}}

\textcolor{red}{\textit{Incorrect. The model interprets Turn~4 as replacing rather than adding to Turn~3.}}
\end{quote}

\vspace{0.25em}

\noindent
\textcolor{brown}{\textbf{ASSISTANT (Assistant-Omitted Context):}} \\
\begin{quote}\small
\textcolor{brown}{Total tea consumed $= 4 \times 1.5 + 16 \times 2 = 6 + 32 = 38$ quarts.\\
Starting amount $=$ consumed $+$ remaining $= 38 + 10 = 48$ quarts.\\
Convert to gallons: $48 \div 4 = 12$ gallons.}

\vspace{0.3em}
\textcolor{brown}{\textbf{There were 12 gallons of tea at the start of the party.}}
\end{quote}

\noindent\rule{\linewidth}{0.4pt}

\vspace{0.5em}
\noindent
\textbf{Analysis.}
The FC model's Turn~3 response provides an intermediate answer of 4~gallons based on partial information. When Turn~4 introduces additional information, the model restarts the calculation using \textit{only} the new group ($16 \times 2 = 32$ quarts) rather than accumulating both groups ($6 + 32 = 38$ quarts).

\vspace{0.25em}
\noindent
\textbf{Gold Answer:} 12 gallons. \quad \textbf{FC Answer:} 10.5 gallons (incorrect). \quad \textbf{AO Answer:} 12 gallons (correct).

%% ==================== MATH EXAMPLE 3 (EXAMPLE 11) ====================
\subsubsection{Example 11: Reasoning error causes confusion downstream}
\label{app:ex-jerry-pools}

\textbf{Generator Model:} Qwen3-4B \quad \textbf{Domain:} Math (GSM8K) \quad \textbf{Context Pollution:} Prior flawed reasoning poisons subsequent turns

\vspace{0.5em}
\noindent
The problem describes Jerry's two pools that both leak water at 4~gallons per minute. In Turn~4, the model incorrectly reasons that ``if both pools lose water at the same rate, the ratio remains constant''. However, these equivalent leak rates would preserve the \emph{difference}, not the \emph{ratio}.
In Turn~5, the FC model inherits this belief and confuses itself ``if they both lose water at the same rate, the ratio between them should stay the same\ldots the ratio changed from 2:1 to 4:1\ldots that doesn't make sense.'' This then leads to a new hallucination, where the model sets up the ``4~minutes ago'' condition as $B - 16 = 2(S - 16)$ (subtracting leaked water from current values, as if pools had \emph{less} water in the past) instead of the correct $B + 16 = 2(S + 16)$, obtaining $S = -8$ and declaring the problem ``impossible.''

\noindent\rule{\linewidth}{0.4pt}

\vspace{0.75em}

% --- Turn 4 ---
\noindent
\textbf{[Turn 4]} \\
\textbf{USER:} \\
4 minutes ago, the big pool had twice as much water as the small pool

\vspace{0.25em}

\noindent
\textbf{ASSISTANT (Full Context):} \\
\begin{quote}\small
Let me denote the current amount of water in the small pool as $S$ and the big pool as $B$.

\vspace{0.3em}
\colorbox{red!15}{If both pools lose water at the same rate (4 gallons per minute), the ratio of water}
\colorbox{red!15}{in the two pools remains constant over time.}

\textcolor{red}{\textit{Flawed reasoning, as equal leak rates preserve the difference ($B - S$), not the ratio ($B/S$).}}

\vspace{0.3em}
So if 4 minutes ago $B' = 2S'$, and both pools lost $4 \times 4 = 16$ gallons since then, we get $B = 2S$ currently as well.

\vspace{0.3em}
But substituting into $B + 16 = 2(S + 16)$: $2S + 16 = 2S + 32 \Rightarrow 16 = 32$, a contradiction.

\vspace{0.3em}
\colorbox{red!15}{\textbf{The problem as stated is underdetermined.}}

\textcolor{red}{\textit{Incorrect conclusion based on the false ``ratio stays constant'' premise.}}
\end{quote}

\vspace{0.75em}

% --- Turn 5 ---
\noindent
\textbf{[Turn 5]} \\
\textbf{USER:} \\
at this moment, the big pool has four times as much water as the small pool. how much water remains in the small pool now?

\vspace{0.25em}

\noindent
\textbf{ASSISTANT (Full Context):} \\
\begin{quote}\small
\colorbox{red!15}{If they both lose water at the same rate, the ratio between them should stay the same.}
\colorbox{red!15}{But the ratio changed from 2:1 to 4:1. That doesn't make sense.}

\textcolor{red}{\textit{Model confuses itself from Turn~4's false ``ratio stays constant'' claim.}}

\vspace{0.3em}
From the current condition: $B = 4S$.

\vspace{0.3em}
Four minutes ago, the big pool had $B - 16$ and the small pool had $S - 16$.

\[
\colorbox{red!15}{$B - 16 = 2(S - 16)$}
\]

\textcolor{red}{$\leftarrow$ \textit{The pool would have had \textbf{more} water, not less, 4~min ago, so the correct system should be $B + 16 = 2(S + 16)$}}

\vspace{0.3em}
Substituting $B = 4S$: $4S - 16 = 2S - 32 \Rightarrow 2S = -16 \Rightarrow S = -8$

\vspace{0.3em}
\colorbox{red!15}{\textbf{The problem is impossible as stated (negative water).}}

\end{quote}

\vspace{0.25em}

\noindent
\textcolor{brown}{\textbf{ASSISTANT (Assistant-Omitted Context):}} \\
\begin{quote}\small
\textcolor{brown}{Let the current amount of water in the small pool be $S$ and in the big pool be $B$.}

\textcolor{brown}{Current condition: $B = 4S$.}

\textcolor{brown}{Four minutes ago, both pools had 16 more gallons (since they each lose 4 gal/min $\times$ 4 min $= 16$ gallons):}

\textcolor{brown}{$\quad (4S + 16) = 2(S + 16)$}

\textcolor{brown}{$\quad 4S + 16 = 2S + 32$}

\textcolor{brown}{$\quad 2S = 16$}

\textcolor{brown}{$\quad S = 8$}

\vspace{0.3em}
\textcolor{brown}{\textbf{There are 8 gallons of water remaining in the small pool.}}
\end{quote}

\noindent\rule{\linewidth}{0.4pt}

\vspace{0.5em}
\noindent
\textbf{Analysis.}
The model's Turn~4 reasoning trace introduces an error: that equal leak rates preserve the ratio between two quantities; as a result, the model declares the problem underdetermined. In Turn~5, the model restates this belief, opening with ``if they both lose water at the same rate, the ratio between them should stay the same\ldots that doesn't make sense.'' The AO response, without the erroneous reasoning in context, correctly reasons that pools had \textit{more} water, not less, in the past.

\vspace{0.25em}
\noindent
\textbf{Gold Answer:} 8 gallons. \quad \textbf{FC Answer:} ``impossible'' (incorrect). \quad \textbf{AO Answer:} 8 gallons (correct).

\end{document}